\documentclass{article}

\usepackage{todonotes}
\usepackage[utf8]{inputenc} 
\usepackage[T1]{fontenc}    
\usepackage[colorlinks=true, citecolor=blue]{hyperref}       
\usepackage{url}            
\usepackage{booktabs}       
\usepackage{amsfonts}       
\usepackage{nicefrac}       
\usepackage{microtype}      
\usepackage{xcolor}         
\usepackage{graphicx}
\usepackage{amsmath}
\usepackage{physics}
\usepackage{colonequals}
\usepackage{textcomp}
\usepackage{multirow}
\usepackage{wrapfig}
\usepackage[compress, numbers]{natbib}
\usepackage{color}
\usepackage{authblk}
\usepackage[margin=1in]{geometry}
\usepackage{enumitem}
\usepackage{subcaption}

\title{OccamNet: A Fast Neural Model for Symbolic Regression at Scale}

\author{Owen Dugan$\phantom{}^{*}$, Rumen Dangovski$\phantom{}^{*}$, Allan Costa$\phantom{}^{*}$, Samuel Kim, Pawan Goyal, Joseph Jacobson, Marin Solja\v{c}i\'{c} \\
Massachusetts Institute of Technology \\
$\phantom{}^{*}$ denotes equal contribution}

\begin{document}

\maketitle

\begin{abstract}
  Neural networks’ expressiveness comes at the cost of complex, black-box models that often extrapolate poorly beyond the domain of the training dataset, conflicting with the goal of finding compact analytic expressions to describe scientific data. We introduce OccamNet, a neural network model that finds interpretable, compact, and sparse symbolic fits to data, \`{a} la Occam's razor. Our model defines a probability distribution over functions with efficient sampling and function evaluation. We train by sampling functions and biasing the probability mass toward better fitting solutions, backpropagating using cross-entropy matching in a reinforcement-learning loss. OccamNet can identify symbolic fits for a variety of problems, including analytic and non-analytic functions, implicit functions, and simple image classification, and can outperform state-of-the-art symbolic regression methods on real-world regression datasets. Our method requires a minimal memory footprint, fits complicated functions in minutes on a single CPU, and scales on a GPU.
\end{abstract}

\section{Introduction}
Deep learning has revolutionized a variety of complex tasks, ranging from language modeling to computer vision ~\cite{lecun2015deeplearning}. Key to this success is designing a large search space in which many local minima sufficiently approximate given data ~\cite{choromanska2014loss}. This requires large, complex models, which often conflicts with the goals of sparsity and interpretability, making neural nets not optimally suited for a myriad of physical and computational problems with compact and interpretable underlying mathematical structures~\cite{Lample2020Deep}. Neural networks also might not preserve desired physical properties (e.g., time invariance) and are typically unable to generalize much beyond observed data.

In contrast, Evolutionary Algorithms (EAs), in particular genetic programming, can find interpretable, compact models that explain observed data~\cite{ schmidt2009distilling, Udrescu_2020, poli2008field}. EAs have been employed as an alternative to gradient descent for optimizing neural networks in what is known as \textit{neuroevolution} ~\cite{265960, 10.1145/2330163.2330207,such2017deep}. Recently, evolutionary strategies that model a probability distribution over parameters, updating this distribution according to their own best samples (i.e., selecting the fittest), were found advantageous for optimization on high-dimensional spaces, including neural networks' hyperparameters \cite{hansen2016cma, loshchilov2016cmaes}. 

A number of evolution-inspired, probability-based models have been explored for Symbolic Regression \cite{mckay2010grammar}. Along these lines, \citet{petersen2021deep} explore deep symbolic regression by using an RNN to define a probability distribution over a space of expressions and sample from it using autoregressive expression generation. More recently, \citet{BiggioNeuroSymResScales} have pretrained Transformer models that receive input-output pairs as input and return functional forms that could fit the data. In the related field of program synthesis, probabilistic program induction using domain-specific languages \cite{NIPS2018_7845, NIPS2018_8006, NIPS2019_9116} has proven successful. \citet{balog2016deepcoder} first train a machine learning model to predict a DSL based on input-output pairs and then use methods from satisfiability modulo theory~\cite{SolarLezama:EECS-2008-177} to search the space of programs built using the predicted DSL.

One approach to symbolic regression which can integrate well with deep learning is the Neural Arithmetic Logic Unit (NALU) and related models \cite{NIPS2018_8027, Madsen2020Neural}, which provide neural inductive bias for arithmetic in neural networks by shaping a neural network towards a gating interpretation of the linear layers. Neural Turing Machines~\cite{Graves2014NeuralTM, Graves2016HybridCU} and their stable versions~\cite{CollierBeel2018} can also discover interpretable programs, simulated by neural networks connected to external memory, via observations of input-output pairs. Another option is Equation Learner (EQL) Networks \cite{EQLOriginal,EQLWithDivision,Kim2019IntegrationON}, which identify symbolic fits to data by training a neural network with symbolic activation functions, such as multiplication or trigonometric functions. However, these methods require strong regularization to be interpretable. NALUs and to a lesser extent EQL Networks can also only use a restricted set of differentiable primitive functions, and Neural Turing Machines do not include the concept of a ``primitive.'' Additionally, these methods often converge to local minima and often converge to uninterpretable models unless they are carefully regularized for sparsity.

In this paper, we consider a mixed approach of connectionist and sample-based optimization for symbolic regression. We propose a neural network architecture, OccamNet, which preserves key advantages of EQL networks and other neural-integrable symbolic regression frameworks while addressing many of these architectures' limitations. Inspired by neuroevolution, our architecture uses a neural network to model a probability distribution over functions. We optimize the model by sampling to compute a reinforcement-learning loss, tunable for different tasks, based on the training method presented in Risk-Seeking Policy Gradients \cite{petersen2021deep}. Our method handles non-differentiable and implicit functions, converges to sparse, interpretable symbolic expressions, and can work across a wide range of symbolic regression problems. 
Further, OccamNet consistently outperforms other symbolic regression algorithms in testing on real-world regression datasets. We also introduce a number of strategies to induce compactness and simplicity a la Occam's Razor.

The main goal of this study is not to replace existing symbolic regression methods, but rather to create a novel hybrid approach that combines the strengths of neural networks and evolutionary algorithms. Our proposed OccamNet method consistently achieves state-of-the-art performance across a wide range of tasks, including a diverse range of synthetic functions, simple programs, raw data classification, and real-world tabular tasks. Additionally, we show how to connect OccamNet to state-of-the-art pretrained vision models, such as ResNets~\citep{he2016deep}. OccamNet has also shown promise in discovering quantitative and formal laws in social sciences, indicating its potential to aid scientific research~\citep{balla2022ai}. By striking a delicate balance between expressiveness and interpretability, OccamNet presents a versatile and powerful solution for symbolic regression challenges.

\section{Model Architecture} \label{sec:model}

\begin{figure*}[!t]
    \centering  
    \includegraphics[width=\textwidth]{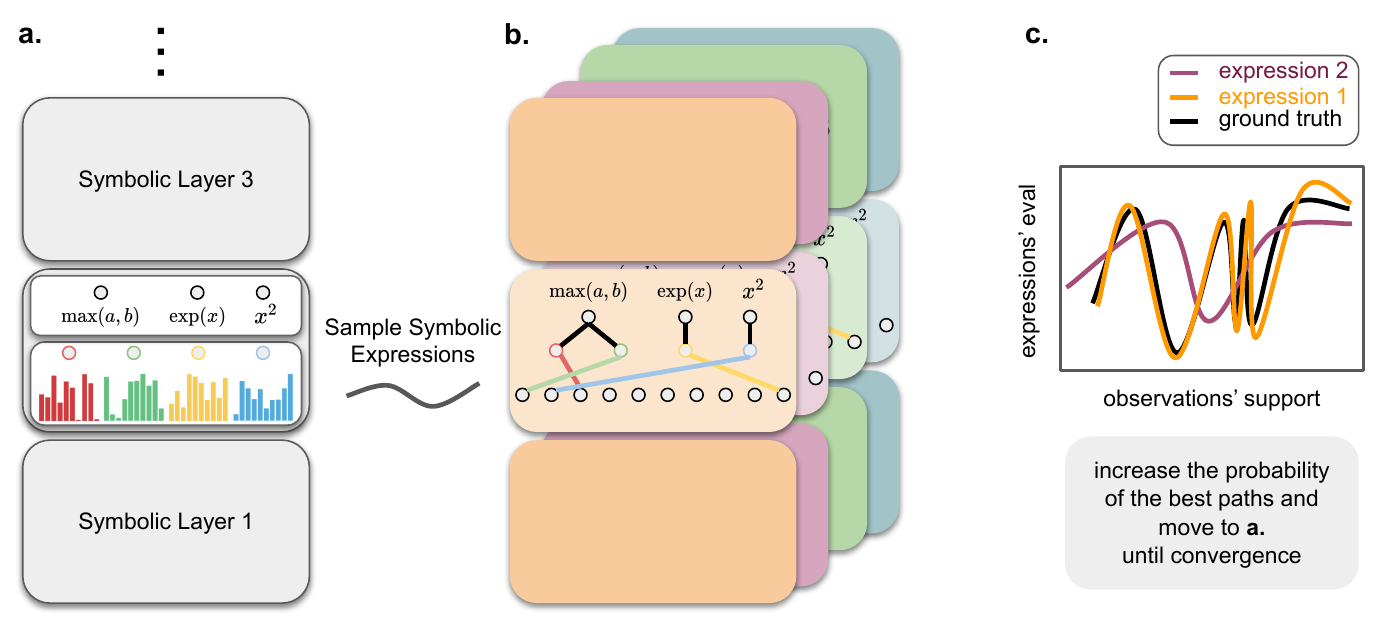}
    \caption{OccamNet architecture and training. \textbf{a.} OccamNet is a stack of ``symbolic layers'' each described by a collection of learned distributions (over the neurons from the previous layer) for each neuron within the layer, as well as non-linearities that are collections of symbolic expressions. \textbf{b.} By sampling from each distribution independently, we are able to sample paths from OccamNet that represent symbolic expressions, ready for evaluation. \textbf{c.} We evaluate each expression by feeding the observations' support data and comparing the outputs with the ground truth. The probability of the best paths is increased and the process is repeated until convergence.
    }   
    \label{fig:pretty_intro}
\end{figure*}

In Figure~\ref{fig:pretty_intro} we sketch the OccamNet architecture and the method for training it, before following with a more detailed description. We can view OccamNet as a fully-connected feed-forward network (a stack of fully connected linear layers with non-linearities) with two key unique features. First, the parameters of the linear layer are substituted with a learned probability distribution associated with the neurons from the preceding layer for each neuron within the layer. Second, the non-linearities form a collection of symbolic expressions. Thus we obtain a collection of ``symbolic layers'' that form OccamNet (Figure 1a). Figure 1b shows a variety of symbolic expressions, representing paths within OccamNet from sampling each probability distribution independently. Figure 1c shows OccamNet's training objective, which increases the probability of the paths that are closest to the ground truth. Below we formalize OccamNet in detail.

\subsection{Layer structure}

A dataset $\mathcal{D}=\{( \vec{x}_p, \vec{y}_p )\}_{p=1}^{|\mathcal{D}|}$ consists of pairs of inputs $\vec{x}_p$ and targets $\vec{y}_p=\vec{f}^*\left(\vec{x}_p\right) =[f^*_{(0)}(\vec{x}_p),\dots, f^*_{(v-1)}(\vec{x}_p)]^\top$.
Our goal is to compose either $f_{(i)}^*(\cdot)$ or an approximation of $f_{(i)}^*(\cdot)$ using a predefined collection of $N$ primitive functions $\mathbf{\Phi}=\{\phi_i(\cdot,\ldots,\cdot)\}_{i=1}^N$. Note that primitives can be repeated, their arity (number of arguments) is not restricted to one, and they may operate over different domains.
The concept of a set of primitives $\Phi$ is similar to that of DSL, \emph{domain-specific languages}~\cite{10.5555/1809745}.

\begin{figure*}[h]
    \centering  
    \includegraphics[width=\textwidth]{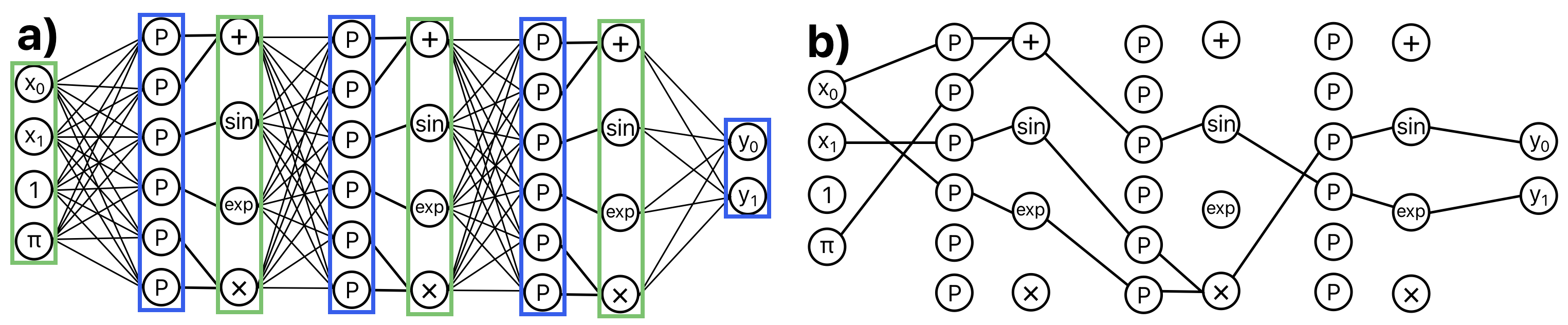}

    \caption{(\emph{a}) A two-output network model with depth $L=2$, $\vec x = [x_0, x_1]$, user-selected constants $\mathcal{C} = [1, \pi ],$ and set of primitive functions  $\mathbf{\Phi} =\left( +(\cdot,\cdot), \sin(\cdot), \exp(\cdot), \times(\cdot,\cdot) \right)$. Boxed in blue are the arguments sublayers (composed of P-nodes). For each arguments sublayer, the associated image sublayer (composed of the basis functions from $\mathbf{\Phi}$) is boxed in green and to the right of the corresponding arguments sublayer. Together, these two sublayers define a single hidden layer of our model. The input layer can be thought of as an image layer and the output layer can be thought as an arguments layer. (\emph{b}) An example of function-specifying directed acyclic graphs (DAGs) that can be sampled from the network in (\emph{a}). These DAGs represent the functions $y_0 = \exp[\sin(x_0+\pi)]$ and $y_1 = \sin(\exp(x_0)\sin(x_1)).$
    }   
    \label{fig:mesh1}
\end{figure*}

To solve this problem, we follow a similar approach as in EQL networks \cite{EQLWithDivision, EQLOriginal, Kim2019IntegrationON}, in which the primitives act as activation functions on the nodes of a neural network. Specifically, each hidden layer consists of an \textit{arguments} sublayer and an \textit{images} sublayer, as shown in Figure \ref{fig:mesh1}a. We use this notation because the arguments sublayer holds the inputs, or arguments, to the activation functions and the images sublayer holds the outputs, or images, of the activation functions. The primitives are stacked in the images sublayer and act as activation functions for their respective nodes. Each primitive takes in nodes from the arguments sublayer. Additionally, we use skip connections similar to those in DenseNet~\cite{huang2017densely} and ResNet~\cite{he2016deep}, concatenating image states with those of subsequent layers.

Next, we introduce a probabilistic modification of the network: instead of computing the inputs to the arguments sublayers using dense feed-forward layers, we compute them probabilistically and sample through the network. This enables many key advantages: it enforces sparsity and interpretability without requiring regularization, it allows the model to avoid backpropagating through the activation functions, thereby allowing non-differentiable and fast-growing functions in the primitives, and it helps our model avoid premature convergence to local minima.

Because they behave probabilistically, we call nodes in the arguments sublayer \emph{P-nodes}. Figure~\ref{fig:mesh1} highlights this sublayer structure, while the methods section describes the complete mathematical formalism behind it.

\subsection{Temperature-controlled connectivity}
\label{sec:connectivity}

Instead of dense linear layers, we use \emph{$T$-softmax layers}. For any temperature $T > 0,$ we define a $T$-softmax layer as a standard $T$-controlled softmax layer with weighted edges connecting an images sublayer and the subsequent arguments sublayer, in which each P-node from the arguments sublayer probabilistically samples a single edge between itself and a node in the images sublayer. Each node's sampling distribution is given by $$\mathbf{p}^{(l,i)}(T_l) = \mathrm{softmax}(\mathbf{w}^{(l,i)};T_l),$$ where $\mathbf{w}^{(l,i)}$ and $\mathbf{p}^{(l,i)}$ are the weights and probabilities for edges leading to the $i$th P-node of the $l$th layer and $T_l$ is the fixed temperature for the $l$th layer. Selecting these edges for all $T$-softmax layers produces a sparse directed acyclic graph (DAG) specifying a function $\vec{f},$ as seen in Figure~\ref{fig:mesh1}b. While controlling the temperature adjusts the entropy of the distributions over nodes, OccamNet automatically enforces sparsity by sampling a single input edge to each P-node. Adjusting the temperature has no impact on sparsity, but it allows for balancing exploration and exploitation during training.

\subsection{A neural network as a probability distribution over functions}
\label{sec:prob}

Through the temperature-controlled connectivity described above, OccamNet can be sampled to produce DAGs corresponding to functions $\vec{f}.$ Based on the weights of OccamNet, some DAGs may be sampled with higher or lower probability. In this way, OccamNet can be considered as representing a probability distribution over the set of all possible DAGs, or equivalently over all possible functions sample-able from OccamNet.

Let $\mathbf{W}=\left \{\mathbf{w}^{(l,i)} ; 1 \leq l \leq L, 1 \leq i\leq N \right \}.$ The probability of the model sampling $f_{(i)}$ as its $i$th output, $q_i(f_{(i)}|\mathbf{W}),$ is the product of the probabilities of the edges of $f_{(i)}$'s DAG. Similarly, $q(\vec{f}|\mathbf{W}),$ the probability of the model sampling $\vec{f},$ is given by the product of $\vec{f}$'s edges, or $q(\vec{f}|\mathbf{W})=\prod_{i=0}^{v-1}q_i(f_{(i)}|\mathbf{W}).$ For example, in Figure \ref{fig:mesh1}b, the probabilities of sampling the DAG shown is given by the product of the probabilities sampling each of the edges shown.

In practice, we compute an approximation of this probability which we denote $q_{apx},$ as described in Methods Section \ref{sec:probability}. We find that OccamNet performs well with this approximation. For all other sections of this paper, unless explicitly mentioned, we use $q$ to mean $q_{apx}$.

We initialize the network with weights $\mathbf{W}_\text{i}$ such that $q_{apx}(\vec{f}_1|\mathbf{W}_\text{i}) = q_{apx}(\vec{f}_2|\mathbf{W}_\text{i})$ for all $\vec{f}_1$ and $\vec{f}_2$ in $\mathcal{F}_{\mathbf{\Phi}}^L.$ After training (Section \ref{sec:optimization}), the network has weights $\mathbf{W}_\text{f}.$ The network then selects the function $\vec{f}_\text{f}$ with the highest probability $q_{apx}(\vec{f}_\text{f}|\mathbf{W}_\text{f}).$ We discuss our algorithms for initialization and function selection in the Methods section. 
A key benefit of OccamNet is that, unlike other approaches such as \citet{petersen2021deep}, it allows for efficiently identifying the function with the highest probability.

\section{Training} \label{sec:optimization}
To express a wide range of functions, we include non-differentiable and fast-growing primitives. Additionally, in symbolic regression, we are interested in finding global minima. To address these constraints, we implement a loss function and training method that combine gradient-based optimization and sampling-based strategies for efficient global exploration of the possible functions. Our loss function and training procedure are closely related to those proposed by \citet{petersen2021deep}, differing mainly in the fitness function and regularization terms.

Consider a mini-batch $\mathcal{M} = ( X, Y )$ and a sampled function from the network $\vec{f}(\cdot) \sim q(\cdot|\mathbf{W})$. We compute the \textit{fitness} of each $f_{(i)}(\cdot)$ with respect to a training pair $( \vec{x}, \vec{y} )$ by evaluating 
$$k_i\left(f_{(i)}(\vec{x}),\vec{y}\right)=(2\pi\sigma^2)^{-1/2}\exp(-\left[ f_{(i)}(\vec{x})-(\vec{y})_i\right]^2/(2\sigma^2)),$$ which measures how close $f_{(i)}(\vec{x})$ is to the target $(\vec{y})_i$. 
The total fitness is determined by summing over the entire mini-batch: $K_i\left(\mathcal{M},f_{(i)}\right) = \sum_{(\vec{x},\vec{y}) \in \mathcal{M}} k_i\left(f_{(i)}(\vec{x}),\vec{y}\right)$.

We then define the loss function \begin{equation}
\label{eq:cross_entropy}
\small
H_{q_i}[f_{(i)}, \mathbf{W}, \mathcal{M}] = - K_i\left(\mathcal{M},f_{(i)}\right)\cdot\log \left[q_i(f_{(i)}|\mathbf{W})\right].
\end{equation} as in \citet{petersen2021deep}. As in \cite{petersen2021deep}, we train the network by sampling functions, selecting the number $\lambda$ of functions with the highest fitness for each output, and performing a gradient step based on these highest-fitness functions using the loss defined in Equation \ref{eq:cross_entropy}. In practice, $\lambda$ is a critical hyperparameter to tune as it adjusts the balance between updating toward higher-fitness functions and receiving information about all sampled functions.

To improve implicit function fitting, we implement regularization terms that punish trivial solutions by reducing the fitness $K$, as discussed in the Methods (Section \ref{sec:regularization}). We also introduce regularization to restrict OccamNet to solutions that preserve units (Section \ref{sec:units}).

OccamNet can also be trained to find recurrent functions, as discussed in the Methods (Section \ref{sec:recurrence2}).

\section{Results} \label{sec:experiments}
To empirically validate our model, we first develop a diverse collection of benchmarks in four categories: \emph{Analytic Functions}, simple, smooth functions; \emph{Implicit Functions}, functions specifying an implicit relationship between inputs; \emph{Non-Analytic Functions}, discontinuous and/or non-differentiable functions; \textit{Image/Pattern Recognition}, patterns explained by analytic expressions. We then test OccamNet's performance and ability to scale on real-world symbolic regression datasets. The purpose of these experiments is to demonstrate that OccamNet can perform competitively with other symbolic regression frameworks in a diverse range of applications.

We compare OccamNet with several other symbolic regression methods: Eureqa \cite{schmidt2009distilling}, a genetic algorithm with Epsilon-Lexicase (Eplex) selection  \cite{EPLEX}, AI Feynman 2.0 (AIF) \cite{Udrescu_2020,AIFeynman2.0}, and Deep Symbolic Regression (DSR) \cite{petersen2021deep}. We do not compare to Transformer-based models such as \citet{BiggioNeuroSymResScales} because, unlike our method, these methods utilize a prespecified and immutable set of primitive functions which are not always sufficiently general to fit our experiments.
The results are shown in Tables~\ref{table:analyticbenchmarks}, \ref{table:programbenchmarks}, and \ref{table:implicitbenchmarks}, and we discuss them below. More details about the experimental setup are given in the methods section.

\begin{figure}[t]
\centering
\includegraphics[width=\textwidth]{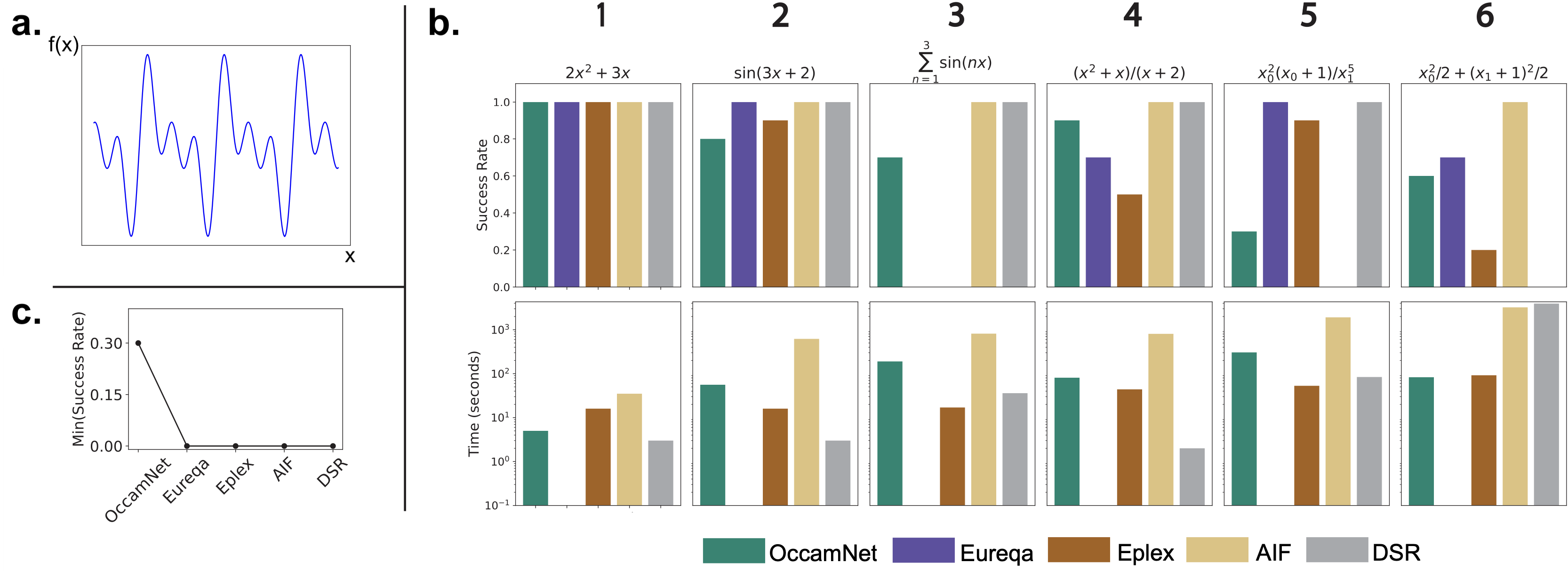}
\caption{Experiment on analytic functions. \textbf{a.} A sketch of the function $\sum_{n=1}^{3}\sin(nx)$ as an example of the analytics functions we consider in our work. \textbf{b.} Success rate (out of 10 trials) for each of the five methods considered: OccamNet, Eureqa, Eplex, AI Feynman 2.0 (AIF) and Deep Symbolic Regression (DSR) (at the top). Training time for the methods (at the bottom). Eureqa almost always finishes much more quickly than the other methods, so we do not provide training times for Eureqa. We enumerate the functions to ease the discussion. \textbf{c.} The ``worst-case'' performance for each methods, showing the minimal success rate across the six tasks.
\label{fig:analyticbenchmarks}
}
\end{figure}

\subsection{Analytic functions}\label{sec:analytic}

In Figure~\ref{fig:analyticbenchmarks} and Table~\ref{table:analyticbenchmarks} (in the Methods) we present our results on analytic functions. Figure 3a presents an analytic function that is particularly challenging for Eureqa. Figure 3b shows that OccamNet gives competitive success rate to state-of-the-art symbolic regression methods. For all methods besides OccamNet, there is at least one function for which the method gets zero accuracy; in contrast, OccamNet gets non-zero accuracy on every single
considered function (Figure 3c).

We highlight the large success rate for function 4, 
which we originally speculated could easily trick the network with the local minimum $f(x) \approx x + 1$ for large enough $x$. In contrast, as with the difficulties faced by AI Feynman 2.0, we find that OccamNet often failed to converge for function 5 because it approximated the factor $x_0^2 (x_0+1)$ to $x_0^3$; even when convergence did occur, it required a relatively large number of steps for the network to resolve this additional constant factor. Notably, Eureqa and Eplex had difficulty finding function 3. 

AI Feynman 2.0 consistently identifies many of the functions, but it struggles with function 5 and is also generally much slower than other approaches. Eplex also performs well on most functions and is fast. However, like Eureqa, Eplex struggles with functions 3 and 6. We suspect that this is because evolutionary approaches require a larger sample size than OccamNet's training procedure to adequately explore the search space. DSR consistently identifies many of the functions and is very fast. However, DSR struggles to fit Equation 6, which we suspect is because such an equation is complex but can be simplified using feature reuse. OccamNet's architecture allows such feature reuse, demonstrating an advantage of OccamNet's inductive biases.

\subsection{Non-analytic functions}

\begin{figure}[t]
\centering
\includegraphics[width=\textwidth]{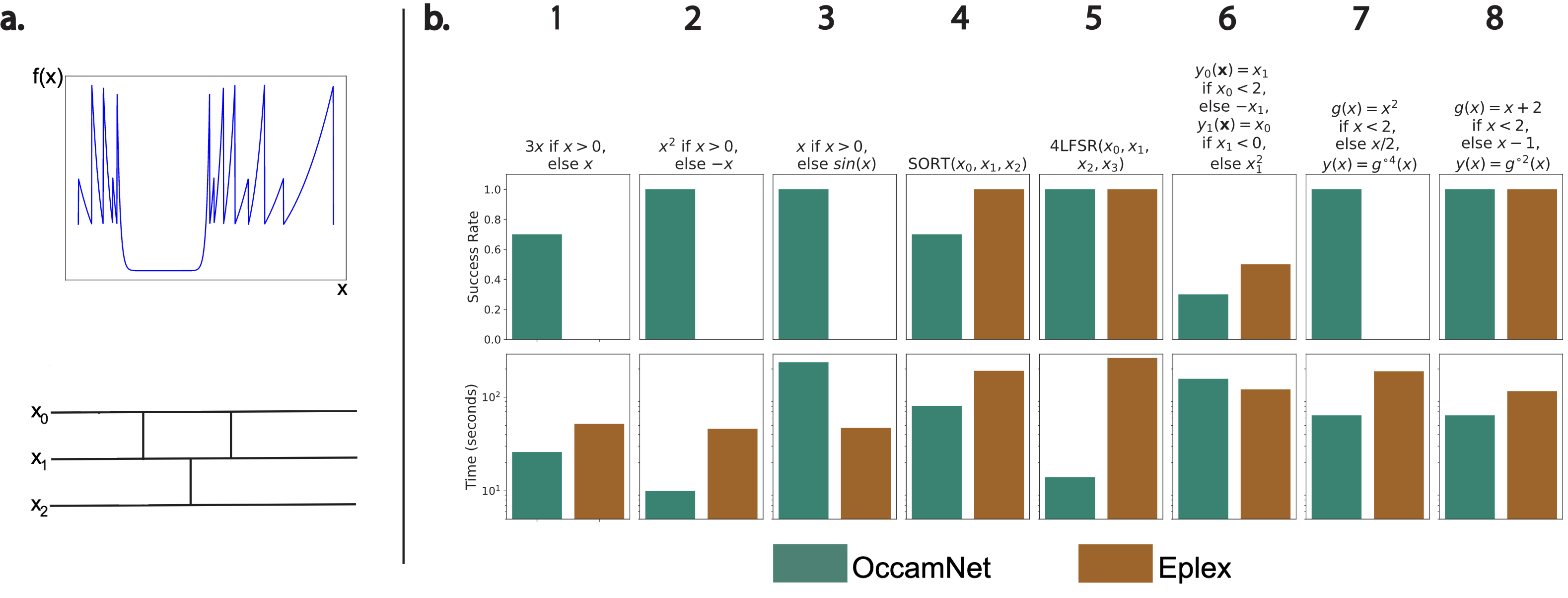}
\caption{Experiments on non-analytic functions. \textbf{a.} Two prominent examples of non-analytic functions: The challenging recursion $g(x)=x^2$ if $x < 2$, else $x/2$, $y(x) = g^{\circ 4}(x) = g(g(g(g(x))))$ (top) and a sorting circuit of three numbers (bottom). \textbf{b.} Success rate (out of 10 trials) and training time for OccamNet and Eplex. We enumerate the functions to ease the discussion.
\label{figure:programbenchmarks}
}
\end{figure}

In Figure~\ref{figure:programbenchmarks} and Table~\ref{table:programbenchmarks} (in the Methods) we benchmark the ability to find several non-differentiable, potentially recursive functions. 
From our experiments, we highlight both the network's fast convergence to the correct functional form and the discovery of the correct recurrence depth of the final expression. This is pronounced for function 7 in, which is a challenging chaotic series on which Eureqa and Eplex struggle. Interestinly, Eplex fails to identify the simpler functions 1-3 correctly. We suspect that this may be because, for these experiments, we restrict both OccamNet and Eplex to smaller expression depths. Although OccamNet is able to identify the correct functions with small expression depth, we suspect that Eplex often identifies expressions by producing more complex equivalents to the correct program and so cannot identify the correct function when restricted to simpler expressions.

We also investigated the usage of primitives such as $\mathsf{MAX}$ and $\mathsf{MIN}$ to sort numbers (function 4), obtaining relatively well-behaved final solutions: the few solutions that did not converge fail only in deciding the second component, $y_2$, of the output vector. Finally, we introduced binary operators and discrete input sets for testing function 5, a simple 4-bit Linear Feedback Shift Register (LFSR), the function $(x_0,x_1,x_2,x_3) \to(x_0+x_3 \mod 2, x_0, x_1, x_2)$, which converges fast with a high success rate.

We do not compare to AI Feynman 2.0 in these experiments because AI Feynman does not support the required primitive functions.

\subsection{Implicit Functions and Image Recognition}

\begin{figure}[t]
\centering
\includegraphics[width=\textwidth]{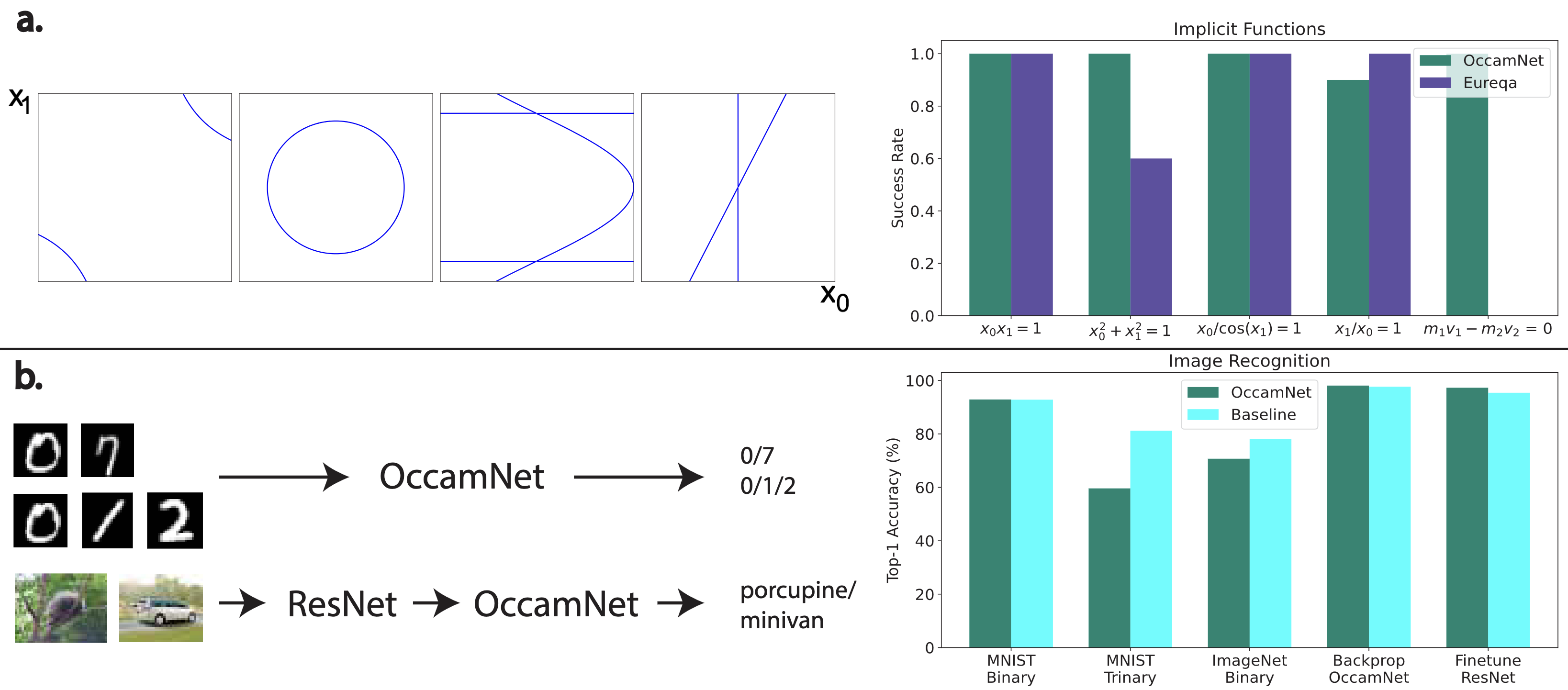}
\caption{Experiments on implicit functions and standard vision benchmarks. \textbf{a.} Examples of implicit functions' loci (left) and the corresponding success rate on a suite of implicit functions (right). \textbf{b.} Examples of image recognition tasks (left) and the best accuracy from 10 trials for both OccamNet and the baseline. The baseline for MNIST Binary/ Trinary and ImageNet Binary is the HeuristicLab symbolic regression algorithm~\cite{wagner2014}. The baseline for Backprop OccamNet and Finetune ResNet is a feed-forward neural network with the same number of parameters as OccamNet.
\label{figure:implicitbenchmarks}
}
\end{figure}

Figure~\ref{figure:implicitbenchmarks}a and Table~\ref{table:implicitbenchmarks} show OccamNet's performance on implicit functions. OccamNet demonstrates an advantage on challenging implicit functions. Notably, Eureqa is unsuccessful in fitting $m_1v_1 - m_2v_2 = 0$ (conservation of momentum). 
Note that we only compare OccamNet to Eureqa for Implicit Functions because none of the other methods include the regularization that would be necessary to fit such functions.

Figure~\ref{figure:implicitbenchmarks}b and Table~\ref{table:implicitbenchmarks} demonstrate applications of OccamNet in image recognition, a domain that are not natural for standard symbolic regression baselines, but is somewhat more natural for OccamNet due to its interpretation as a feed-forward neural network.

We train OccamNet to classify MNIST~\cite{lecun-gradientbased-learning-applied-1998}\footnote{Creative Commons Attribution Share Alike 3.0 License} in a binary setting between the digits 0 and 7 (\emph{MNIST Binary}). For this high-dimensional task, we implement OccamNet on an Nvidia V100 GPU, yielding a sizable 8x speed increase compared to a CPU. For MNIST Binary, one of the successful functional fits that OccamNet finds is $
    y_0\left(\vec{x}\right) = \tanh{(10(\max(x_{25,15},x_{26,19})}+\tanh(x_{15,15})+2x_{25,10} +2x_{25,13}))$ and
    $y_1\left(\vec{x}\right) = \tanh\left(10\tanh(10\left(x_{18,8} + x_{20,6}\right))\right).$
The model learns to incorporate pixels into the functional fit that are indicative of the class: 
here $x_{18,8}$ and $x_{20,6}$ are indicative of the digit 7. 
These observations hold when we further benchmark the integration of OccamNet with deep feature extractors. We extract features from ImageNet~\cite{imagenet_cvpr09}\footnote{The Creative Commons Attribution (CC BY) License
} images using a ResNet 50 model, pre-trained on ImageNet~\cite{he2016deep}. For simplicity, we select two classes, 
``minivan'' and ``porcupine'' (\emph{ImageNet Binary}). 
OccamNet significantly improves its accuracy by backpropagating through our model using a standard cross-entropy signal. We either freeze the ResNet weights (\emph{Backprop OccamNet}) or finetune ResNet through OccamNet (\emph{Finetune ResNet}). In both cases, the converged OccamNet represents simple rules, ($y_0(\vec{x}) = x_{1838}$, $y_1(\vec{x}) = x_{1557}$), suggesting that replacing the head in deep neural networks with OccamNet might be promising.

\subsection{Real-world regression datasets} \label{sec:PMLBTests}
We also test OccamNet's ability to fit real-world datasets, selecting 15 datasets with 1667 or fewer datapoints from the Penn Machine Learning Benchmarks (PMLB\footnote{Creative Commons Attribution 4.0
International License}) regression datasets \cite{PMLB}. These are real-world datasets, and based on their names, we infer that many are from social science, suggesting that they are inherently noisy and likely to follow no known symbolic law. Additionally, 1/3 of the datasets we choose have feature sizes of 10 or greater. These factors make the PMLB datasets challenging symbolic regression tasks. We again compare OccamNet to Eplex and AI Feynman 2.0.\footnote{AIF's regression algorithm examines all possible feature subsets, the number of which grows exponentially with the number of features. Accordingly, we only test the datasets with ten or fewer features. AI Feynman 2.0 failed to run on a few datasets. All remaining datasets are included in tables and figures.}

We test OccamNet twice. For the first test, ``OccamNet-Small,'' we test exactly 1,000,000 functions, the same number as we test for Eplex. For the second test, ``OccamNet-GPU,'' we exploit our architecture's integration with the deep learning framework by running OccamNet on an Nvidia V100 GPU and testing a much larger number of functions. We allow AIF to run for approximately as long or longer than OccamNet for each dataset.

As discussed in the SM, we perform grid search on hyperparameters and identify the fits with the best training, validation, and testing Mean Squared Error (MSE) losses. The raw data from these experiments are shown in the SM.

\begin{figure*}
    \centering
    
    \includegraphics[width=\linewidth]{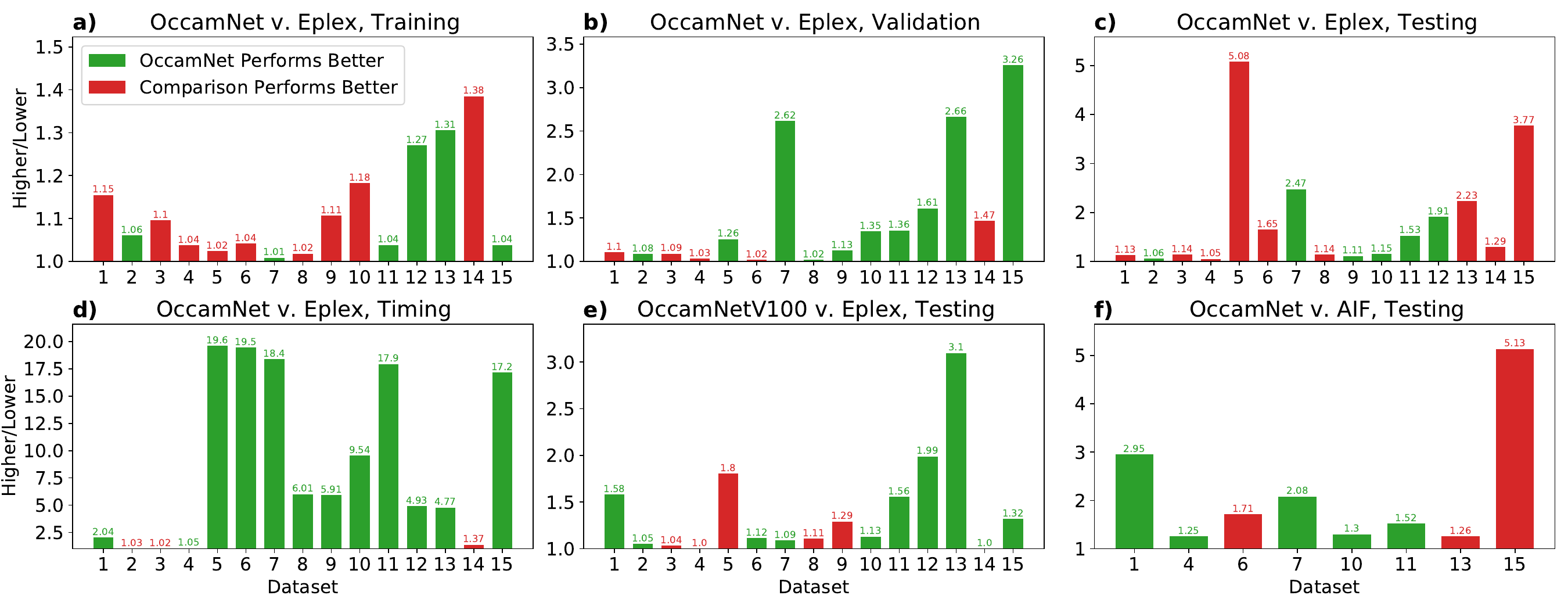}
    
    \caption{Bar charts showing the relative performance between OccamNet-CPU, OccamNet-GPU, and two baseline methods, Eplex and AIF. The x-axis is the dataset involved. The y-axis is the relative performance according to the given metric: the MSE on the training, validation, or testing set or the training time. To compute this relative performance, we divide the higher (worse) performance value by the lower (better) performance value for each dataset. The green bars represent datasets where OccamNet has a lower (better) performance value than the comparison baseline method, and the red bars represent the datasets where the comparison method has a better performance than OccamNet.}
    
    \label{fig:PMLBChart}
\end{figure*}

Figure \ref{fig:PMLBChart}
shows the relative performance of OccamNet-CPU, OccamNet-GPU, and baselines according to several metrics. As shown in Figure~\ref{fig:PMLBChart}a-c, overall, Eplex outperforms OccamNet-CPU in training and testing MSE loss, but OccamNet-CPU outperforms Eplex in validation loss. We speculate that OccamNet-CPU's performance drop between the validation and testing datasets being larger than Eplex's performance drop results from overfitting from the larger set of hyperparameter combinations used by OccamNet-CPU (details in the SM).

Additionally, OccamNet-CPU runs faster than Eplex in nearly all datasets tested, often by an order of magnitude (Figure \ref{fig:PMLBChart}d). Furthermore, OccamNet is highly parallel and can easily scale on a GPU. Thus, a major advantage of OccamNet is its speed and scalability (see Section \ref{sec:PMLBScaleTests} for a further discussion of OccamNet's scaling). Comparing OccamNet-GPU and Eplex demonstrates that OccamNet continues to improve when testing more functions. The testing MSE is where OccamNet-GPU performs worst in comparison to Eplex (see Figure \ref{fig:PMLBChart}e), but it still outperforms Eplex at 10 out of 15 of the datasets while running more than nine times faster on average. Thus OccamNet's speed and scalability can be exploited to greatly increase its accuracy at symbolic regression. This demonstrates that OccamNet is a powerful alternative to genetic algorithms for interpretable data modeling.

Additionally, OccamNet outperforms AIF for training, validation, and testing MSE, while running faster. OccamNet-CPU achieves a lower training and validation MSE than AIF for every dataset tested. For training loss, OccamNet-CPU performs better than AIF in 4 out of 7 datasets (Figure \ref{fig:PMLBChart}f). Additionally, unlike OccamNet, AIF performs polynomial fitting, giving it an additional advantage. However, the datasets we test are likely a worst-case for AIF; the datasets are small, have no known underlying formula, and we normalize the data prior to training, meaning that AIF will likely struggle not to overfit with its neural network and will also be unlikely to identify graph modularities.

\subsection{Scaling on real-world regression datasets} \label{sec:PMLBScaleTests}

\begin{figure}[t]
    \centering
    \begin{subfigure}{0.49\textwidth}
        \includegraphics[width=\textwidth]{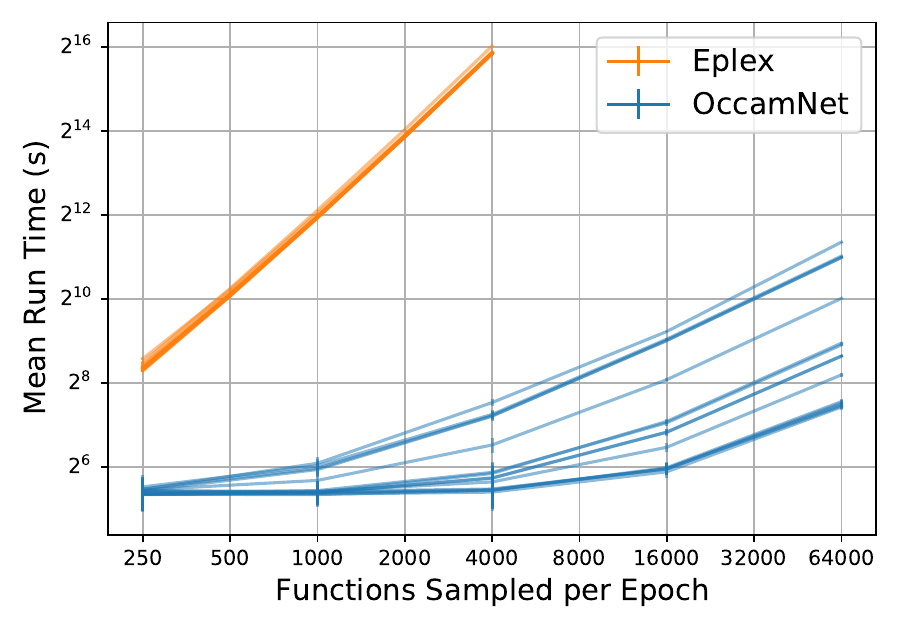}
    \end{subfigure}
    \begin{subfigure}{0.49\textwidth}
        \includegraphics[width=\textwidth]{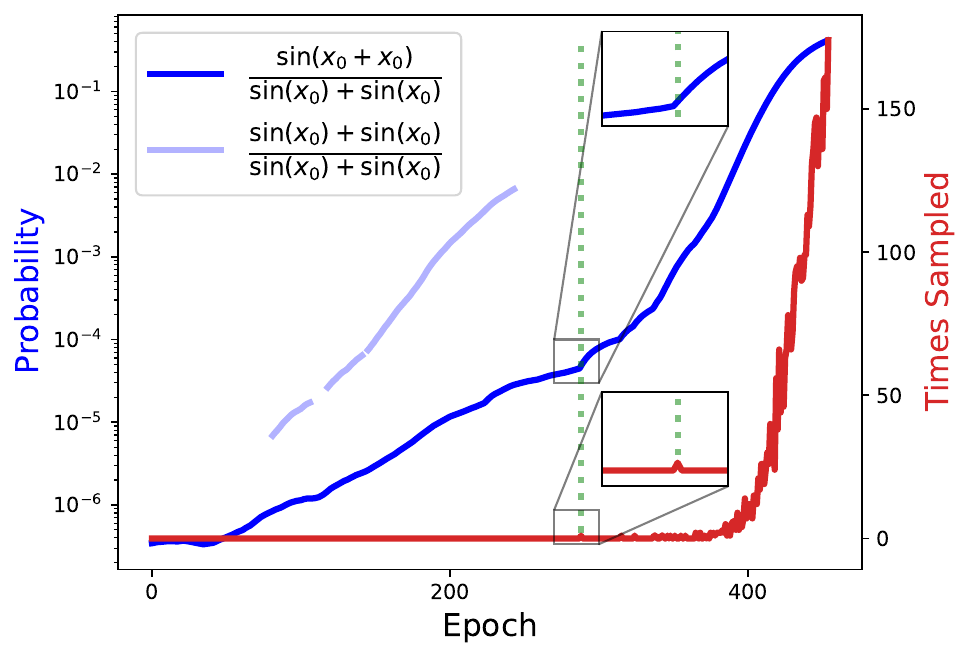}
    \end{subfigure}
    \caption{\textit{Left:} The run time for OccamNet-GPU or Eplex as a function of the number of functions sampled per epoch. Each curve represents one of the 15 datasets. \textit{Right:} Gradual modularity with training. Dark blue is the probability of the correct function. Light blue is the probability of a suboptimal fit with a high probability early in training. Red corresponds to the number of samples of the correct function. The insets zoom in on the curves around the epoch where the correct function is first sampled.}
    \label{fig:ScalingAndModularity}
\end{figure}

As discussed in Section \ref{sec:PMLBTests}, OccamNet-CPU runs far more quickly than Eplex on the same hardware, meaning that it can scale to testing far more functions per epoch than Eplex in the same runtime. To explore this advantage, we compare OccamNet running on an Nvidia V100 GPU (OccamNet-GPU) against Eplex while varying the number of functions sampled per epoch for each method. Since Eplex is not designed to scale on a GPU, we run Eplex on a CPU as before. We benchmark both methods on the same 15 PMLB datasets (see the methods section for more details).

We include and discuss the complete results of this experiment in Appendix \ref{sec:PMLBScaleAnalysis}. In this section, we highlight key results. Figure \ref{fig:ScalingAndModularity} shows that OccamNet-GPU is often more than an order of magnitude faster than Eplex. Eplex scales quadratically with the number of functions, whereas OccamNet's runtime asymptotes to linear growth. However, the V100 GPU's extreme parallelism initially suppresses OccamNet-GPU's linear time complexity, demonstrating an advantage of OccamNet's ability to scale on a GPU.

In all of the 15 datasets, OccamNet-GPU's training loss decreases with larger runtimes, demonstrating that OccamNet can utilize the greater number of sampled functions that its efficient scaling allows. 
Additionally, for 11 of the training datasets, the OccamNet-GPU best fit has a MSE that is lower than or equal to the Eplex best fit MSE. Interestingly, OccamNet-GPU's validation and testing loss do not always show such a clear trend of improvement with increasing sample size. Given that the training loss does improve, we suspect that this is a case of overfitting. OccamNet-GPU's validation loss does decrease with increasing number of functions sampled for most of the datasets.

\section{Discussion} \label{sec:discussion}
Since our experimental settings did not require very large depths, we have not tested the limits of OccamNet-GPU in terms of depth rigorously (preliminary results on increasing the depth for pattern recognition are in the SM). We expect increasing depth to yield significant complications as the search space grows exponentially. We recognize the need to create symbolic regression benchmarks that would require expressions that are large in depth. We believe that other contributions to symbolic regression would also benefit from such benchmarks. Another direction where OccamNet might be improved is low-level optimization that would make the method more efficient to train. For example, in our PMLB experiments, we estimate that OccamNet performs >8x as many computations as necessary. Eplex may also benefit from optimization. Finally, similarly to other symbolic regression methods, OccamNet requires a specified set of primitives to fit a dataset. While it is a notable advantage of OccamNet to have non-differentiable primitives, further work needs to be done to explore optimization at a meta level that identifies appropriate primitives for the datasets of interest without having them provided ahead of time.

OccamNet's learning procedure allows it to combine partial solutions into better results. For example in Figure \ref{fig:ScalingAndModularity}, the correct function's probability increases monotonically by more than 100 times \emph{before being sampled} because OccamNet samples similar approximate solutions.

OccamNet successfully fits many implicit functions that other neurosymbolic architectures struggle to fit because of the non-differentiable regularization terms required to avoid trivial solutions. Although Eureqa also fits many of these equations, we find that it sometimes requires the data to be ordered by some latent variable and struggles when the dataset is very small. This is likely because Eureqa numerically evaluates implicit derivatives from the dataset \cite{ImplicitFitting}, which can be noisy when the data is sparse. While \citet{ImplicitFitting} propose methods for analyzing unordered data, it is unclear whether these methods have been implemented in Eureqa. Thus, OccamNet seems to shine in its ability to fit unordered and small datasets described by implicit equations (e.g., momentum conservation in line 5 in Table~\ref{table:implicitbenchmarks}).

To our knowledge, a unique advantage of our method compared to other symbolic regression approaches is that OccamNet represents complete analytic expressions with a single forward pass. This allows sizable gains when using an AI accelerator, as demonstrated by our experiments on a V100 GPU (Figure~\ref{fig:PMLBChart}). Furthermore, because of this property, OccamNet can be easily integrated with components from the standard deep learning toolkit. For example, lines 9-10 in Table~\ref{table:implicitbenchmarks} demonstrate integrating OccamNet with other neural networks and optimizing both together, which is not possible with Eureqa. We also conjecture that such integration with autoregressive approaches such as DSR~\cite{petersen2021deep} might be challenging as the memory and latency would increase.

An advantage of OccamNet over transformer-based approaches to symbolic regression is that OccamNet can find fits to data regardless of the primitive functions it is given, whereas transformer-based models \cite{BiggioNeuroSymResScales} can only fit functions that contain a certain set of primitive functions chosen at pretraining time. Thus, although transformer-based approaches may outperform OccamNet for functions similar to their training distribution, OccamNet and other similar approaches are more flexible and broadly applicable than transformer-based models. As discussed above, this is the reason that we do not compare against transformer-based methods in our experiments.

\section*{Acknowledgements}
We would like to thank Isaac Chuang, Thiago Bergamaschi, Kristian Georgiev, Andrew Ma, Peter Lu, Evan Vogelbaum, Laura Zharmukhametova, Momchil Tomov, Rumen Hristov, Charlotte Loh, Ileana Rugina and Lay Jain for fruitful discussions.

Research was sponsored in part by the United States Air Force Research Laboratory and was accomplished under Cooperative Agreement Number FA8750-19-2-1000. The views and conclusions contained in this document are those of the authors and should not be interpreted as representing the official policies, either expressed or implied, of the United States Air Force or the U.S. Government. The U.S. Government is authorized to reproduce and distribute reprints for Government purposes notwithstanding any copyright notation herein.  

This material is based upon work supported in part by the U.S. Army Research Office through the Institute for Soldier Nanotechnologies at MIT, under Collaborative Agreement Number W911NF-18-2-0048.

Additionally, Owen Dugan would like to thank the United States Department of Defence for sponsoring his research and for allowing him to attend the Research Science Institute for free.

\section{Methods}

We divide our methods section into two parts. In Section \ref{sec:completeModel}, we provide a more detailed description of OccamNet, and in \ref{sec:fullSetup} we fully describe the setup for all of our experiments. 

\subsection{Complete Model Description} \label{sec:completeModel}
We divide this section as follows:\begin{enumerate}
    \item In Section \ref{sec:additional_exp_results}, we present additional materials that support the figures from the main text.
    \item In Section \ref{sec:sampling}, we describe of OccamNet's sampling process.
    \item In Section \ref{sec:probability}, we describe OccamNet's probability distribution.
    \item In Section \ref{sec:initialization}, we describe OccamNet's initialization process.
    \item In Section \ref{sec:functionSelection}, we describe OccamNet's function selection.
    \item In Section \ref{sec:loss}, we describe OccamNet's loss function.
    \item In Section \ref{sec:training2}, we describe OccamNet's outer training loop.
    \item In Section \ref{sec:constantFitting}, we describe OccamNet's two-step training method for fitting constants.
    \item In Section \ref{sec:recurrence2}, we describe OccamNet's handling of recurrence.
    \item In Section \ref{sec:regularization}, we describe OccamNet's regularization for fitting implicit functions.
    \item In Section \ref{sec:undefOutputs}, we describe OccamNet's procedure for handling functions with undefined outputs.
    \item In Section \ref{sec:units}, we describe OccamNet's method for regularizing to respect units.
\end{enumerate}

\subsubsection{Supporting Materials for the Main Figures}
\label{sec:additional_exp_results}

Tables \ref{table:analyticbenchmarks}, \ref{table:programbenchmarks}, and \ref{table:implicitbenchmarks} present our experiments in a tabular format.

\begin{table}[h]
\centering
\caption{Analytic Functions. The proportion of 10 trials that converge to the correct analytic function for OccamNet, Eureqa, Eplex, AI Feynman 2.0, and Deep Symbolic Regression (DSR). \emph{sec.} is the average number of seconds for convergence. Eureqa almost always finishes much more quickly than the other methods, so we do not provide training times for Eureqa.}
\label{table:analyticbenchmarks}
\begin{tabular}{rlcrccrcrcr}
\toprule
\multicolumn{11}{c}{Analytic Functions}\\
\cmidrule(r){1-11}
\multicolumn{1}{c}{\#} & \multicolumn{1}{c}{Targets} & \multicolumn{1}{c}{OccamNet} & \multicolumn{1}{c}{sec.} & \multicolumn{1}{c}{Eureqa} & \multicolumn{1}{c}{Eplex} & \multicolumn{1}{c}{sec.} &  \multicolumn{1}{c}{AI Feynman} & \multicolumn{1}{c}{sec.} & \multicolumn{1}{c}{DSR} & \multicolumn{1}{c}{sec.}\\
\cmidrule(r){1-1}
\cmidrule(r){2-2}
\cmidrule(r){3-3}
\cmidrule(r){4-4}
\cmidrule(r){5-5}
\cmidrule(r){6-6}
\cmidrule(r){7-7}
\cmidrule(r){8-8}
\cmidrule(r){9-9}
\cmidrule(r){10-10}
\cmidrule(r){11-11}
1 & $2x^2+3x$ & 1.0 & 5 & 1.0 & 1.0 & 16 & 1.0 & 35 & 1.0 & 3\\
2 & $\sin(3x+2)$ & 0.8 & 56 & 1.0 & 0.9 & 16 & 1.0 & 620 & 1.0 & 3\\
3 & $\sum_{n=1}^3\sin(nx)$ & 0.7 & 190 & 0.0 & 0.0 & 17 & 1.0 & 815 & 1.0 & 36\\
4 & $(x^2+x)/(x+2)$ & 0.9 & 81 & 0.7 & 0.5 & 44 & 1.0 & 807 & 1.0 & 2\\
5 & $x_0^2(x_0+1)/x_1^5$ & 0.3 & 305 & 1.0 & 0.9 & 53 & 0.0 & 1918 & 1.0 & 84\\
6 & $x_0^2/2+(x_1+1)^2/2$ & 0.6 & 83 & 0.7 & 0.2 & 92 & 1.0 & 3237 & 0.0 & 3935\\
\bottomrule
\end{tabular}
\end{table}

\begin{table}[h]
\centering
\caption{Non-analytic Functions. The proportion of 10 trials that converge to the correct function for OccamNet, Eureqa, and Eplex. \emph{sec.} is the average number of seconds for convergence. Eureqa almost always finishes much more quickly than OccamNet and Eplex, so we do not provide training times for Eureqa. ${}^*$For program \#6, Eplex fits $y_1$ every time and never fits $y_0$ correctly, so we give it a score of 0.5.}
\label{table:programbenchmarks}
\begin{tabular}{rlcrccr}
\toprule
\multicolumn{7}{c}{Non-analytic Functions} \\
\cmidrule(r){1-7}
\multicolumn{1}{c}{\#} & \multicolumn{1}{c}{Targets} & \multicolumn{1}{c}{OccamNet} & \multicolumn{1}{c}{sec.} & \multicolumn{1}{c}{Eureqa} & \multicolumn{1}{c}{Eplex} & \multicolumn{1}{c}{sec.}\\
\cmidrule(r){1-1}
\cmidrule(r){2-2}
\cmidrule(r){3-3}
\cmidrule(r){4-4}
\cmidrule(r){5-5}
\cmidrule(r){6-6}
\cmidrule(r){7-7}
1 & $3x$ if $x>0$, else $x$ & 0.7 & 26 & 1.0 & 0.0 & 52\\
2 & $x^2$ if $x>0$, else $-x$ & 1.0 & 10 & 1.0 & 0.0 & 46\\ 
3 & $x$ if $x>0$, else $\sin(x)$ & 1.0 & 236 & 1.0 & 0.0 & 47\\
4 & $\mathsf{SORT}(x_0,x_1,x_2)$ & 0.7 & 81 & 1.0 & 1.0 & 191\\
5 & $\mathsf{4LFSR}(x_0,x_1,x_2,x_3)$ & 1.0 & 14 & 1.0 & 1.0 & 262\\
\cmidrule(r){1-7}
\multirow{4}{*}{6} & $y_0(\vec{x})=x_1$ if $x_0<2$, & \multirow{4}{*}{0.3} & \multirow{4}{*}{157} & \multirow{4}{*}{0.1} & \multirow{4}{*}{${}^*$0.5} & \multirow{4}{*}{121}\\
 & \hspace{50pt}else $-x_1$\\
 & $y_1(\vec{x})=x_0$ if $x_1<0$,\\
 & \hspace{50pt}else $x_1^2$\\
\cmidrule(r){1-7}
\multirow{3}{*}{7} & $g(x)=x^2$ if $x<2$, & \multirow{3}{*}{1.0} & \multirow{3}{*}{64} & \multirow{3}{*}{0.0} & \multirow{3}{*}{0.0} & \multirow{4}{*}{189}\\
 & \hspace{50pt}else $x/2$\\
 & $y(x)=g^{\circ 4}(x)$ \\
\cmidrule(r){1-7}
\multirow{3}{*}{8} & $g(x)=x+2$ if $x<2$, & \multirow{3}{*}{1.0} & \multirow{3}{*}{64} & \multirow{3}{*}{0.6} & \multirow{3}{*}{1.0} & \multirow{4}{*}{116}\\
 & \hspace{50pt} else $x-1$\\
 & $y(x)=g^{\circ 2}(x)$\\
\bottomrule
\end{tabular}
\end{table}

\begin{table}[h]
\centering
\caption{Implicit Functions: The proportion of 10 trials that converge to the correct implicit function for OccamNet and Eureqa. Image Recognition: The best accuracy from 10 trials for both OccamNet and the baseline. The baseline above the mid-line is HeuristicLab~\cite{wagner2014}, and the baseline below the mid-line is a feed-forward neural network with the same number of parameters as OccamNet. \emph{sec.} is the average number of seconds for convergence. The baselines almost always finish much more quickly than OccamNet, so we do not provide baseline training times.}
\small
\label{table:implicitbenchmarks}
\begin{tabular}{rlcrcrlcrc}
\toprule
\multicolumn{5}{c}{Implicit Functions} & \multicolumn{5}{c}{Image Recognition}\\
\cmidrule(r){1-5}
\cmidrule(r){6-10}
\multicolumn{1}{c}{\#} & \multicolumn{1}{c}{Target} & \multicolumn{1}{c}{OccamNet} & \multicolumn{1}{c}{sec.} & \multicolumn{1}{c}{Eureqa} & \multicolumn{1}{c}{\#} & \multicolumn{1}{c}{Target} & \multicolumn{1}{c}{OccamNet} & \multicolumn{1}{c}{sec.} & \multicolumn{1}{c}{Baseline} \\
\cmidrule(r){1-1}
\cmidrule(r){2-2}
\cmidrule(r){3-3}
\cmidrule(r){4-4}
\cmidrule(r){5-5}
\cmidrule(r){6-6}
\cmidrule(r){7-7}
\cmidrule(r){8-8}
\cmidrule(r){9-9}
\cmidrule(r){10-10}
1 & $x_0x_1=1$&1.0&294          & 1.0 & 6 & MNIST Binary & 92.9 & 150 & 92.8\\
2 & $x_0^2+x_1^2=1$&1.0&153     & 0.6 & 7 & MNIST Trinary & 59.6 & 400 & 81.2\\
3 & $x_0/\cos(x_1)=1$&1.0&131   & 1.0 & 8 & ImageNet Binary & 70.7 & 400 & 78.0\\
\cmidrule{6-10}
4 & $x_1/x_0=1$&0.9&232        & 1.0 & 9 & Backprop OccamNet & 98.1 & 37 & 97.7\\
5 & $m_1v_1-m_2v_2$ = 0 & 1.0 & 270 & 0.0 & 10 & Finetune ResNet & 97.3 & 200 & 95.4\\
\bottomrule
\end{tabular}
\end{table}

\subsubsection{Sampling from OccamNet}
\label{sec:sampling}

In this section, we more carefully describe OccamNet's sampling process. As described in the main text, we start from a predefined collection of $N$ primitive functions $\mathbf{\Phi}=\{\phi_i(\cdot)\}_{i=1}^N$. Each neural network layer is defined by two sublayers, the \textit{arguments} and \textit{image} sublayers. For a network of depth $L$, each of these sublayers is reproduced $L$ times. Now let us introduce their corresponding hidden states: for $1\le l\le L$, each $l$'th arguments sublayer defines a hidden state vector $\widetilde{\mathbf{h}}^{(l)}$, and each $l$'th image sublayer defines a hidden state $\mathbf{h}^{(l)}$, as follows:
\begin{gather}
\mathbf{\widetilde h}^{(l)}=\left[\widetilde h^{(l)}_1,\dots, \widetilde h^{(l)}_M\right], \;\;
\mathbf{h}^{(l)}=\left[h^{(l)}_1,\dots,h^{(l)}_N\right],
\end{gather} where \begin{equation*}
    M = \sum_{0\leq k \leq N} \alpha(\phi_k)
\end{equation*} and $\alpha(\phi)$ is the arity of function $\phi(\cdot, \dots, \cdot).$ We also define $\mathbf{h}^{(0)}$ to be the input layer (an image sublayer) and $\mathbf{\widetilde h}^{(L+1)}$ to be the output layer (an arguments sublayer).
These image and arguments sublayer vectors are related through the primitive functions
\begin{equation}
\label{eq:argument-to-image}
h^{(l)}_i = \phi_i\left(\widetilde h_{j+1}^{(l)}, \dots, \widetilde h_{j + \alpha(\phi_i)}^{(l)}\right),\;\;
j = \sum_{0 \leq k < i} \alpha(\phi_k).
\end{equation} This formally expresses how the arguments connect to the images in any given layer, visualized as the bold edges between sublayers in Figure 1 in the main paper. To complete the architecture and connect the images from layer $l$ to the arguments of layer $(l+1)$, we sample from the softmax of the weights\footnote{we define for any $\mathbf{z}=[z_1,\dots,z_{N_l}]$ the softmax function as follows $
\mathsf{softmax}(\mathbf{z};T) \colonequals \left[\frac{\exp(z_1/T)}{\sum_{i=1}^{N_l}\exp(z_i/T)},\dots,\frac{\exp(z_{N_l}/T)}{\sum_{i=1}^{N_l}\exp(z_i/T)} \right]
$}: 
\begin{equation}
\label{eq:inductive-bias}
\widetilde{\mathbf{h}}^{(l+1)} = \begin{bmatrix}\widetilde h_1^{(l+1)} \\ \vdots \\ \widetilde h_{M_{l+1}}^{(l+1)}\end{bmatrix} \equiv \mathsf{SAMPLE}\left(\begin{bmatrix} {\mathsf{softmax}(\mathbf{w}_1^{(l)};T^{(l)})} \\ \vdots \\ {\mathsf{softmax}(\mathbf{w}_{M_{l+1}}^{(l)};T^{(l)}}\end{bmatrix}\right)\begin{bmatrix}h_1^{(l)} \\ \vdots \\ h_{N_l}^{(l)}\end{bmatrix}
\end{equation} where the $\mathsf{SAMPLE}$ function samples a one-hot row vector for each row based on the categorical probability distribution defined by $\mathsf{softmax}(\mathbf{w};T)^\top$. Here the hidden states $\vb{h}^{(l)}$ and $\widetilde{\vb{h}}^{(l+1)}$ have $N_l$ and $M_{l+1}$ coordinates, respectively, and the vectors $\mathbf{w}_{i}^{(l)}$ represent the $i$th row of the weights for the $l$th layer. In practice, we set $T^{(l)}$ to a fixed, typically small, number. The last layer is usually set to a higher temperature to allow more compositionality. These sampled edges are encoded as sparse matrices, through which a forward pass evaluates $\vec{f}$.

It is also possible to implement OccamNet without the sampling part of the propagation. In this case, the softmax of the weight matrices is treated as the weights of linear layers, and we minimize the MSE loss between the outputs and the desired outputs. In practice, however, we find that this approach leads to solutions which are less sparse, which makes this approach less interpretable and often converge to suboptimal local minima.

\begin{figure}[t]
    \centering
    \includegraphics[width=0.8\textwidth]{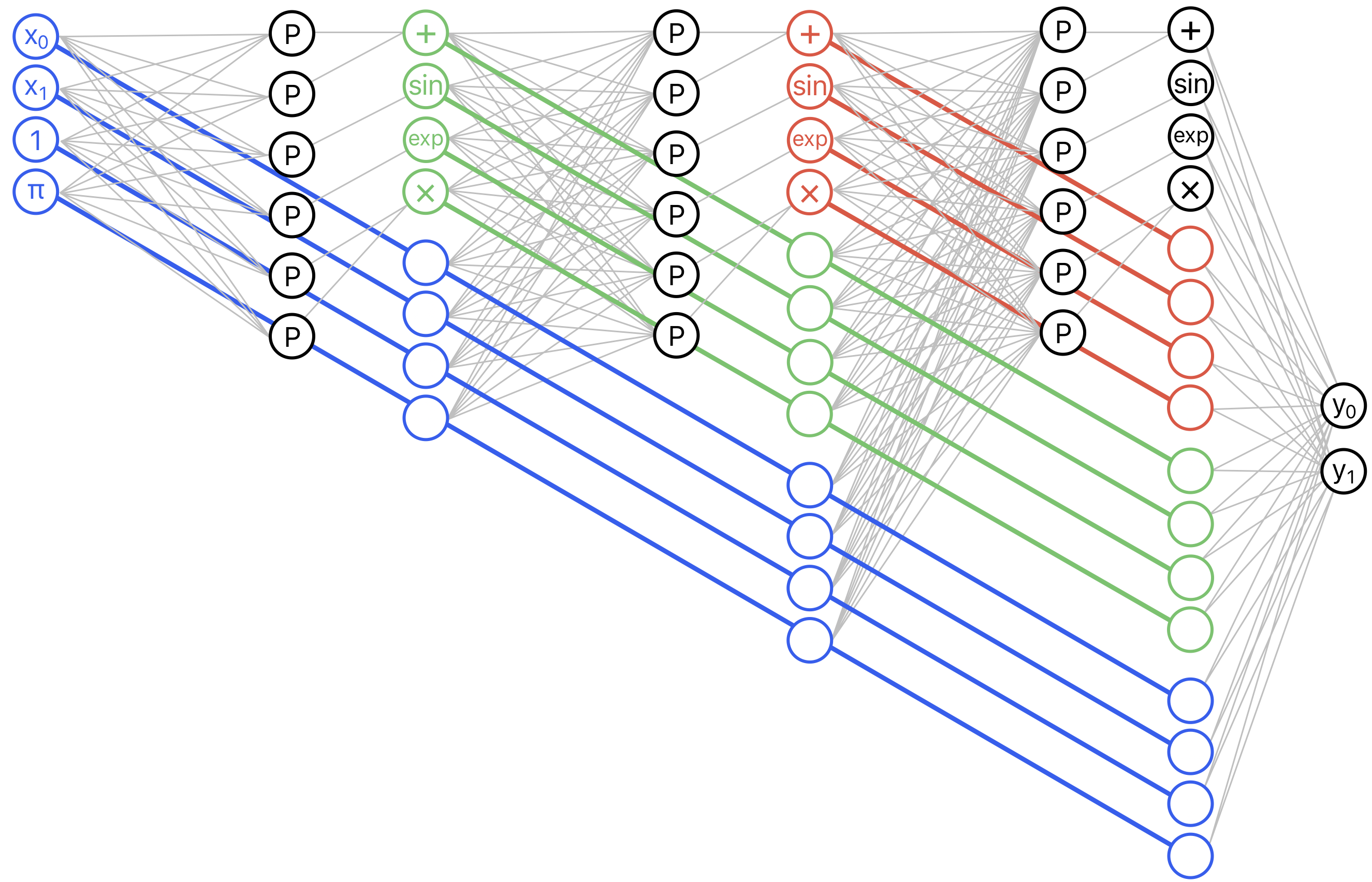}
    \caption{
    Skip connections.
    Nodes are color coded with lines indicating the origin of the reused neurons. 
    }
    \label{fig:mesh2}
\end{figure}

As shown in Figure \ref{fig:mesh2}, we use skip connections similar to those in DenseNet~\cite{huang2017densely} and ResNet~\cite{he2016deep}, concatenating each image layer with prior image layers. In particular, such a network now has argument layers computed as
\begin{equation}
\widetilde{\mathbf{h}}^{(l+1)} = \begin{bmatrix}\widetilde h_1^{(l+1)} \\ \vdots \\ \widetilde h_{M_{l+1}}^{(l+1)}\end{bmatrix} \equiv \mathsf{SAMPLE}\left(\begin{bmatrix} {\mathsf{softmax}(\mathbf{w}_1^{(l)};T^{(l)})} \\ \vdots \\ {\mathsf{softmax}(\mathbf{w}_{M_{l+1}}^{(l)};T^{(l)}}\end{bmatrix}\right)\mathsf{CONCAT}(\mathbf{h}^{(0)},\mathbf{h}^{(1)},\ldots,\mathbf{h}^{(l)}),
\end{equation}
where now each vector $\mathbf{w}_i^{(l+1)}$ has $\sum_{i=0}^l N_i$ components instead of $N_l$. Skip connections yield several desirable properties: (\emph{i}) The depth of equations is not fixed, lifting the requirement that the number of layers of the solution be known in advance. (\emph{ii}) The network can find compact solutions as it considers all levels of composition. This promotes solution sparsity and interpretability. (\emph{iii}) Primitives in shallow layers can be reused, analogous to feature reuse in DenseNet. (\emph{iv}) Subsequent layers may behave as higher-order corrections to the solutions found in early layers. Additionally, if we implement OccamNet without sampling, shallow layers are trained before or alongside the subsequent layers due to more direct supervision because gradients can propagate to shallow layers more easily to avoid exploding or vanishing gradients.

From Equation~\eqref{eq:argument-to-image}, we see that $M_{l+1} = M = \sum_{0\leq k \leq N} \alpha(\phi_k)$. If no skip connections are used, $N_l = N = |\mathbf{\Phi}|$. If skip connections are used, however, $N_l$ grows as $l$ increases. We demonstrate how the scaling grows as follows.
Let $u$ be the number of inputs and $v$ be the number of outputs. When learning connections from images to arguments at layer $l$ ($1 \leq l \leq L$), there will be skip connections from the images of the previous $l$ layers $0,1,\dots,l-1$. Hence the $i$th layer has an image size of $u+iN,$ as shown in Figure \ref{fig:mesh2}. We learn linear layers from these images to arguments, and the number of arguments is always $M$. Thus, in total, we have the following number of parameters: 
$$
v(u + (L+1)N) + M \sum_{i=0}^{L-1}(u + iN)\in O(NML^2).
$$

Note that in the above discussion we assume that $M$ remains constant. However, to be able to represent all functions up to a particular depth, we must repeat primitives in earlier layers, causing $M$ to grow exponentially. For small numbers of layers, this is not problematic. If a larger expression depth is required, one can avoid primitives and increase the number of layers beyond what is necessary. This makes additional copies of each primitive available for use without requiring an exponential growth in the layer size.

\subsubsection{OccamNet's Probability Distribution}
\label{sec:probability}
OccamNet parametrizes not only the probability of sampling a given function $\vec{f} = (f_{(0)},\ldots,f_{(v-1)})^\top$ but also the probability of sampling each $f_{(i)}$ independently of the other components of $\vec{f}.$ As discussed in the main text, the probability of the model sampling $f_{(i)}$ as its $i$th output, $q_i(f_{(i)}|\mathbf{W}),$ is the product of the probabilities of the edges of $f_{(i)}$'s DAG. Similarly, $q(\vec{f}|\mathbf{W}),$ the probability of the model sampling $\vec{f},$ is given by the product of $\vec{f}$'s edges, or $q(\vec{f}|\mathbf{W})=\prod_{i=0}^{v-1}q_i(f_{(i)}|\mathbf{W}).$ 

Because $q(\cdot|\cdot)$ is a probability distribution, we have $\sum_{\vec{f} \in \mathcal{F}_{\mathbf{\Phi}}^L}q(\vec{f}|\mathbf{W})=1$ and $q(\vec{f}|\mathbf{W})\geq 0$ for all $\vec{f}$ in $\mathcal{F}_{\mathbf{\Phi}}^L$. Similar results hold for the probability distributions of each component $f_{(i)}$.

\begin{figure*}
    \centering
    
    \includegraphics[width=0.6\linewidth]{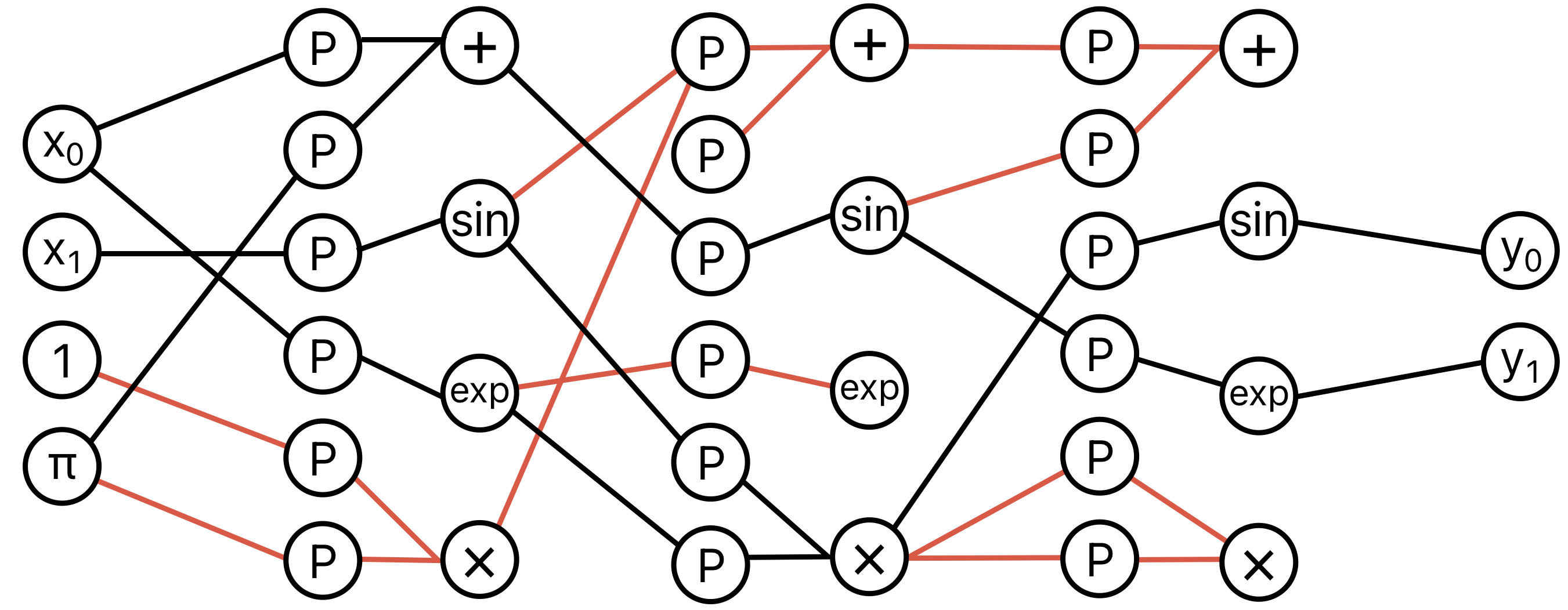}
    
    \caption{A demonstration of the dropped connections from sampled paths in OccamNet. All red paths are dropped from the final symbolic form of the sampled function because they are not directly connected to the outputs. These paths are unnecessarily computed during OccamNet's training process, leading to potential slowdowns in training.}
    
    \label{fig:DroppedConnections}
\end{figure*}

OccamNet's sampling process involves independently sampling connections from each layer. Although each of OccamNet's layers represents an independent probability distribution, when sampling a function, the layers do not act independently. This is because the samples from layers closer to the outputs inform which of the sampled connections from previous layers are used. In particular, the full DAG that OccamNet samples has many disconnected components, and all components of the DAG which are not connected to any of the output nodes are effectively trimmed (See Figure \ref{fig:DroppedConnections}). This is advantageous as it allows OccamNet to produce very different distributions of functions for different choices of connections in the final few layers, thereby allowing OccamNet to explore multiple classes of functions simultaneously.

As discussed in the main text, $q(\vec{f}|\mathbf{W})$ is the product of the probabilities of the sampled connections in $\vec{f}$'s DAG which are connected to the output nodes. However, in practice, we compute probabilities of functions in a feed-forward manner. This computation underestimates some probabilities; it actually computes an estimate $q_{apx}(\vec{f}|\mathbf{W})$ of $q(\vec{f}|\mathbf{W}).$

To compute the probability of a given function, we assign each image and argument node a probability given this function's DAG. We denote the probability of the $i$'th node of the $l$'th image layer with $p^{(l)}_i$ and the probability of the $i$'th node of the $l$'th argument layer with $\tilde{p}^{(l)}_i.$

We propagate probabilities as follows. If the $i$'th image node in layer $l$ is connected to the $j$'th argument node in layer $l+1$, the probability of the $j$'th argument node in layer $l$ is 
\begin{equation}
    \tilde{p}^{(l+1)}_j = p^{(l)}_i\cdot p^{(l,j)}_i(T^{(l)}).
\end{equation} 
The $i$th image node of the $l$th layer then has probability given by 
\begin{equation}
p_i^{(l)} = \prod_{k=n+1}^{n+\alpha(\phi_i)}\widetilde p_k^{(l)}, \hspace{10pt} n = \sum_{j=1}^{i-1} \alpha(\phi_j),
\end{equation} 
where $\alpha(f)$ denotes the number of inputs to $f$. Finally, to calculate the probability of a function, we multiply the probabilities of the output nodes.

This algorithm computes function probabilities correctly unless a function's DAG has multiple nodes connecting to the same earlier node in the DAG. In this case, the probability of the earlier node is included multiple times in the final function probability, producing an estimate that is below the true probability of sampling the function.

In practice, we find that this biased evaluation of probabilities does not substantially affect OccamNet training. Note that when we equalize all functions to have the same probability (Section \ref{sec:initialization}) or sample the highest probability function (Section \ref{sec:functionSelection}), we do so with respect to the probability estimate $q_{apx}$, not with respect to $q.$ In this paper, we use $q$ to mean $q_{apx}$ unless otherwise specified.

\subsubsection{Initialization}
\label{sec:initialization}

When beginning this project, we originally initialized all model weights to 0. However, this initializes complex functions, which have DAGs with many more edges than simple functions, to low probabilities. As a result, we found in practice that the network sometimes struggled to converge to complex functions with high fitness $K(\mathcal{M},f)$ because their initial low probabilities meant that they were sampled far less often than simple functions. This is because even if complex functions have a higher probability increase than simple functions when they are sampled, the initial low probabilities caused the complex functions to be sampled far less and to have an overall lower expected probability increase.

To address this issue, we now use a second initialization algorithm, which initializes all functions to equal probability. This initialization algorithm iterates through the layers of the network. In practice, to balance effects discussed at the end of this section, we initialize to weights interpolated between 0 and the algorithm discussed below. More details are given at the end of this section.

The algorithm to initialize all functions with equal probability establishes as an invariant that, after assigning the weights up to the $l$th layer, all paths leading to a given node in the $l$th argument layer have equal probabilities. Then, each argument layer node has a unique corresponding probability, the probability of all paths up to that node. We denote the probability of the $i$th node in the $l$th argument sublayer as $\widetilde p_i^{(l)},$ because it is the probability of \textit{any} path leading to the $i$th node in the $l$th argument sublayer.
Because each argument layer node has a corresponding probability, each image layer node must also have a unique corresponding probability, which, for the $i$th node in the $l$th image sublayer, we denote as $p_i^{(l)}.$ Again, we use the notation $p_i^{(l)}$ because this is the probability of \textit{any} path leading to the $i$th node in the $l$th image sublayer. These image layer probabilities are given by\begin{equation}
\label{eq:argtoimgApp}
p_i^{(l)} = \prod_{k=n+1}^{n+\alpha(\phi_i)}\widetilde p_k^{(l)}, \hspace{10pt} n = \sum_{j=1}^{i-1} \alpha(\phi_j).
\end{equation}

Our algorithm starts with input layer, or the 0th image layer. Paths to any node in the input layer have no edges so they all have probability 1. Thus, we initialize $p_i^{(0)}=1$ for all $i$. As the algorithm iterates through all subsequent $T$-Softmax layers, the invariant established above provides a system of linear equations involving the desired connection probabilities, which the algorithm solves. The algorithm groups the previous image layer according to the node probabilities, obtaining a set of ordered pairs $\{({p'}_a^{(l)},n_a^{(l)})\}_{i=a}^k$ representing $n_a^{(l)}$ nodes with probability ${p'}_a^{(l)}$ in the $l$th layer. Note that if two image nodes have the same probability $p_i^{(l)} = p_j^{(l)}$, then  the edges between any argument node in the next layer and the two image nodes must have the same probability in order to satisfy the algorithm's invariant: $p_i^{(l,k)} = p_j^{(l,k)}$. Then, we define ${p'}_a^{(l,i)}$ as the probability of the edges between the image nodes with probability ${p'}_a^{(l)}$ and the $i$th argument $P$-node of the $l$th layer. The probabilities of the edges to a given $P$-node sum to 1, so for each $j$, we must have $\sum_{a} n_a^{(l)}{p'}_a^{(l,i)} =1.$ Further, the algorithm requires that the probability of a path to a $P$-node through a given connection is the same as the probability of a path to that $P$-node through any other connection. The probability of a path to the $i$th $P$-node through a connection with probability ${p'}_a^{(l,i)}$ is ${p'}_a^{(l)}{p'}_{a}^{(l,i)},$ so we obtain the equations ${p'}_0^{(l)}{p'}_{0}^{(l,i)}={p'}_a^{(l)}{p'}_{a}^{(l,i)},$ for all $a$ and $i$. These two constraints give the vector equation \begin{equation}
    \begin{bmatrix}
        n_0^{(l)} & n_1^{(l)} & n_2^{(l)} &\cdots & n_k^{(l)}\\
        {p'}_0^{(l)} & -{p'}_1^{(l)} & 0 & \cdots & 0\\
        {p'}_0^{(l)} & 0 & -{p'}_2^{(l)} & \cdots & 0\\
        \vdots&\vdots&\vdots&\ddots&\vdots\\
        {p'}_0^{(l)} & 0 & 0 & \cdots & -{p'}_k^{(l)}
    \end{bmatrix}\begin{bmatrix}
        {p'}_{0}^{(l,j)}\\
        {p'}_{1}^{(l,j)}\\
        {p'}_{2}^{(l,j)}\\
        \vdots\\
        {p'}_{k}^{(l,j)}
    \end{bmatrix} = \begin{bmatrix}
        1\\
        0\\
        0\\
        \vdots\\
        0
    \end{bmatrix},
    \label{eq:initialization_matrix}
\end{equation} for all $1\le j\le M$. The algorithm then solves for each ${p'}_{a}^{(l,j)}.$

After determining the desired probability of each connection of the $l$th layer, the algorithm computes the SPL weights ${\mathbf{w}'}^{(l,j)}$ that produce the probabilities ${p'}_{a}^{(l,j)}$. Since there are infinitely many possible weights that produce the correct probabilities, the algorithm sets ${w'}_0^{(l,j)}=0.$ Then, the algorithm uses the softmax definition of the edge probabilities to determine the required value of $\sum_{n=1}^k\exp({w'}_n^{(l,j)}/T^{(l)})$:\begin{align*}
{p'}_{0}^{(l,j)} &= \frac{\exp({w'}_0^{(l,j)}/T^{(l)})}{\sum_{n=1}^k\exp({w'}_n^{(l,j)}/T^{(l)})}\\
&= \frac{1}{\sum_{n=1}^k\exp({w'}_n^{(l,j)}/T^{(l)})} \end{align*} so \begin{equation*}
    \sum_{a=1}^k\exp({w'}_a^{(l,j)}/T)=1/{p'}_{0}^{(l,j)}.
\end{equation*} Substituting this equation into the expression for the other probabilities gives \begin{align*}
    {p'}_{a}^{(l,j)} &= \exp({w'}_a^{(l,j)}/T^{(l)})/\left(\sum_{n=1}^k\exp({w'}_n^{(l,j)}/T^{(l)})\right)\\
    &= {p'}_{0}^{(l,j)}\exp({w'}_a^{(l,j)}/T^{(l)}).
\end{align*} Solving for ${w'}_i^{(l,j)}$ gives \begin{equation}
    {w'}_a^{(l,j)} = T^{(l)}\log\left({p'}_{a}^{(l,j)}/{p'}_{0}^{(l,j)}\right),
    \label{eq: initialization_log}
\end{equation} which the algorithm uses to compute ${w'}_i^{(l,j)}$. 

After determining the weights ${w'}_a^{(l,j)}$ the algorithm assigns them to the corresponding $w_a^{(l,j)}.$ In particular, if the $i$th image node has probability ${p'}_a^{(l)},$ the weights of edges to the $i$th node are given by $w_i^{(l,j)} = {w'}_k^{(l,j)},$ for all $j.$ The algorithm then determines the values of $\widetilde p_j^{(l+1)},$ given by $\widetilde p_j^{(l+1)} = p_i^{(l)}p_i^{(l,j)}.$ Finally, the algorithm determines $p_i^{(l+1)}$ using Equation \ref{eq:argtoimgApp} and repeats the above process for subsequent layers until it reaches the end of the network.

In summary, the algorithm involves the following steps:
\begin{enumerate}
    \item Set $l=0$.
    \item Set $p_i^{(l)}=1$.
    \item Increment $l$ by 1.
    \item Compute $\{({p'}_a^{(l)},n_a^{(l)})\}_{i=a}^k$ and use Equation \ref{eq:initialization_matrix} to compute ${p'}_{a}^{(l,j)}.$
    \item Set ${w'}_a^{(l,j)}$ according to Equation \ref{eq: initialization_log}.
    \item If $l < L+1$, Compute $\widetilde p_i^{(l+1)}$ and $p_i^{(l+1)}$.
    \item Return to step 3 until $l = L+1.$
\end{enumerate}

This algorithm efficiently equalizes the probabilities of all functions in the network. In practice, however, we find that perfect equalization of functions causes activation functions with two inputs to be highly explored. This is because there are many more possible functions containing activation functions with two inputs than with one input. Additionally, as mentioned in Section \ref{sec:probability}, in this section we have implicitly been using the approximate probability $\tilde{q}.$ This probability underestimates many functions that include activation functions with two or more inputs because these functions are those which can use a node multiple times in their DAG. As a result, although all functions will have an equal $\tilde{q}$, some functions with multiple inputs will have larger $q$ than other functions, and $q$ is what determines the probability of being sampled. In practice, therefore, we find that a balance between initializing all weights to one and initializing all functions to equal probability is most effective for exploring all types of functions.

To implement this balance, we create an equalization hyperparameter, $E.$ If $E=0,$ we initialize all weights to 1 as in the original OccamNet architecture. If $E\neq 0,$ we use the algorithm presented above to initialize the weights and then divide all of the weights by $E$. For $E>1,$ this has the effect of initializing weights between the two initialization approaches. In practice, we find that values of $E=1$ and $E=5$ are most effective for exploring all types of functions (See Section \ref{sec:PMLBSetup}).

\subsubsection{Function Selection}
\label{sec:functionSelection}
As discussed in the main text, after training using a sampling strategy, the network selects the function $\vec{f}$ with the highest probability $q(\vec{f}|\mathbf{W}).$

We develop a dynamic programming algorithm that determines the DAG with the highest probability. The algorithm steps sequentially through each argument layer, and at each argument layer it determines the maximum probability path to each argument node. Knowing the maximum probability paths to the previous argument layer nodes allows the algorithm to easily determine the maximum probability paths to the next argument layer.

As with the network initialization algorithm, the function selection algorithm associates the $i$th $P$-node of the $l$th argument sublayer with a probability, $\widetilde p_i^{(l)},$ which represents the highest probability path to that node. Similarly, we let $p_i^{(l)}$ represent the assigned probability of the $i$th node of the $l$th image sublayer, defined as the highest probability path to a given image node. $p_i^{(l)}$ can once again be determined from $\widetilde p_i^{(l)}$ using Equation \ref{eq:argtoimgApp}. Further, the algorithm associates each node with a function, $\widetilde f_i^{(l)}$ for argument nodes and $f_i^{(l)}$ for image nodes, which represents the highest probability function to the corresponding node. Thus, $\widetilde f_i^{(l)}$ has probability $\widetilde p_i^{(l)},$ and $f_i^{(l)}$ has probability $p_i^{(l)}.$ Further, $f_i^{(l)}$ is determined from $\widetilde f_i^{(l)}$ using\begin{equation}
    \label{eq:argtoimgfuncApp}
    f_i^{(l)}(\vec{x}) = \phi_i\left(\widetilde f_{n+1}^{(l)}(\vec{x}),\ldots,\widetilde f_{n+\alpha(\phi_j)}^{(l)}(\vec{x})\right),
    \hspace{10pt} n = \sum_{j=1}^{i-1} \alpha(\phi_j).
\end{equation}

The algorithm iterates through the networks layers. At the $l$th layer, it determines the maximum probability path to each argument node, computing \begin{align*}
    \widetilde p_i^{(l+1)} &= \text{MAX}\left(p_0^{(l)}p_0^{(l,i)},\ldots,p_N^{(l)}p_N^{(l,i)}\right)\\
    \widetilde f_i^{(l+1)} &= \begin{cases}
                                   f_0^{(l)} & \text{if } \widetilde p_i^{(l+1)} = p_0^{(l)}p_0^{(l,i)}\\
                                   f_1^{(l)} & \text{if } \widetilde p_i^{(l+1)} = p_1^{(l)}p_1^{(l,i)}\\
                                   \;\;\vdots & \hspace{39pt}\vdots\\
                                   f_N^{(l)} & \text{if } \widetilde p_i^{(l+1)} = p_N^{(l)}p_N^{(l,i)}
                              \end{cases}.
\end{align*} Next, it determines the maximum probability path up to each image node, computing $p_i^{(l+1)}$ and $f_i^{(l+1)}$ using Equations \ref{eq:argtoimgApp} and \ref{eq:argtoimgfuncApp}, respectively. The algorithm repeats this process until it reaches the output layer, at which point it returns $\vec{f}_\text{max} = [\widetilde f_1^{(L)},\ldots,\widetilde f_N^{(L)}]^\top$ and $p_\text{max} = \prod_{i=1}^N \widetilde p_i^{(L)}.$

An advantage of this process is that identifying the highest probability function has the same computational complexity as sampling functions. In particular, the complexity at each layer is $O(MN_i),$ leading to an overall complexity of $O(NML^2)$ if skip connections are included.

\subsubsection{Loss Function and its Gradient} \label{sec:loss}
We train our network on mini-batches of data to provide flexibility for devices with various memory constraints. Consider a mini-batch $\mathcal{M} = ( X, Y )$, and a sampled function from the network $\vec{f}(\cdot) \sim q(\cdot|\mathbf{W})$. We compute the \textit{fitness} of each $f_{(i)}(\cdot)$ with respect to a training pair $( \vec{x}, \vec{y} )$ by evaluating the likelihood $$k_i\left(f_{(i)}(\vec{x}),\vec{y}\right)=(2\pi\sigma^2)^{-1/2}\exp(-\left[ f_{(i)}(\vec{x})-(\vec{y})_i\right]^2/(2\sigma^2)),$$ which is a Normal distribution with mean $(\vec{y})_i$ and variance $\sigma^2$, and measures how close $f_{(i)}(\vec{x})$ is to the target $(\vec{y})_i$. The fitness can be also viewed as a Bayesian posterior with a noninformative prior. The total fitness is determined by summing over the entire mini-batch: $K_i\left(\mathcal{M},f_{(i)}\right) = \sum_{(\vec{x},\vec{y}) \in \mathcal{M}} k_i\left(f_{(i)}(\vec{x}),\vec{y}\right)$.

The variance $\sigma^2$ of $k_i\left(f_{(i)}(\vec{x}),\vec{y}\right)$ characterizes the fitness function's smoothness. 
As $\sigma^2 \rightarrow 0$, the fitness is a delta function with nonzero fitness 
for some $( \vec{x}, \vec{y} )$ only if $f_{(i)}(\vec{x}) = (\vec{y})_i$. Similarly, a large variance characterizes a fitness in which potentially many solutions provide accurate approximations, increasing the risk of convergence to local minima. In the former case, learning becomes harder as few $f_{(i)}(\cdot)$ out of exponentially many sampleable functions result in any signal, whereas in the latter case learning might not converge to the optimal solution. We let $\sigma^2$ be a network hyperparameter, tuned for the tradeoff between ease of learning and solution optimality for different tasks.

Similar to \citet{petersen2021deep}, we use a loss function for backpropagating on the weights of $q(\cdot|\mathbf{W})$:
\begin{equation}
\label{eq:cross_entropy2}
\small
H_{q_i}[f_{(i)}, \mathbf{W}, \mathcal{M}] = - K_i\left(\mathcal{M},f_{(i)}\right)\cdot\log \left[q_i(f_{(i)}|\mathbf{W})\right].
\end{equation}
We can interpret~\eqref{eq:cross_entropy2} as the cross-entropy of the posterior for the target and the probability of the sampled function $f_{(i)}.$  If the sampled function $f_{(i)}$ is close to $f_{(i)}^*$, then $K_i(\mathcal{M},f_{(i)})$ will be large, and the gradient update below will also be large:
\begin{equation}
\label{eq:gradient_update}
\small
\nabla_{\mathbf{W}} H_{q_i}\left[f_{(i)}, \mathbf{W},\mathcal{M}\right]=-\frac{\nabla_{\mathbf{W}}q_i(f_{(i)}|\mathbf{W})}{q_i(f_{(i)}|\mathbf{W})} K_i\left(\mathcal{M},f_{(i)}\right).
\end{equation}

The first term on the right-hand side (RHS) of update~\eqref{eq:gradient_update} increases the probability of the function $f_{(i)}.$ The second term on the RHS is maximal when $f_{(i)}(\vec{x}) = f_{(i)}^*(\vec{x})$. Importantly, the second term approaches zero as $f_{(i)}$ deviates from $f_{(i)}^*$. If the sampled function is far from the target, then the probability update is suppressed by $K_i(\mathcal{M},f_{(i)})$. Therefore, we only optimize the probability for functions close to the target. 
Note that in~\eqref{eq:gradient_update} we backpropagate only through the probability of the function $f_{(i)}$ given by $q_i\left(f_{(i)}|\mathbf{W}\right),$ whose value \emph{does not} depend on the primitives in $\mathbf{\Phi},$ implying that the primitives can be non-differentiable. This is particularly useful for applications requiring non-differentiable primitive functions.
Furthermore, this loss function allows non-differentiable regularization terms, which greatly expands the regularization possibilities.

\subsubsection{Sample-based Training}
\label{sec:training2}
We use a sampling-based strategy to update our model, explained below without constant fitting for simplicity. This training procedure was first proposed in Risk-Seeking Policy Gradients \cite{petersen2021deep}. We denote $\mathbf{W}^{(t)}$ as the set of weights at training step $t,$ and we fix two hyperparameters: $R$, the number of functions to sample at each training step, and $\lambda$, or the \textit{truncation parameter}, which defines the number of the $R$ paths chosen for optimization via~\eqref{eq:gradient_update}. We initialize $\mathbf{W}^{(0)}$ as described in Section \ref{sec:connectivity}. We then proceed as follows:

\begin{enumerate}
\item Sample $R$ functions $\vec{f}_1,\dots, \vec{f}_R \sim q(\cdot|\mathbf{W}^{(t)}).$ We denote the $j$th output of $\vec{f}_i$ as $f_{i(j)}.$
\item For each output $j$, sort $f_{i(j)}$ from greatest to least value of $K_j\left(\mathcal{M},f_{i(j)}\right)$ and select the top $\lambda$ functions, yielding a total of $v \lambda$ selected functions $g_{1,j},\ldots,g_{\lambda,j}$. The total loss is then given by
$\sum_{i=1}^\lambda\sum_{j=0}^{v-1} H_{q_j}[g_{i,j},\mathbf{W},\mathcal{M}],$ which yields the training step gradient update:
\begin{equation}
\label{eq:total_update}
-\sum_{i=1}^\lambda\sum_{j=0}^{v-1} \frac{\nabla_{\mathbf{W}}q_j(g_{i,j}|\mathbf{W})}{q_j(g_{i,j}|\mathbf{W})} \; K_j(\mathcal{M},g_{i,j}).
\end{equation}
Notice that through~\eqref{eq:total_update} we have arrived at a modified REINFORCE update~\cite{10.1007/BF00992696}, where the policy is $q_i(\cdot|\cdot)$ and the regret is the fitness $K_i(\cdot,\cdot).$
\item Perform the gradient step~\eqref{eq:total_update} on $\mathbf{W}^{(t)}$ for all selected paths to obtain $\mathbf{W}^{(t+1)}.$ In practice, we find that the Adam algorithm~\cite{Kingma2015AdamAM} works well for this step.
\item Set $t = t + 1$ and repeat from Step 1 until a stop criterion is met. 
\end{enumerate}
Note that Equations~\eqref{eq:gradient_update} and~\eqref{eq:total_update} represent different objective functions -- we use~\eqref{eq:total_update}. The benefit of using Equation~\eqref{eq:total_update} is that accumulating over the top $v\lambda$ best fits to the target allows for explorations of function compositions that contain desired components but are not fully developed.
In practice, we find that reweighting the importance of the top-$v\lambda$ routes, substituting $K_j'(\mathcal{M},g_{i,j}) = K_j(\mathcal{M},g_{i,j}) / i,$ improves convergence speed by biasing updates towards the best routes.

\subsubsection{Constant Fitting} \label{sec:constantFitting}
Thus far, our method works for functions with constants known \textit{a priori}. Examples of such functions include $x^2$ or $x+\pi$ if $(\cdot)^2$ and $\pi$ are provided respectively as primitives and constants ahead of time. In some cases, however, we may wish to fit functions that involve constants that are not known \textit{a priori}. To fit such undetermined constants, we use activation functions with unspecified constants, such as $x^c$ and $c\cdot x$ ($c$ is undefined). We then combine the training process described in Section \ref{sec:training2} with a constant fitting training process. 

The two-step training process works as follows: We first sample a batch $\mathcal{M}$ and a function batch $(\vec{f}_1,\ldots,\vec{f}_R).$ Next, for each function $\vec{f}_i$, we fit the unspecified constants to $\mathcal{M}$ in $\vec{f}_i$ using gradient descent. Any other constant optimization method would also work. Finally, we update the network weights according to Section \ref{sec:training2}, using the fitness $K$ of the constant-fitted function batch. To increase training speed, we store each function's fitted constants for reuse.

\subsubsection{Recurrence}  \label{sec:recurrence2}
OccamNet can also be trained to find recurrence relations, which is of particular interest for programs that rely on $\mathsf{FOR}$ or $\mathsf{WHILE}$ loops. 
To find such recurrence relations, we assume a \emph{maximal} recursion $D$. We use the following notation for recurring functions: $f^{\circ (n+1)}(x) \equiv f^{\circ n}(f(x))$, with base case $f^{\circ 1}(x) \equiv f(x)$. 

To augment the training algorithm, we first sample $(\vec{f}_1,\dots, \vec{f}_R) \sim q(\cdot|\mathbf{W}^{(t)})$. For each $\vec{f}_i$, we compute its recurrence to depth $D$ as follows $\left(\vec{f}_i^{\circ 1}, \vec{f}_i^{\circ 2}, \dots, \vec{f}_i^{\circ D}\right)$, obtaining a collection of $RD$ functions. Training then continues as usual; we compute the corresponding $K_j(\mathcal{M},\vec{f}_{i(j)}^{\circ n})$, select the best $v\lambda$, and update the weights. It is important to note that we consider all depths up to $D$ since our maximal recurrence depth might be larger than the one for the target function. 

Note that we do not change the network architecture to accommodate for recurrence depth $D > 1$. As described in the main text, we can efficiently use the network architecture to evaluate a sampled function $\vec{f}(\vec x)$ for a given batch of $\vec x$. To incorporate recurrence, we take the output of this forward pass and feed it again to the network $D$ times, similar to what is typically done for recurrent neural networks. The resulting outputs are evaluations $\left(\vec{f}_i^{\circ 1}(\vec x), \vec{f}_i^{\circ 2}(\vec x), \dots, \vec{f}_i^{\circ D}(\vec x)\right)$ for a given batch of $\vec x.$

\subsubsection{Regularization}
\label{sec:regularization}
As discussed in the main text, to improve implicit function fitting, we implement a regularized loss function, \begin{equation*}
    K_i'(\mathcal{M},f)=K_i(\mathcal{M},f)-s\cdot r[f],
\end{equation*} for some regularization function $r$, where $s= n(\mathcal{M})/\sqrt{2\pi\sigma^2}$ is the maximum possible value of $K_i(\mathcal{M},f).$ We define\begin{equation*}
    r[f] = w_{\phi}\cdot \phi{[f]}+w_{\psi}\cdot \psi{[f]}+w_{\xi}\cdot \xi{[f]}+w_{\gamma}\cdot\gamma{[f]},
\end{equation*}where $\phi[f]$ measures trivial operations, $\psi[f]$ measures trivial approximations, $\xi[f]$ measures the number of constants in $f$, $\gamma[f]$ measures the number of activation functions in $f,$ and $w_{\phi},$ $w_{\psi},$ $w_{\xi},$ and $w_{\gamma},$ are weights for their respective regularization terms. We now discuss each of these regularization terms in more detail.

\textit{The $\phi[f]$ Regularization Term}:
The $\phi[f]$ term measures whether the unsimplified form of $f$ contains trivial operations, by which we mean operations that simplify to 0, 1, or the identity. For example, division is a trivial operation in $x/x,$ because the expression simplifies to $1.$ Similarly, $1\cdot x,$ $x^1,$ and $x^0$ are all trivial operations. We punish these trivial operations because they produce constant outputs without adding meaning to an expression.

To detect trivial operations, we employ two procedures. The first uses the \texttt{SymPy} package \cite{SymPy} to simplify $f$. If the simplified expression is different from the original expression, then there are trivial operations in $f$, and this procedure returns 1. Otherwise the first procedure returns 0. Unfortunately, the \texttt{SymPy} \texttt{==} function to test if functions are equal often incorrectly indicates that nontrivial functions are trivial. For example, \texttt{SymPy}'s \texttt{simplify} function, which we use to test if a function can be simplified, converts $x+x$ to $2\cdot x,$ and the \texttt{==} function states that $x+x\neq 2\cdot x$. To combat this, we develop a new function, \texttt{sympyEquals} which corrects for these issues with \texttt{==}. The \texttt{sympyEquals} is equivalent to \texttt{==}, except that it does not take the order of terms into account, and it does not mark expressions such as $x+x$ and $x\cdot x$ as unsimplified. We find that this greatly improves implicit function fitting.

The constant fitting procedure often produces functions that only differ from a trivial operation because of imperfect constant fitting, such as $f(x_0) = x_0^{0.0001},$ which is likely meant to represent $x_0^0.$ \texttt{SymPy}, however, will not mark this function as trivial. The second procedure addresses this issue by counting the constant activations, such as $x_0^{0.0001},$ $1.001\cdot x_0,$ and $x_0+0.001,$ which approximate trivial operations. For the activation function $f(x) = x+c,$ if the fitted $c$ satisfies $-0.1<c<0.1,$ the procedure adds 1 to its counter. Similarly, for the activation functions $f(x) = cx$ and $f(x) = x^c,$ if the fitted $c$ satisfies $-0.1<c<0.1$ or $0.9<c<1.1,$ the procedure adds 1 to its counter. We select these ranges to capture instances of imperfect constant fitting without labeling legitimate solutions as trivial. After checking all activation functions used, the procedure returns the counter. 

The $\phi[f]$ term returns the sum of the outputs of the first and second procedures. We find that a weight of $w_\phi\approx 0.7$ for $\phi[f]$ is most effective in our loss function. This value of $w_\phi$ ensures that most trivial $f$ have $K_i(\mathcal{M},f)-s\cdot w_{\phi}\cdot \phi[f]<0$, thus actively reducing the weights corresponding to functions with trivial operations, without over punishing functions and hindering learning.

\textit{The $\psi[f]$ Regularization Term}:
When punishing trivial operations using the $\phi$ term, we find that the network discovers many nontrivial operations which very closely approximate trivial operations by exploiting portions of functions with near-zero derivatives, which can be used to artificially compress data. For example, $\cos(x/2)$ closely approximates 1 if $-1<x<1.$ Unfortunately, it is often difficult to determine if a function approximates a trivial function simply from its symbolic representation. This issue is also identified in \cite{ImplicitFitting}.

To detect these trivial function approximations, we develop an approach that analyzes the activation functions' outputs \textit{during} the forward pass. The $\psi[f]$ term counts the number of activation functions which, during a forward pass, the network identifies as possibly approximating trivial solutions, as well as a metric for how close to trivial these functions are. For each primitive function, the network stores values around which outputs of that function often cluster artificially. Table \ref{tab:trivialApprox} lists the primitives which the network tests for clustering. 

The procedure for determining $\psi$ is as follows. The algorithm begins with a counter of 0. During the forward pass, if the network reaches a primitive function $\phi$ listed in Table \ref{tab:trivialApprox}, the algorithm tests each ordered tuple $(\phi,a,\delta)$ from Table \ref{tab:trivialApprox}, where $a$ is the point tested for clustering and $\delta$ is the cluster tolerance. If the mean of all the outputs of the primitive function, $\Bar{y},$ for a given batch satisfies $\left| \Bar{y}-a\right|<\delta$, the algorithm adds $\text{min}(5,0.1/\left|\Bar{y}-a\right|)$ to the counter. These expressions increase with the severity of clustered data; the more closely the outputs are clustered, the higher the punishment term. The minimum term ensures that $\psi[f]$ is never infinite. 

We also test for the approximation $\sin(x)\approx x$ by testing the inputs and outputs of the sine primitive function. If the inputs and outputs $x$ and $y$ of the sine primitive satisfy $\overline{\left|y-x\right|}<0.1,$ the algorithm adds $\text{min}(5,0.05/\overline{\left|y-x\right|})$ to the counter. In the future, we plan to consider more approximations similar to the small angle approximation.

\begin{table}[tb]
    \centering
    \caption{Primitive functions tested for clustering}
    \begin{tabular}{lcr}
         \toprule
         \multicolumn{1}{c}{Primitive Function} & \multicolumn{1}{c}{Cluster Points} & \multicolumn{1}{c}{Cluster Tolerance}\\
         \cmidrule(r){1-1}
         \cmidrule(r){2-2}
         \cmidrule(r){3-3}
         $(\cdot)^2$ & $\{0\}$ & 0.25\\
         $(\cdot)^3$ & $\{0\}$ & 0.25\\
         $\sin(\cdot)$ & $\{1,-1\}$ & 0.25\\
         $\cos(\cdot)$ & $\{1,-1\}$ & 0.25\\
         $(\cdot)^c$ & $\{1\}$ & 0.5\\
         \bottomrule
    \end{tabular}
    \label{tab:trivialApprox}
\end{table}

$\psi[f]$ should not artificially punish functions involving the primitives listed in Table \ref{tab:trivialApprox} that are not trivial approximations because no proper use of these primitive functions will always produce outputs very close to the clustering points. Because $\psi[f]$ flags functions based on their batch outputs, each batch will likely give different outcomes. This allows $\psi[f]$ to better discriminate between trivial function approximations and nontrivial operations: $\psi[f]$ should flag trivial function approximations often, but it should only flag nontrivial operations rarely when the inputs statistically fluctuate to produce clustered outputs. In practice, we find that a weight of $w_\psi\approx 0.3$ for $\psi[f]$ is most effective in our loss function.

\textit{The $\xi[f]$ Regularization Term}:
When our network converges to the correct solution, it may converge to a more complicated expression equivalent to the desired expression. To promote simpler expressions, we slightly punish functions based on their complexity. The $\xi[f]$ term counts the number of activation functions used to produce $f,$ which serves as a measure of $f$'s complexity. We find that a small weight of $w_\xi\approx 0.1$ for $\xi[f]$ is most effective in our loss function. This small value has little significance when distinguishing between a function that fits a dataset well and a function that does not, but it is enough to promote simpler functions over complex functions when they have approximately the same loss otherwise.

\textit{The $\gamma[f]$ Regularization Term}:
The $\gamma[f]$ term also punishes functions for their complexity. The $\gamma[f]$ term counts the number of constants in $f$, which, like the number of activation functions, serves as a metric for $f$'s complexity. We find that a weight of $w_\gamma\approx 0.15$ for $\gamma[f]$ is most effective in our loss function. Just as with $\xi[f],$ this small value has little significance when distinguishing between a function that fits a dataset well and a function that does not, but it is enough to slightly promote simpler functions over complex functions when they are otherwise equivalent. We weight $\gamma[f]$ slightly higher than $\xi[f]$ because many functions with constants can be simplified.

\subsubsection{Functions with Undefined Outputs}
\label{sec:undefOutputs}

One difficulty that may arise when training OccamNet is that many sampled functions are undefined on the input data range. Two cases of undefined functions are: 1) the function is undefined on part of the input data range for all values of a set of constants, or 2) the function is only undefined when the function's constants take on certain values. An example function satisfying case 1 is $f_1(x_0) = c_0/(x_0-x_0),$ which divides by $0$ regardless of the value of $c_0.$ An example function satisfying case 2 is $f_2(x_0) = x_0^{c_0},$ which is undefined whenever $x_0$ is negative and $c_0$ is not an integer.

In the first case, the network should abandon the function. In the second case, the network should try other values for the constants. However, the network cannot easily determine which case an undefined function satisfies. To balance both cases, if the network obtains an undefined result, such as \texttt{NaN} or \texttt{inf}, for the forward pass, the network tests up to 100 more randomized sets of constants. If none of these attempts produce defined results, the network returns the array of undefined outputs. For example, with $c_0/(x_0-x_0),$ the network tests a first set of constants, determines that they produce an undefined output, and tests 100 more constants. None of these functions are defined on all inputs, so the network returns the undefined outputs. 

In contrast, with $x_0^{c_0},$ the network might find that the first set of constants produces undefined outputs, but after 20 retries, the network might discover that $c_0 = 2$ produces a function defined on all inputs. The network will then perform gradient descent and return the fitted value of $c_0$. Further, if at any point in the gradient descent, the forward pass yields undefined results, the network returns the well-defined constants and associated output from the previous forward pass. For example, for $x_0^{c_0},$ after the network discovers that $c_0=2$ works, the gradient for the constants will be undefined because $c_0$ can only be an integer. Thus, the network will return the outputs of $x_0^{c_0},$ for $c_0=2,$ before the undefined gradients.

We find that if the network simply ignores functions with undefined outputs, these functions will increase in probability because our network regularization punishes many other functions. Since these punished functions decrease in probability during training, the functions with undefined outputs begin to increase in probability. To combat this, instead of ignoring undefined functions, we use a modified fitness for undefined functions, $K'_i(\mathcal{M},f) = -w_{\text{undef}} s,$ where $w_{\text{undef}}$ is a hyperparameter that can be tuned. This punishes undefined functions, causing their weights to decrease. In practice, we find that a value of $w_{\text{undef}}$ between 0 and 1 is most effective, depending on the application. Larger penalties may overly decrease probabilities of valid functions which are similar to an undefined function.

\subsubsection{OccamNet with Units}\label{sec:units}
Although we do not use this feature in our experiments, we also allow users to provide units for inputs and outputs. OccamNet will then regularize its functions so that they preserve the desired units.

To determine if a function $f$ preserves units, we first encode the units of each input and output. We encode an input parameter's units as a NumPy array in which each entry represents the power of a given base unit. For example, if for a problem the relevant units are kg, m, and s, and we have an input $F$ with units $\text{kg}\cdot\text{m}/\text{s}^2,$ we would represent $F$'s units as $[1,1,-2]$.

We then feed these units through the sampled function. Each primitive function receives a set of variables with units, may have requirements on those units for them to be consistent, and returns a new set of units. For example, $\sin(\cdot)$ receives one variable which it requires to have units $[0,\ldots,0]$ and returns the units $[0,\ldots,0]$. Similarly, $+(\cdot,\cdot)$ takes two variables with units that it requires to be equal and returns the same units. Using these rules, we propagate units through the function until we obtain units for the output. If at any point the input units for a primitive function do not meet that primitive's requirements, that primitive returns $[\infty,\ldots,\infty].$ Any primitive functions that receive $[\infty,\ldots,\infty]$ also return $[\infty,\ldots,\infty]$. Finally, if the output units of $f$ do not match the desired output units (including if $f$ outputs $[\infty,\ldots,\infty]$), we mark $f$ as not preserving units.

For the multiplication by a constant primitive function, $\cdot c$, we have to be careful. Because the units of $c$ are unspecified, this primitive function can produce any output units. As such, it returns $[\text{NaN},\ldots,\text{NaN}].$ If any primitive function receives $[\text{NaN},\ldots,\text{NaN}],$ it will either return $[\text{NaN},\ldots,\text{NaN}]$ if it has no constraints on the input units, or it will treat the $[\text{NaN},\ldots,\text{NaN}]$ as being the units required to meet the primitive function's consistency conditions. For example, if the $\sin(\cdot)$ function receives $[\text{NaN},\ldots,\text{NaN}]$, it will treat the input as $[0,\ldots,0],$ and if the $+(\cdot,\cdot)$ function receives $[1,2,3]$ and $[\text{NaN},\text{NaN},\text{NaN}],$ it will return $[1,2,3]$.

After sampling functions from OccamNet, we determine which functions do not preserve units. Because we wish to avoid these functions entirely, we bypass evaluating their normal fitness (thereby saving compute time) and instead assign a fitness of $K'_i(\mathcal{M},f) = -w_{\text{units}} s,$ where $s= n(\mathcal{M})/\sqrt{2\pi\sigma^2}$ is the maximum possible value of $K_i(\mathcal{M},f)$ and $w_{\text{units}}$ is a hyperparameter that can be tuned (set to 1 by default).

\subsection{Experimental Setup} \label{sec:fullSetup}
We divide this section as follows:\begin{enumerate}
    \item Section \ref{sec:hyper} describes the hyperparameters used for the experiments described in the main text testing OccamNet on Analytic Functions, Non-Analytic Functions, Implicit Functions, and Image Recognition.
    \item Section \ref{sec:PMLBSetup} describes the experimental setup for our tests with the PMLB datasets.
\end{enumerate}
\subsubsection{Experimental Setup and Hyperparameters for Non-PMLB Experiments}
\label{sec:hyper}
For the non-PMLB experiments, we terminate learning when the top-$v\lambda$ sampled functions all return the same fitness $K(\cdot, f)$ for 30 consecutive epochs. If this happens, these samples are equivalent function expressions. 

Computing the most likely DAG allows retrieval of the final expression. If this final expression matches the correct function, we determine that the network has converged. For pattern recognition, there is no correct target composition, so we measure the accuracy of the classification rule on a test split, as is conventional. Note that in the experiments where $E=0,$ we instead take an approximate of the highest probability function by taking the argmax of the weights into each argument node.

In all experiments, if termination is not met in a set number of steps, we consider it as not converged. We also keep a constant temperature for all the layers except for the last one. An increased last layer temperature allows the network to explore higher function compositionality, as shallow layers can be further trained before the last layer probabilities become concentrated; this is particularly useful for learning functions with high degrees of nesting. More details on hyperparameters for experiments are in the SM. Our network converges rapidly, often in only a few seconds and at most a few minutes.

In Tables~\ref{table:hyperparameters1} and \ref{table:hyperparameters2}, we present and detail the hyperparameters we used for our experiments in the main paper. Note that detail about the setup for each experiment is provided in the open source repositories available at \url{https://github.com/druidowm/OccamNet_Versions}.

In Tables~\ref{table:hyperparameters1} and \ref{table:hyperparameters2}, $+$ is addition (2 arguments); $-$ is subtraction (2 arguments) $\cdot$ is multiplication (2 arguments); $/$ is division (2 arguments); $\sin(\cdot)$ is sine, $+c$ is addition of a constant, $\cdot c$ is multiplication of a constant, $(\cdot)^c$ is raising to the power of a constant, $\leq$ is an if-statement (4 arguments: comparing two numbers, one return for a true statement, and one for a false statement); $-(\cdot)$ is negation. $\mathsf{MIN},$ $\mathsf{MAX}$, and $\mathsf{XOR}$ all have two arguments. Here, $\mathsf{SIGMOID}'$ is a sigmoid layer, and $\tanh'$ is a tanh layer where the inputs to both functions are scaled by a factor of 10, $+_4$, and $+_9$ are the operations of adding 4 and 9 numbers respectively, and $\mathsf{MAX}_4$, $\mathsf{MIN}_4$, $\mathsf{MAX}_9$ and $\mathsf{MIN}_9$ are defined likewise. The primitives for pattern recognition experiments are given as follows: $\mathbf{\Phi}_A$ consists of $\mathsf{SIGMOID}',$ $\mathsf{SIGMOID}',$ $\tanh',$ $\tanh',$ $+_4,$ $+_4,$ $+_9,$ $+_9,$ $+,$ $+,$ $\mathsf{MIN},$ $\mathsf{MIN},$ $\mathsf{MAX}$ and $\mathsf{MAX};$ $\mathbf{\Phi}_B$ consists of $\mathrm{id},$ $\mathrm{id},$ $\mathrm{id},$ $\mathrm{id},$ $+,$ $+,$ $+,$ $+_4,$ $+_4,$ $+_9,$ $+_9,$ $+_9,$ $\tanh,$, $\tanh,$ $\mathsf{SIGMOID},$ and $\mathsf{SIGMOID}.$ Additionally, the constants used for pattern recognition are $\mathbf{C}=\{-1,-1,0,0,1,1,1\}.$

In Tables~\ref{table:hyperparameters1} and \ref{table:hyperparameters2}, $L$ is the depth, $T$ is the temperature, $T_\mathrm{last}$ is the temperature of the final layer, $\sigma$ is the variance, $R$ is the sample size, $\lambda$ is the fraction of best fits, $\alpha$ is the learning rate, $E$ is the initialization parameter described in Section \ref{sec:initialization}, and $w_\phi,$ $w_\psi,$ $w_\xi,$ and $w_\gamma$ are as defined in Appendix \ref{sec:regularization}. Table \ref{table:hyperparameters1} does not include $E$ as a listed hyperparameter because for all experiments listed, $E=0.$ With $^*$ we denote the experiments for which the best model is without skip connections. We do not regularize for any experiments in Table~\ref{table:hyperparameters1}. NA entries mean that the corresponding hyperparameter is not present in the experiment. Note that the first three equations in Table~\ref{table:hyperparameters2} are not discussed in the main text. Instead, they are smaller experiments that we performed and which we discuss in the SM.

For all experiments in Table \ref{table:hyperparameters2}, we use a learning rate of 0.01 and, when applicable, a constant-learning rate of 0.05. We also set the temperature to 1 and the final layer temperature to 10 for all experiments in the table. For the equation $m_1v_1-m_2v_2 = 0,$ we sample $m_1,$ $v_1,$ and $m_2$ from $[-10,10]$ and compute $v_2$ using the implicit function.

All our experiments in Table \ref{table:hyperparameters1} use a batch size of 1000, except for \emph{Backprop OccamNet} and \emph{Finetune ResNet}, for which we use batch size 128. All our experiments in Table \ref{table:hyperparameters2} use a batch size of 200. For each of our pattern recognition experiments, we use a 90\%/10\% train/test random split for the corresponding datasets. The input pixels are normalized to be in the range [0, 1]. During validation, for \emph{MNIST Binary}, \emph{MNIST Trinary} and \emph{ImageNet Binary} the outputs of OccamNet are thresholded at 0.5. If the output matches the one-hot label, then the prediction is accurate; otherwise, it is inaccurate. For \emph{Backprop OccamNet} and \emph{Finetune ResNet} the outputs of OccamNet are viewed as the logits of a negative log likelihood loss function, so the prediction is the argmax of the logits. Backprop OccamNet and Finetune ResNet use an exponential decay of the learning rate with decay factor $0.999.$

\begin{table*}[hbt]
  \tiny 
  \centering
  \caption{Hyperparameters for Experiments Where $E=0$}
  \begin{tabular}{lccccc}
    \toprule
    \multicolumn{1}{c}{Target} & \multicolumn{1}{c}{Primitives} & \multicolumn{1}{c}{Constants} & \multicolumn{1}{c}{Range} & \multicolumn{1}{c}{$L$/ $T$/ $T_\mathrm{last}$/ $\sigma$} & 
    \multicolumn{1}{c}{$R$/ $\lambda$/ $\alpha$} 

    \\
    \cmidrule(r){1-1}
    \cmidrule(r){2-2}
    \cmidrule(r){3-3}
    \cmidrule(r){4-4}
    \cmidrule(r){5-5}
    \cmidrule(r){6-6}
    \multicolumn{6}{c}{Analytic Functions} \\
    \cmidrule(r){1-6}
     $2x^2+3x$ & $( \cdot,\cdot, +,+)$ & $\emptyset$ & $[-10,10]$ &
     2/1/1/0.01 & 50/5/0.05\\
    $\sin(3x+2)$ & $( \cdot,\sin,\sin,+,+)$ & 1, 2 & $[-10,10]$ &
     3/1/1/0.001 & 50/5/0.005\\
 $\sum_{n=1}^3\sin(nx)$ & $(\sin,\sin,+,+,+)$ & 1, 2 & $[-20,20]$ &
     5/1/1/0.001 & 50/5/0.005\\
     $(x^2+x)/(x+2)$ & $( \cdot,\cdot,+,+,/,/)$ & 1 & $[-6,6]$ &
     2/1/2/0.0001 & 100/5/0.005\\
     $x_0^2(x_0+1)/x_1^5$ & $(\cdot,\cdot,+,+,/,/)$ & 1 & $[[-10,10],[0.1,3]]$ &
     4/1/3/0.0001 & 100/10/0.002\\
     $x_0^2/2+(x_1+1)^2/2$ & $(\cdot,\cdot,+,+,/)$ & 1, 2 & $[[-20,-2],[2,20]]$ &
     3/1/2/0.1 & 150/5/0.005\\
    \cmidrule(r){1-6}
    \multicolumn{6}{c}{Program Functions} \\
    \cmidrule(r){1-6}
     $3x$ if $x>0$, else $x$ & $(\leq,\leq,\cdot,+,+,/)$ & 1 & $[-20,20]$ & 2/1/1.5/0.1 & 100/5/0.005 \\
     $x^2$ if $x>0$, else $-x$ & $(\leq,\leq,-(\cdot),+,+,-,\cdot)$ & 1 & $[-20,20]$ & 2/1/1.5/0.1 & 100/5/0.005 \\
     $x$ if $x>0$, else $\sin(x)$ & $(\leq,\leq,+,+,\sin,\sin)$ & 1 & $[-20,20]$ & 3/1/1.5/0.01 & 100/5/0.005 \\
     \multirow{2}{*}{$\mathsf{SORT}(x_0,x_1,x_2)$} & $(\leq,+,\mathsf{MIN},\mathsf{MAX},$  & \multirow{2}{*}{1, 2} & \multirow{2}{*}{$[-50,50]^4$} & \multirow{2}{*}{3/1/4/0.01} & \multirow{2}{*}{100/5/0.004} \\
     & $\mathsf{MAX}/,\cdot,-)$ \\
     $\mathsf{4LFSR}(x_0,x_1,x_2,x_3)$ & $( +, +, \mathsf{XOR},
     \mathsf{XOR})$ & $\emptyset$ & $\{0,1\}^4$ & 2/1/1/0.1 & 100/5/0.005 \\
     \\
    \cmidrule(r){1-1}
    $y_0(\vec{x})=x_1$ if $x_0<2$, else $-x_1$  & \multirow{2}{*}{$( \leq, \leq, -(\cdot), \cdot )$}
    & \multirow{2}{*}{1, 2} & \multirow{2}{*}{$[-5,5]^{2}$} & \multirow{2}{*}{3/1/3/0.01}
    & \multirow{2}{*}{100/5/0.002}\\
    $y_1(\vec{x})=x_0$ if $x_1<0$, else $x_1^2$ \\
    \cmidrule(r){1-1}
    \\
    \cmidrule(r){1-1}
    $g(x)=x^2$ if $x<2$, else $x/2$ & $( \leq, \leq, +, \cdot, \cdot, /,/)$
    & \multirow{2}{*}{1, 2}
    &\multirow{2}{*}{$[-8,8]$} & \multirow{2}{*}{2/1/2/0.01} & \multirow{2}{*}{100/5/0.005}
    \\
    $y(x)=g^{\circ 4}(x)$ \\
    \cmidrule(r){1-1}
    \\
    \cmidrule(r){1-1}
    $g(x)=x+2$ if $x<2$, else $x-1$ & $( \leq, \leq, +, +,$ 
    & \multirow{2}{*}{1, 2}
    &\multirow{2}{*}{$[-3,6]$} & \multirow{2}{*}{2/1/1.5/0.01} & \multirow{2}{*}{100/5/0.005}\\
    $y(x)=g^{\circ 2}(x)$ & $+, -,-)$
    \\
    \cmidrule(r){1-1}
    
    \cmidrule(r){1-6}
    \multicolumn{6}{c}{Pattern Recognition} \\
    \cmidrule(r){1-6}
    MNIST Binary & $\mathbf{\Phi}_A$& $\mathbf{C}$ & $[0,1]^{784}$& 2/1/10/0.01 & 150/ 10/0.05 \\
 MNIST Trinary  & $\mathbf{\Phi}_A$ & $\mathbf{C}$& $[0,1]^{784}$& 2/1/10/0.01 & 150/ 10/0.05 \\
    ImageNet Binary$^*$  & $\mathbf{\Phi}_A$& $\mathbf{C}$& $[0,1]^{2048}$& 4/1/10/10 & 150/10/0.0005 \\
    Backprop OccamNet$^*$  & $\mathbf{\Phi}_B$& $\mathbf{C}$& $[0,1]^{2048}$& 4/1/10/NA & NA/NA/0.1  \\
    Finetune ResNet$^*$  & $\mathbf{\Phi}_B$& $\mathbf{C}$& $[0,1]^{3\times224\times224}$& 4/1/10/NA & NA/NA/0.1  \\
    \bottomrule
  \end{tabular}
  \label{table:hyperparameters1}
\end{table*}

\begin{table*}[hbt]
  \tiny
  \centering
  \caption{Hyperparameters for Experiments Where $E=1$}
  \begin{tabular}{lcccccccc}
    \toprule
    \multicolumn{1}{c}{Target} & \multicolumn{1}{c}{Primitives} & \multicolumn{1}{c}{Constants} & \multicolumn{1}{c}{Range} & \multicolumn{1}{c}{$L$} & \multicolumn{1}{c}{$\sigma$} & 
     \multicolumn{1}{c}{$R$} & \multicolumn{1}{c}{$\lambda$} & \multicolumn{1}{c}{$w_\phi/w_\psi/w_\xi/w_\gamma$}

    \\
    \cmidrule(r){1-1}
    \cmidrule(r){2-2}
    \cmidrule(r){3-3}
    \cmidrule(r){4-4}
    \cmidrule(r){5-5}
    \cmidrule(r){6-6}
    \cmidrule(r){7-7}
    \cmidrule(r){8-8}
    \cmidrule(r){9-9}
    \multicolumn{9}{c}{Analytic Functions}\\
    \cmidrule(r){1-9}
     \multirow{2}{*}{$10.5x^3.1$} & $( +,-,\cdot,/,\sin,$ & \multirow{2}{*}{$\emptyset$} & \multirow{2}{*}{$[0,1]$} &
     \multirow{2}{*}{2}&\multirow{2}{*}{0.0005} & \multirow{2}{*}{200}&\multirow{2}{*}{10}&\multirow{2}{*}{0/0/0/0}\\
     & $\cos,+c,\cdot c,(\cdot)^c)$\\
     $\cos(x)$ & $( +,/,\sin)$ & $2,\pi$ & $[-100,100]$ &
     3&0.01 & 400&50&0/0/0/0\\
     $e^x$ & $( +,\cdot c,(\cdot)^c)$ & $10$ & $[0,1]$ &
     3&0.05 & 200&1&0.7/0.3/0.05/0.03\\
    \cmidrule(r){1-9}
     \multicolumn{9}{c}{Implicit Functions}\\
    \cmidrule(r){1-9}
     $x_0x_1=1$ & $( +,-,\cdot,/,\sin,\cos)$ & $\emptyset$ & $[-1,1]$ &
     2&0.01 & 400&1&0.7/0.3/0.15/0.1\\
    $x_0/x_1=1$ & $( +,-,\cdot,/,\sin,\cos)$ & $\emptyset$ & $[-1,1]$ &
     2&0.01 & 400 & 1 & 0.7/0.3/0.15/0.1\\
    $x_0^2+x_1^2=1$ & $( +,-,\cdot,/,\sin,\cos)$ & $\emptyset$ & $[-1,1]$ &
     2&0.01 & 200&10&0.7/0.3/0.15/0.1\\
     $x_0/\cos(x_1)=1$ & $( +,-,\cdot,/,\sin,\cos)$ & $\emptyset$ & $[-1,1]$ &
     2&0.01 & 200&10&0.7/0.3/0.15/0.1\\
     $m_1v_1-m_2v_2 = 0$ & $( +,-,\cdot,/,\sin,\cos)$ & $\emptyset$ & $[-10,10]^3$ &
     2&0.01 & 200&10&0.7/0.3/0.15/0.1\\
    \bottomrule
  \end{tabular}
\label{table:hyperparameters2}
\end{table*}

\subsubsection{PMLB Experiments Setup} \label{sec:PMLBSetup}
As described in the main text, we test OccamNet on 15 datasets from the Penn Machine Learning Benchmarks (PMLB) repository \cite{PMLB}. The 15 datasets chosen and the corresponding numbers we use to reference them, are shown in Table \ref{table:Datasets}. We chose these datasets by selecting the first 15 regression datasets with fewer than 1667 datapoints. These 15 datasets are the only datasets from PMLB we examine.

 We test four methods on these datasets. OccamNet-CPU, OccamNet-GPU, Eplex, AIF, and Extreme Gradient Boosting (XGB) \cite{XGBoost}. We have described all of these methods except for XGB in the main text. XGB is a tree-based method that was identified by \cite{WhereAreWe} as the best machine learning method based on validation MSE for modeling the PMLB datasets. However, XGB is not interpretable and thus cannot be used as a one-to-one comparison with OccamNet. Hence, although we provide the raw data for XGB's performance, we do not analyze it further. We train all methods except OccamNet-GPU on a single core of an Intel Xeon E5-2603 v4 @ 1.70GHz. For all methods, we use the primitive set $\Phi = \left(+(\cdot,\cdot),-(\cdot,\cdot),\times(\cdot,\cdot),\divisionsymbol(\cdot,\cdot),\sin(\cdot),\cos(\cdot),\exp(\cdot),\log|\cdot|\right).$ 

\begin{table}[bth]
  \centering
  \caption{Datasets Tested}
  \footnotesize
  \begin{tabular}{cccc}
    \toprule
    \# & Dataset & Size & \# Features\\
    \cmidrule(r){1-1}
    \cmidrule(r){2-2}
    \cmidrule(r){3-3}
    \cmidrule(r){4-4}
    1 & 1027\_ESL & 488 & 4\\
    2 & 1028\_SWD & 1000 & 10\\
    3 & 1029\_LEV & 1000 & 4\\
    4 & 1030\_ERA & 1000 & 4\\
    5 & 1089\_USCrime & 47 & 13\\
    6 & 1096\_FacultySalaries & 50 & 4\\
    7 & 192\_vineyard & 52 & 2\\
    8 & 195\_auto\_price & 159 & 15\\
    9 & 207\_autoPrice & 159 & 15\\
    10 & 210\_cloud & 108 & 5\\
    11 & 228\_elusage & 55 & 2\\
    12 & 229\_pwLinear & 200 & 10\\
    13 & 230\_machine\_cpu & 209 & 6\\
    14 & 4544\_GeographicalOriginalofMusic & 1059 & 117\\
    15 & 485\_analcatdata\_vehicle & 48 & 4\\
    
    \bottomrule
  \end{tabular}
  \label{table:Datasets}
\end{table}

For each dataset, we perform grid search to identify the best hyperparameters. The hyperparameters searched for the two OccamNet runs are shown in Table \ref{table:gridHyperparameters}. The other hyperparameters not used in the grid search are set as follows: $T=10,$ $T_{\text{last}} =10,$ $w_\phi = w_\psi = w_\xi = w_\gamma = 0,$ and the dataset batch size is the size of the training data. For OccamNet-GPU, we set $R$ to be approximately as large as can fit on the V100 GPU, which varies between datasets. See Table \ref{table:functionSampleSize} for the exact number of functions tested for each dataset for OccamNet-GPU. For XGBoost, we use exactly the same hyperparameter grid as used in \citet{WhereAreWe}. For Eplex, we use the same hyperparameter grid as used in \citet{WhereAreWe}, with the exception that we use a depth of 4 to match that of OccamNet.

\begin{table*}[bth]
  \centering
  \caption{OccamNet Hyperparameters}
  \footnotesize
  \begin{tabular}{ccc}
    \toprule
    Hyperparameter & OccamNet-CPU & OccamNet-GPU \\
    \cmidrule(r){1-1}
    \cmidrule(r){2-2}
    \cmidrule(r){3-3}
    $\alpha$ & $\{0.5,1\}$ & $\{0.5,1\}$\\
    $\sigma$ & $\{0.5,1\}$ & $\{0.1,0.5,1\}$\\
    $E$ & $\{1,5\}$ & $\{0,1,5\}$\\
    $\lambda/R$ & $\{0.1,0.5,0.9\}$ & $\{0.1,0.5,0.9\}$\\
    $R$ & $\{500,1000,2000\}$ & max\\
    $N$ & $1000000/R$ & 1000\\
    
    \bottomrule
  \end{tabular}
  \label{table:gridHyperparameters}
\end{table*}

\begin{table*}[tbh]
  \centering
  \caption{Number of Functions Sampled Per Epoch}
  \footnotesize
  \begin{tabular}{cr}
    \toprule
    \# & \multicolumn{1}{c}{$R$} \\
    \cmidrule(r){1-1}
    \cmidrule(r){2-2}
    1 & 17123\\
    2 & 8333\\
    3 & 8333\\
    4 & 8333\\
    5 & 178571\\
    6 & 166666\\
    7 & 161290\\
    8 & 52631\\
    9 & 52631\\
    10 & 78125\\
    11 & 151515\\
    12 & 41666\\
    13 & 40000\\
    14 & 7874\\
    15 & 178571\\
    \bottomrule
  \end{tabular}
  \label{table:functionSampleSize}
\end{table*}

We select the best run from the grid search as follows. For each hyperparameter combination, we first identify the models with the lowest training MSE and the lowest validation MSE:\begin{itemize}
    \item For OccamNet-CPU and OccamNet-GPU, we examine the highest probability function after each epoch. From these functions, we select the function with the lowest testing MSE and the function with the lowest validation MSE.
    \item For Eplex, we examine the highest-fitness individual from each generation. From these individuals, we select the individual with the lowest training MSE and the individual with the lowest validation MSE.
    \item For XGBoost, we train the model until the validation loss has not decreased for 100 epochs. We then return this model as the model with the best training MSE and validation MSE.
\end{itemize} Once we have the models with the lowest training and validation MSE for each hyperparameter combination, we identify the overall model with the lowest training MSE from the set of lowest training MSE models, and we identify the overall model with the lowest validation MSE from the set of lowest validation MSE models. We then record these models' training MSE and validation MSE as the best training MSE and validation MSE, respectively. Finally, we test the model with the overall lowest validation MSE on the testing dataset and record the result as the grid search testing MSE.

For our test of OccamNet and Eplex's scalability on the PMLB datasets, we use the same hyperparameter combinations as those listed described above, except that, as described in the main text, we run OccamNet-GPU with 250, 1000, 4000, 16000, and 64000 functions sampled per epoch and Eplex with 250, 500, 1000, 2000 and 4000 functions sampled per epoch. Our evaluation of training, validation, and testing loss is exactly the same as described above, except that we evaluate the lowest losses for each value of $N$ instead of grouping $N$ with all of the other hyperparameters.

\appendix
\section*{Supplemental Material}
\renewcommand{\thesubsection}{\Alph{subsection}}

We have organized the Supplemental Material as follows: \begin{itemize}
    \item  In Section~\ref{sec:PMLBResults} we provide the results of our experiments on the PMLB datasets.
    \item  In Section~\ref{sec:PMLBFits} we examine the fits each method provides for the PMLB Datasets.
    \item  In Section~\ref{sec:PMLBScaleAnalysis} analyze the results of the experiment scaling OccamNet-GPU on the PMLB datasets.
    \item In Section~\ref{sec:ablation} we present a series of ablation studies.
    \item In Section~\ref{sec:neural} we discuss neural models for sorting and pattern recognition.
    \item In Section~\ref{sec:moreExperiments} we discuss a few small experiments we tested.
    \item In Section~\ref{sec:related} we discuss research related to the various applications of OccamNet.
    \item In Section~\ref{sec:SymbolicRegression} we discuss the evolutionary strategies for fitting functions and programs that we use as benchmarks.
    \item In Section~\ref{sec:codeApp}, we catalog our code and video files.
\end{itemize}

\section{PMLB Experiment Results}\label{sec:PMLBResults}
\begin{figure*}
    \centering
    
    \includegraphics[width=\linewidth]{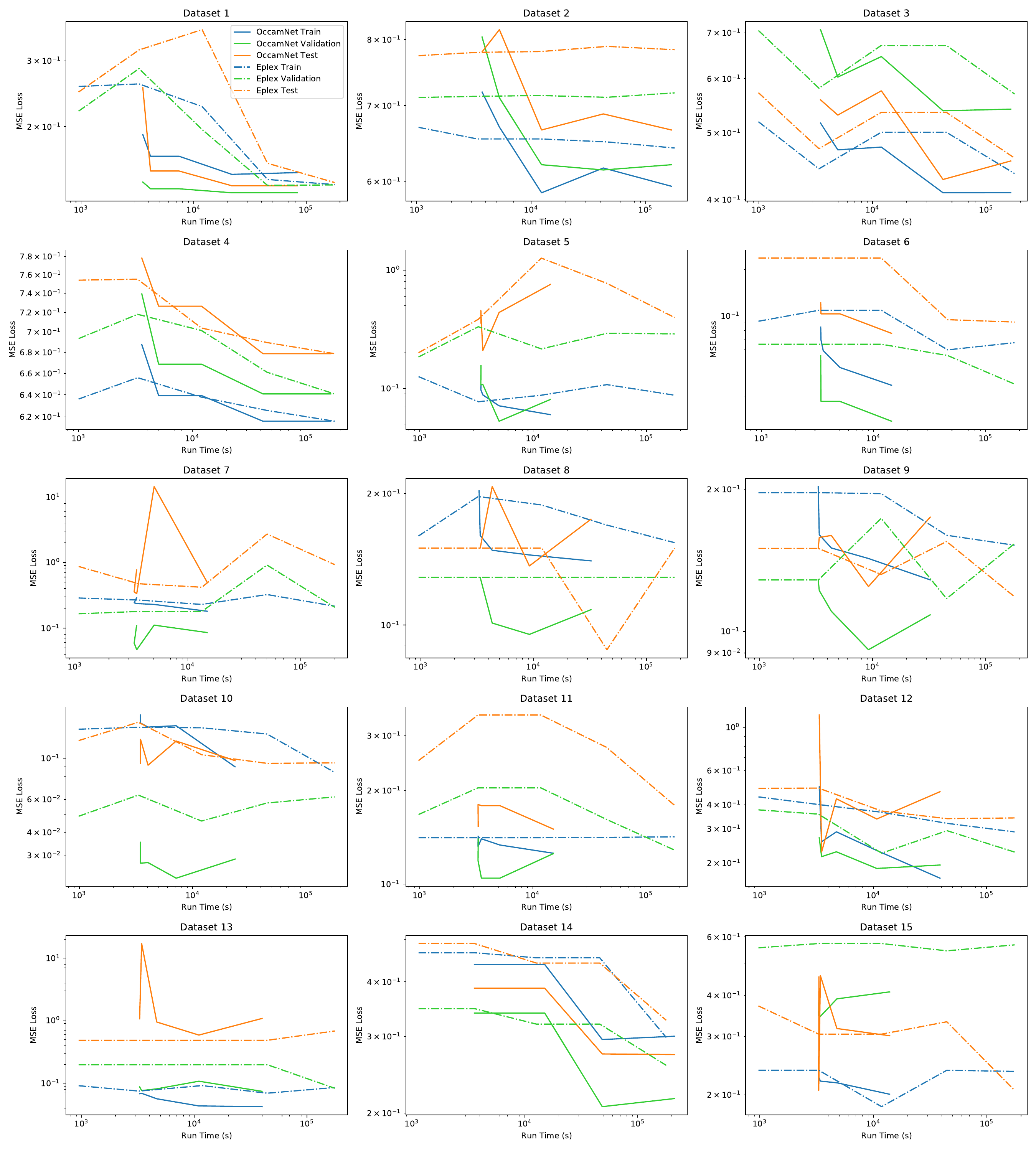}
    
    \caption{OccamNet-GPU and Eplex's Training, Validation, and Testing MSE as a function of run time for the 15 PMLB datasets discussed above.}
    
    \label{fig:PMLBScalingFullHyperparams_App}
\end{figure*}

\begin{figure*}
    \centering
    
    \includegraphics[width=\linewidth]{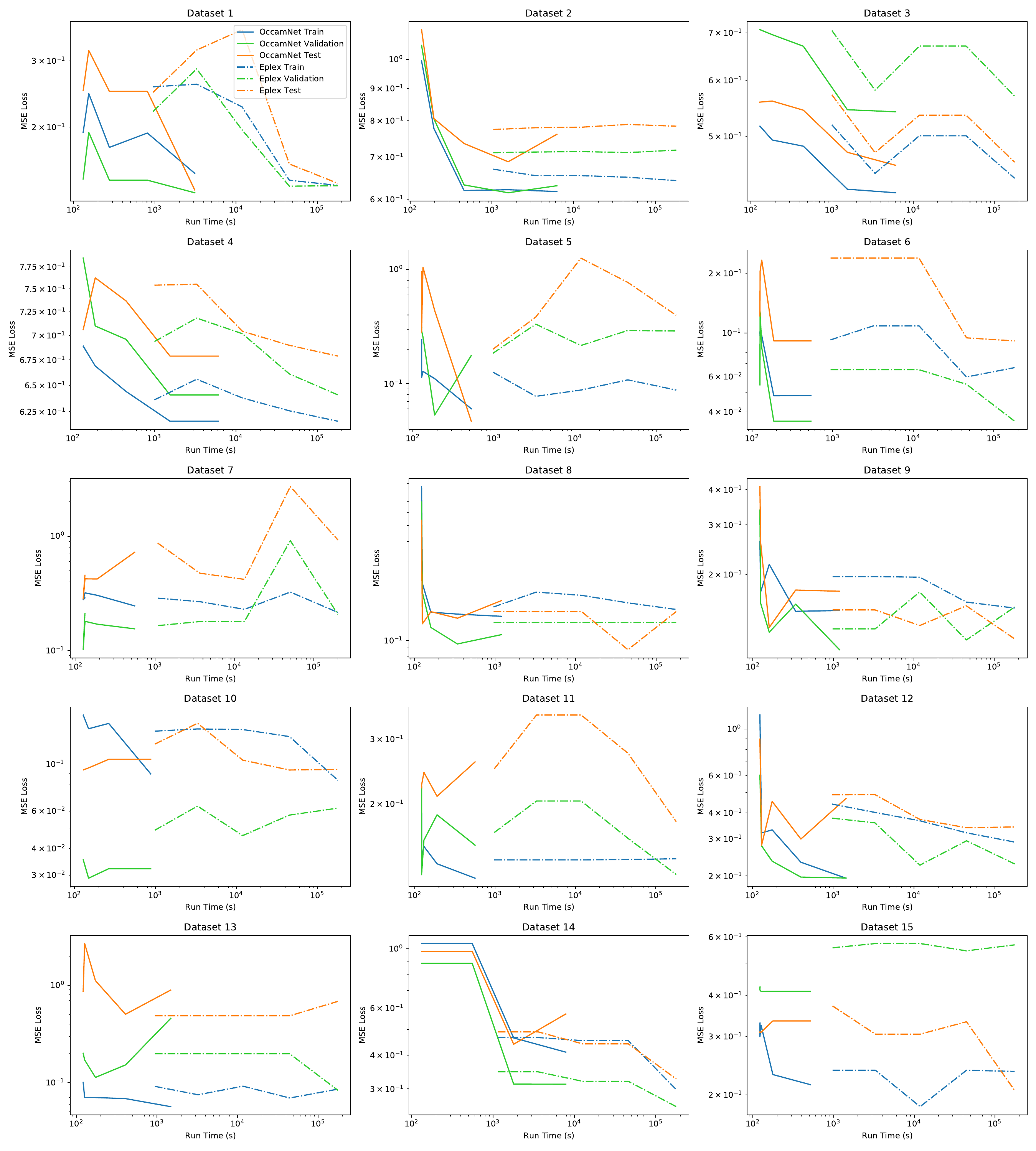}
    
    \caption{OccamNet-GPU and Eplex's Training, Validation, and Testing MSE as a function of run time for the 15 PMLB datasets discussed above. For this figure, we only consider the losses for a restricted subset of hyperparameter combinations.}
    
    \label{fig:PMLBScalingRestrictedHyperparams_App}
\end{figure*}

The raw data for the PMLB experiments are shown in Table \ref{tab:rawData}. To improve readability, we use red highlighting and bold text to illustrate the best-performing model for each dataset and metric. We compare OccamNet-CPU, OccamNet-GPU, Eplex, and AIF, marking the method with the lowest MSE or training time in red. We also compare OccamNet-GPU, Eplex, and AIF, marking the method with the lowest MSE or training time in bold.  Plots of the full results for the PMLB scaling experiment are shown in Figures \ref{fig:PMLBScalingFullHyperparams_App} and \ref{fig:PMLBScalingRestrictedHyperparams_App}. As discussed in Section \ref{sec:PMLBScaleAnalysis}, Figure \ref{fig:PMLBScalingRestrictedHyperparams_App} shows OccamNet-GPU's performance when only considering a restricted set of hyperparameters.

\begin{table*}[t]
  \centering
  \caption{Raw data from the PMLB experiments. Hyperparameters and best fits are in the following path in our code (see Section~\ref{sec:codeApp}): \texttt{pmbl-experiments/pmlb-results}.}
  \label{tab:rawData}
  \small
  \begin{tabular}{lcccccccccc}
    \toprule
    \multicolumn{6}{c}{Training Loss (MSE)} & \multicolumn{5}{c}{Validation Loss (MSE)}\\
    \cmidrule(r){1-6}
    \cmidrule(r){7-11}
    \# & OccamNet-CPU & OccamNet-GPU & Eplex & AIF & XGB & OccamNet-CPU & OccamNet-GPU & Eplex & AIF & XGB\\
    \cmidrule(r){1-1}
    \cmidrule(r){2-2}
    \cmidrule(r){3-3}
    \cmidrule(r){4-4}
    \cmidrule(r){5-5}
    \cmidrule(r){6-6}
    \cmidrule(r){7-7}
    \cmidrule(r){8-8}
    \cmidrule(r){9-9}
    \cmidrule(r){10-10}
    \cmidrule(r){11-11}
    1 & 0.177 & \textbf{0.139} & {\color{red} 0.153} & 0.465 & 0.056 & 0.141 & 0.137 & \textbf{\color{red} 0.128} & 0.449 & 0.133 \\
    2 & {\color{red} 0.607} & \textbf{0.605} & 0.643 &       & 0.443 & {\color{red} 0.647} & \textbf{0.640} & 0.702 &       & 0.567 \\
    3 & 0.486 & \textbf{0.432} & {\color{red} 0.443} &       & 0.326 & 0.634 & 0.597 & \textbf{\color{red} 0.581} &       & 0.556 \\
    4 & 0.639 & 0.616 & \textbf{\color{red} 0.616} & 0.886 & 0.547 & 0.662 & 0.641 & \textbf{\color{red} 0.641} & 1.040 & 0.649 \\
    5 & 0.107 & \textbf{0.054} & {\color{red} 0.105} &       & 0.000 & {\color{red} 0.145} & \textbf{0.108} & 0.182 &       & 0.134 \\
    6 & 0.070 & \textbf{0.035} & {\color{red} 0.067} & 0.162 & 0.000 & 0.037 & \textbf{0.017} & {\color{red} 0.036} & 0.066 & 0.114 \\
    7 & {\color{red} 0.228} & \textbf{0.161} & 0.230 & 0.713 & 0.039 & {\color{red} 0.047} & \textbf{0.099} & 0.122 & 0.802 & 0.175 \\
    8 & 0.155 & \textbf{0.145} & {\color{red} 0.152} &       & 0.000 & {\color{red} 0.095} & 0.115 & \textbf{0.097} &       & 0.105 \\
    9 & 0.168 & \textbf{0.141} & {\color{red} 0.152} &       & 0.000 & {\color{red} 0.114} & \textbf{0.097} & 0.129 &       & 0.105 \\
    10 & 0.154 & \textbf{0.101} & {\color{red} 0.130} & 0.171 & 0.000 & {\color{red} 0.027} & \textbf{0.021} & 0.036 & 0.044 & 0.162 \\
    11 & {\color{red} 0.136} & \textbf{0.129} & 0.141 & 0.177 & 0.029 & {\color{red} 0.119} & 0.162 & \textbf{0.161} & 0.178 & 0.106 \\
    12 & {\color{red} 0.255} & \textbf{0.167} & 0.324 &       & 0.000 & {\color{red} 0.193} & \textbf{0.177} & 0.310 &       & 0.083 \\
    13 & {\color{red} 0.062} & \textbf{0.042} & 0.082 & 0.103 & 0.004 & {\color{red} 0.074} & \textbf{0.076} & 0.198 & 0.289 & 0.163 \\
    14 & 0.573 & 0.438 & \textbf{\color{red} 0.414} &       & 0.000 & 0.470 & \textbf{0.312} & {\color{red} 0.320} &       & 0.196 \\
    15 & {\color{red} 0.208} & \textbf{0.183} & 0.216 & 0.456 & 0.000 & {\color{red} 0.174} & \textbf{0.411} & 0.567 & 0.524 & 0.175 \\
    \cmidrule(r){1-6}
    \cmidrule(r){7-11}
    \multicolumn{6}{c}{Testing Loss (MSE)} & \multicolumn{5}{c}{Average Run Time (s)}\\
    \cmidrule(r){1-6}
    \cmidrule(r){7-11}
    \# & OccamNet-CPU & OccamNet-GPU & Eplex & AIF & XGB & OccamNet-CPU & OccamNet-GPU & Eplex & AIF & XGB\\
    \cmidrule(r){1-1}
    \cmidrule(r){2-2}
    \cmidrule(r){3-3}
    \cmidrule(r){4-4}
    \cmidrule(r){5-5}
    \cmidrule(r){6-6}
    \cmidrule(r){7-7}
    \cmidrule(r){8-8}
    \cmidrule(r){9-9}
    \cmidrule(r){10-10}
    \cmidrule(r){11-11}
    1 & 0.251 & \textbf{0.141} & {\color{red} 0.223} & 0.741 & 0.147 & {\color{red} 2246} & 319 & 4574 & 4498 & 5\\
    2 & {\color{red} 0.704} & \textbf{0.706} & 0.742 &       & 0.648 & 4789 & 302 & {\color{red} 4651} &      & 7\\
    3 & 0.542 & 0.491 & \textbf{\color{red} 0.474} &       & 0.464 & 4767 & 300 & {\color{red} 4696} &      & 6\\
    4 & 0.713 & 0.679 & \textbf{\color{red} 0.679} & 0.895 & 0.659 & 4511 & 308 & 4729 & {\color{red} 4222} & 5\\
    5 & 0.589 & 0.209 & \textbf{\color{red} 0.116} &       & 0.113 & {\color{red} 239} & 504 & 4684 &       & 3\\
    6 & 0.151 & \textbf{0.082} & {\color{red} 0.091} & 0.088 & 0.234 & {\color{red} 244} & 479 & 4755 & 5076  & 3\\
    7 & {\color{red} 0.335} & 0.761 & 0.829 & \textbf{0.698} & 0.316 & {\color{red} 249} & 468 & 4583 & 1222  & 1\\
    8 & 0.137 & 0.132 & \textbf{\color{red} 0.119} &       & 0.094 & {\color{red} 751} & 359 & 4516 &       & 5\\
    9 & {\color{red} 0.135} & 0.194 & \textbf{0.150} &       & 0.094 & {\color{red} 764} & 354 & 4517 &       & 5\\
    10 & {\color{red} 0.084} & \textbf{0.086} & 0.097 & 0.109 & 0.080 & {\color{red} 483} & 373 & 4605 & 6196 & 2\\
    11 & {\color{red} 0.180} & \textbf{0.177} & 0.275 & 0.274 & 0.150 & {\color{red} 259} & 455 & 4639 & 1223 & 2\\
    12 & {\color{red} 0.223} & \textbf{0.214} & 0.426 &       & 0.165 & {\color{red} 944} & 335 & 4653 &      & 3\\
    13 & 1.088 & \textbf{0.157} & {\color{red} 0.487} & 0.864 & 0.622 & {\color{red} 956} & 364 & 4561 & 6899 & 5\\
    14 & 0.570 & \textbf{0.440} & {\color{red} 0.442} &       & 0.228 & 6562 & 354 & {\color{red} 4785} &     & 139\\
    15 & 1.664 & \textbf{0.334} & 0.441 & {\color{red} 0.324} & 0.188 & {\color{red} 264} & 487 & 4533 & 586 & 29\\
    \bottomrule
  \end{tabular}
  \label{table:PMLBErrorTable}
\end{table*}

\section{Analysis of Fits to PMLB Datasets}\label{sec:PMLBFits}
 In this section, we analyze the fits that the methods discussed in Section \ref{sec:PMLBSetup} provide for the PMLB dataset.
 
 OccamNet-CPU and OccamNet-GPU provide solutions which are all short, easy to comprehend fits to the data. We find that OccamNet uses addition, subtraction, multiplication, and division most extensively, exploiting $\sin(\cdot)$ and $\cos(\cdot)$ for more nonlinearity. Interestingly, OccamNet uses $\exp(\cdot)$ and $\log|\cdot|$ less frequently, perhaps because both functions can vary widely with small changes in input, making functions with these primitives more likely to represent poor fits.
 
 OccamNet's solutions demonstrate its ability to exploit modularity and reuse components. These solutions often have repeated components, for example in dataset \#1, 1027\_ESL, where OccamNet-CPU's best fit to the training data is \begin{equation*}
     y_0=\frac{(\sin(x_2)+x_3+x_1)\cdot(\sin(x_2)+x_3+x_1)}{(\sin(x_2)+x_3+x_1)+(x_3+x_1)+(x_1+x_3)}.
 \end{equation*} In this fit, OccamNet-CPU builds $\sin(x_2)+x_3+x_1$ in the first two layers of the network and then reuses it three times. Solutions like the above demonstrate OccamNet's ability to identify successful subcomponents of a solution and then to rearrange the subcomponents into a more useful form. Examples like the above, however, also demonstrate that OccamNet often overuses modularity, potentially restricting the domain of functions it can search. We suspect that the main reason that OccamNet may rely too heavily on modularity in some fits is that OccamNet uses an extremely high learning rate of 1 for its training. We used such a large learning rate to allow OccamNet to converge even when faced with $10^{30}$ or more functions. However, we suspect that this may also cause OccamNet to converge to certain paths before exploring sufficiently. For example, with the function above, OccamNet may have identified that $\sin(x_2)+x_3+x_1$ is a useful component and, because of its high learning rate, used this pattern several times instead of the one time needed. This hypothesis is supported by the fact that OccamNet-GPU, which samples many more functions before taking a training step, repeats patterns less frequently than OccamNet-CPU. For example OccamNet-GPU's best-fit solution for the training dataset of dataset \#1 is \begin{equation*}
     y_0=\cos(x_1/x_1)\cdot\cos(x_1/x_1)\cdot(x_2+x_3+\sin(x_0)+x_3+x_1-\sin(x_3)),
 \end{equation*} which contains almost no repetition.
 
 Remarkably, for dataset \#4, 1030\_ERA, both Eplex and OccamNet-GPU discover equivalent functions for both training and validation: OccamNet-GPU discovers \begin{equation*}
     y_0=\cos(\sin(x_1-x_2))\cdot(\sin(x_2)+x_0+x_1)\cdot\cos(x_2/x_2),
 \end{equation*} and Eplex discovers \begin{equation*}
     y_0 = \cos(x_1/x_1)\cdot (\sin(x_2)+x_0+x_1)\cdot\cos(\sin(x_2-x_1)).
 \end{equation*} As a result, the two methods' losses are identical up to 7 decimal places. Still, we mark Eplex as performing better on this dataset because after the seventh decimal place it has a slightly lower loss, likely due to differences in rounding or precision between the two approaches. Two different methods identifying the same function is extremely unlikely; OccamNet's search space includes $2\cdot 10^{30}$ paths for this dataset, meaning that the probability of both methods identifying this function purely by chance is minuscule. In combination with the fact that this function was the best fit to both the training and validation datasets for both methods, this suggests that the identified function is a nearly optimal fit to the data for the given search space. Given the size of the search space, this result thus provides further evidence that OccamNet and Eplex perform far better than brute-force search. Interestingly, although OccamNet-CPU did not discover this function, it's best fit for the validation,\begin{equation*}
     y_0=\sin(x_2/x_2)\cdot(\sin(x_2)+x_0+x_1)\cdot\cos(\cos(x_3))\cdot\cos(\sin(x_3)),
 \end{equation*} does include several features present in the fits found by OccamNet-GPU and Eplex, such as the $\sin(x_2)+x_0+x_1$ term, the $\cos(\sin(\cdot))$ term, and the $x_2/x_2$ inside of the trigonometric function. This suggests that OccamNet-CPU may also have been close to converging to the function discovered by Eplex and OccamNet-GPU. OccamNet-CPU's loss was also always within 5\% of Eplex's loss on this dataset, again suggesting that OccamNet-CPU had identified a function close to that of Eplex and OccamNet-GPU.
 
 Interestingly, AI Feynman 2.0's fits generally tend to be very simple compared to those of OccamNet-CPU and OccamNet-GPU. For example, AIF's fit for the training dataset \#11 is \begin{equation*}
     y_0 = -0.050638447726+\log(x_0/\sin(x_0))-x_0,
 \end{equation*} whereas OccamNet-CPU's fit is \begin{equation*}
     y_0=\sin(x_0)\cdot x_1\cdot x_0\cdot \sin(x_0)\cdot\log|x_0|-\cos(x_1\cdot x_0-x_1).
 \end{equation*} AI Feynman's fit is slightly simpler and easier to interpret, but it comes at the cost of having a 35\% higher loss. We suspect that because the PMLB datasets likely do not have modular representations, AI Feynman must rely mainly on its brute-force search, which ultimately produces shorter expressions. AI Feynman can also produce constants because of its polynomial fits, and it uses constants in nearly every solution it proposes. We did not allow the other symbolic methods to fit constants, but they still consistently performed better than AI Feynman, suggesting that fitting constants may not be essential to accurately modeling the PMLB datasets.
 
 As discussed in the main text, OccamNet-CPU is considerably faster than Eplex, often running faster by more than an order of magnitude. This may be in part because we train Eplex with the DEAP evolutionary computation framework \cite{DEAP}, which is implemented in Python and utilizes NumPy arrays for computation. Thus, our implementation of Eplex may be somewhat slower than an implementation written in C. However, because of its selection based on many fitness cases, Eplex is also by nature considerably slower than many other genetic algorithms, running in $O(TN^2),$ where $T$ is the number of fitness cases and $N$ is the population size \cite{EplexIsSlow}. This suggests that even a pure C implementation of Eplex may not be as fast as OccamNet-CPU. More recent selection algorithms perform comparably to Eplex but run significantly faster, for example Batch Tournament Selection \cite{EplexIsSlow}. However, because these methods did not exist at the time of \citet{WhereAreWe}, they have not been compared to other methods on the PMLB datasets. Thus, we have not tested these methods here. On the other hand, our current implementation of sampling and the forward pass work with DAGs in which an edge leads to each argument node, regardless of whether the argument node is connected to the outputs. The result is that our implementation of OccamNet evaluates more than $|\Phi|$ times more primitive functions than is necessary, where $|\Phi|$ is the number of primitive functions. In the case of these experiments, this amounts to more than eight times the number of calculations necessary. It may be possible to optimize OccamNet by not evaluating such unused connections, thereby obtaining a much faster runtime.

\section{Analysis of PMLB Scaling Tests}\label{sec:PMLBScaleAnalysis}

As can be seen in Figure \ref{fig:PMLBScalingFullHyperparams_App}, OccamNet-GPU's training loss decreases with increasing sample size for every dataset -- the training loss for 64000 functions sampled is always less than the training loss for 250 functions sampled. For some datasets, OccamNet-GPU's training loss is not monotonically decreasing, but this is to be expected given OccamNet-GPU's inherent randomness and the size of the search space. 

For datasets 1, 2, 3, 4, 13, and 14, the training loss does not drop noticeably when increasing the sample size beyond a certain point. There are two possible explanations for this. (\emph{i}) For all of these datasets, OccamNet-GPU's training loss is very close to or lower than Eplex's best training loss, suggesting that OccamNet-GPU may be approaching an optimal fit and that there is little room to further decrease the loss. (\emph{ii}) There may be critical sample sizes before which OccamNet-GPU's training loss is stagnant and beyond which its training loss begins to decrease. This is apparent in datasets 1, 3, 4, 10, 12, and 14, where the OccamNet training loss temporarily stops decreasing at 1000 or 4000 functions sampled. It is possible that for some datasets another such critical sample size exists beyond 64000 functions.

For datasets 1, 2, 3, 4, 6, 12, and 14, OccamNet-GPU's validation loss also decreases with the number of functions sampled. However, for datasets 5, 7, 8, 9, 10, and 11, it initially decreases and then begins to increase, and for datasets 13 and 15 it does not decrease. Interestingly, the datasets for which OccamNet-GPU's testing loss does not decrease are generally the same as the datasets for which OccamNet-GPU's validation loss does not decrease. The datasets where the validation and testing loss do not decrease are generally very small, with around 200 or fewer datapoints. This suggests that OccamNet-GPU is overfitting. Such overfitting is to be expected given the small number of samples in the PMLB datasets. On the other hand, Eplex only seems to overfit in datasets 5 and 9. Because overfitting results from fitting a training or validation dataset too well, this is further evidence that OccamNet-GPU is fitting the training datasets better than Eplex.

Note that for each number of functions sampled, we tested 81 different hyperparameter combinations for OccamNet-GPU and only 3 for Eplex. This is largely because, as a new architecture, OccamNet-GPU's optimal hyperparameters are not known. For a fair comparison, the runtimes we report in Figure \ref{fig:PMLBScalingFullHyperparams_App} are the times required to run all hyperparameter combinations. Thus, because OccamNet-GPU uses 27 times more hyperparameter combinations, its speed advantage is lessened, although still significant.

After examining the results for all hyperparameter combinations, however, we noted that all hyperparameters but the learning rate had an ``optimal'' value, listed in Appendix \ref{sec:PMLBSetup}. Restricting to only the remaining three hyperparameter combinations produces a best training loss that is often the same as, and is never more than 40\% greater than, the lowest loss among all 81 hyperparameters. Figure \ref{fig:PMLBScalingRestrictedHyperparams_App} shows the results when considering only OccamNet-GPU's restricted hyperparameter combinations for three datasets.

When OccamNet-GPU and Eplex are restricted to the same number of hyperparameter combinations, OccamNet-GPU always runs faster when sampling 64000 functions per epoch than Eplex does when sampling 1000 functions per epoch. OccamNet-GPU's training loss consistently decreases with increasing sample size, although its validation and testing losses do not always follow such a clear trend. Further, OccamNet-GPU almost always converges to training and validation losses that are close to or less than Eplex's training and validation losses. OccamNet-GPU's best training loss is less than or approximately the same as Eplex's best training loss for all but datasets 1, 14, and 15.

Dataset 14 consists of over 1000 datapoints with 117 features, so it is likely one of the most difficult datasets which we test. The fact that OccamNet-GPU does perform comparatively to Eplex when it tests additional hyperparameter combinations suggests that for such difficult problems OccamNet-GPU benefits from additional hyperparameter exploration, particularly involving weight initialization.

For dataset 15, because Eplex only identifies a better function than OccamNet-GPU when it samples 1000 functions and not when it samples 2000 or 4000 functions, Eplex's better performance appears to be somewhat of an outlier.

With the restricted set of hyperparameters, OccamNet-GPU still overfits on every dataset it overfitted on when using the full set of hyperparameters, suggesting that the overfitting is not due to the large number of hyperparameters. 
Interestingly, for both the full and restricted hyperparameter versions of OccamNet-GPU, OccamNet-GPU and Eplex again identify the same fit for Dataset 4, \begin{equation*}
     y_0=\cos(\sin(x_1-x_2))\cdot(\sin(x_2)+x_0+x_1)\cdot\cos(x_3/x_3).
 \end{equation*}

\section{Ablation Studies}
\label{sec:ablation}

We test the performance of various hyperparameters in a collection of ablation studies, as shown in Table \ref{table:ablation}. Here, we focus on what our experiments demonstrate to be the most critical parameters to be tuned: the collection of primitives and constants, the network depth, the variance of our interpolating function, the overall network temperature (as well as the last layer temperature), and, finally, the learning rate of our optimizer. As before, we set the stop criterion and terminate learning when the top-$\lambda$ sampled functions all return the same fitness $K(\cdot, f)$ for 30 consecutive epochs. If this does not occur in a predefined, fixed number of iterations, or if the network training terminates and the final expression does not match the correct function we aim to fit, we say that the network has not converged. All hyperparameters for baselines are specified in Section \ref{sec:hyper}, except for the sampling size, which is set to $R=100$.

\begin{table*}[hbt]
  \centering
  \caption{Ablation studies on representative experiments}
    \begin{tabular}{lcc}
    \toprule
     \multicolumn{1}{c}{Modification} & Convergence fraction $\eta$ & Convergence epochs $T_c$\\
     \cmidrule(r){1-1}
     \cmidrule(r){2-2}
     \cmidrule(r){3-3}
     \multicolumn{3}{c}{Experiment $\sin(3x+2)$} \\
     \midrule
     baseline & 10/10 & 390 \\
     added constants (2) and primitives ($\cdot,(\cdot)^2,-(\cdot)$)& 10/10 & 710 \\
     lower last layer temperature (0.5) & 10/10 & 300 \\
     higher last layer temperature (3) & 10/10 & 450 \\
     lower learning rate (0.001) & 10/10 & 2500 \\
     higher learning rate (0.01) & 10/10 & 170 \\ 
     deeper network (6) & 8/10 & 3100 \\
     lower variance (0.0001) & 10/10 & 390 \\
     higher variance (0.1) & 10/10 & 450 \\
     lower sampling (50) & 10/10 & 680 \\
     higher sampling (250) & 10/10 & 200 \\
     \midrule
     \multicolumn{3}{c}{Experiment $x^2$ if $x>0$, else $-x$}\\
     \midrule
     baseline & 10/10 & 100 \\
     added constants (1, 2) and primitives ($-,-(\cdot)$)& 10/10 & 290 \\
     lower last layer temperature (0.5) & 10/10 & 160 \\
     higher last layer temperature (3) & 10/10 & 150 \\
     lower learning rate (0.001) & 10/10 & 780 \\
     higher learning rate (0.01) & 10/10 & 90 \\ 
     deeper network (6) & 10/10 & 180 \\
     shallower network (2) & 10/10 & 160 \\
     lower variance (0.001) & 10/10 & 160 \\
     higher variance (1) & 10/10 & 180 \\
     lower sampling (50) & 10/10 & 290 \\
     higher sampling (250) & 10/10 & 140 \\
    \bottomrule
    \end{tabular}

    \label{table:ablation}
\end{table*}

Our benchmarks use a sampling size large enough for convergence in most experiments. It is worth noting, however, that deeper networks sometimes failed to converge (with a convergence fraction of $\eta = 8/10$) for the analytic function we tested. Deeper networks allow for more function composition and let approximations emerge as local minima: in practice, we find that increasing the last layer temperature or reducing the variance is often needed to allow for a larger depth $L$. For pattern recognition, we found that \emph{MNIST Binary} and \emph{Trinary} require depth 2 for successful convergence, while the rest of the experiments require depth 4. Shallower or deeper networks either yield subpar accuracy or fail to converge. We also find that for OccamNet without skip connections, larger learning rates usually work best, i.e., 0.05 works best, while OccamNet with skip connections requires a smaller learning rate, usually around 0.0005. We also tested different temperature and variance schedulers in the spirit of simulated annealing. In particular, we tested increasing or decreasing these parameters over training epochs, as well as sinusoidally varying them with different frequencies. Despite the increased convergence time, however, we did not find any additional benefits of using schedulers. As we test OccamNet in larger problem spaces, we will revisit these early scheduling studies and investigate their effects in those domains.

\section{Neural Approaches to Benchmarks}
\label{sec:neural}
Since OccamNet is a neural model that is constructed on top of a fully connected neural architecture, below we consider a limitation of the standard fully connected architectures for sorting and then a simple application of our temperature-controlled connectivity.

\textbf{E-1. Exploring the limits of fully connected neural architectures for sorting}

We made a fully connected neural network with residual connections. We used the mean squared error (MSE) as the loss function. The output size was equal to the input size and represented the original numbers in sorted order. 
We used $L_2$ regularization along with Adam optimization. We tested weight decay ranging from 1e-2 to 1e-6 and found that 1e-5 provided the best training and testing accuracy. Finally, we found that the optimal learning rate was around 1e-3. We used $30,000$ data points to train the model with batch size of 200. Each of the data points is a list of numbers between 0 and 100. For a particular value of input size $x$ (representing the number of points to be sorted), we varied the number of hidden units from $2$ to $20$ and the number of hidden layers from $2$ (just an input and an output layer) to $x!+2.$ Then, the test loss was calculated on 20,000 points, chosen from same distribution. 
Finally, for each input size, Table~\ref{table:sortingApp} records the combination (hidden\_layer, hidden\_unit) for which the loss is less than 5 and (hidden\_ layer * hidden\_units) is minimized. As seen from the table, the system failed to find any optimal combination for any input size greater than or equal to 5. For example, for input size 5, the hidden units were upper capped at 20 and hidden layers at 120 and thus 2400 parameters were insufficient to sort 5 numbers.

\begin{table}[hbt]
  \centering
  \caption{Minimal configurations to sort list of length ``input size.''}
  \small
    \begin{tabular}{cccc}
    \toprule
     Input Size & Hidden units & Hidden Layers & Parameters \\
     \cmidrule(r){1-1}
     \cmidrule(r){2-2}
     \cmidrule(r){3-3}
     \cmidrule(r){4-4}
     2 & 6 & 2 & 12\\
     3 & 8 & 4 & 32\\
     4 & 18 & 4 & 72\\
     5 & -  & - & -\\
    \bottomrule
    \end{tabular}
    \label{table:sortingApp}
\end{table}

\textbf{E-2. Generalization}

The model developed above generalizes poorly on data outside the training domain. For example, consider the model with 18 hidden units and four hidden layers, which is successfully trained to sort four numbers chosen from the range 0 to 100. It was first tested on numbers from 0 to 100 and then on 100 to 200. The error in the first case was around 2 while the average error in the second case was between 6 and 8 (which is $(200/100)^2=4$ times the former loss). Finally, when tested on larger ranges such as $(9900,10000),$ the error exploded to around 0.1 million (which is an order greater than $(10000/100)^2=10000$ times the original loss). This gives a hint that the error might be scaling proportionally to the square of the test domain with respect to the train domain. A possible explanation for this comes from the use of the MSE loss function. Scaling test data by $\rho$ scales the absolute error by approximately the same factor and then taking a square of the error to calculate the MSE scales the total loss by the square of that factor, i.e., $\rho^2$.

\textbf{E-3. Applying temperature-controlled connectivity to standard neural networks for MNIST classification}

We would like to demonstrate the promise of temperature-controlled connectivity as a regularization method for the classification heads of models with a very simple experiment.
We used the ResNet50 model to train on the
standard MNIST image classification benchmark. We studied two variants of the model: the standard ResNet model and ResNet augmented with our temperature-controlled connectivity (with $T=1$) between the flattened layer and the last fully connected layer (on the lines discussed in the main paper). Then we trained both models with a learning rate fixed at 0.05 and a batch size of 64 and ran it for 10 epochs. 
The model with regularization performed slightly better than the one without it. The regularized model achieved the maximum accuracy among all methods, 99.18\%, while the same figure for the standard one was 98.43\%. Another interesting observation is that the regularized model produces much more stable and consistent results across iterations than the unregularized model. These results encourage us to study the above regularization
method in larger experiments.

\section{Small Experiments}
\label{sec:moreExperiments}

To demonstrate its ability to fit functions with constants, we also tested OccamNet on the function $10.5x^{3.1}$ without providing either 10.5 or 3.1 beforehand. OccamNet identified the correct function 10 times out of 10, taking an average of 553s.

We also investigated whether OccamNet could discover a formula for cosine using only the primitives $\sin(\cdot),$ $+(\cdot,\cdot),$ and $\divisionsymbol(\cdot,\cdot)$ and the constants $2$ and $\pi.$ We expected OccamNet to discover $\cos(x) = \sin(x+\pi/2),$ but, interestingly, it instead always identified the double angle identity $\cos(x) = \sin(2x)/(2\sin(x)).$ OccamNet successfully identified an identity for cosine 8 out of 10 times and in an average of 410s. A more optimized implementation of OccamNet takes only 7s for the same task, although its accuracy is somewhat lower, fluctuating between 2 and 6 out of 10 correct identifications.

Similarly, we tested whether OccamNet could discover Taylor polynomials of $e^x.$ OccamNet identified $e^x\approx 1+x+x^2/2,$ but was unable to discover the subsequent $x^3/6$ term.

\section{Related Work}
\label{sec:related}

\subsection{Symbolic regression}
OccamNet was partially inspired by the EQL network \cite{EQLOriginal, EQLWithDivision, Kim2019IntegrationON}, a neural network-based symbolic regression system that successfully finds simple analytic functions. Neural Arithmetic Logic Units (NALU) and related models \cite{NIPS2018_8027, Madsen2020Neural} provide a neural inductive bias for arithmetic in neural networks that can in principle fit some benchmarks in Table~\ref{table:analyticbenchmarks}.
NALU updates the weights by backpropagating through the activations, shaping the neural network towards a gating interpretation of the linear layers. However, generalizing those models to a diverse set of function primitives might be a formidable task: from our experiments, backpropagation through some activation functions (such as division or sine) makes training considerably harder. In a different computational paradigm, genetic programming (GP) has performed exceptionally well at symbolic regression \cite{schmidt2009distilling, Udrescu_2020}, and a number of evolution-inspired, probability-based models have been explored for this goal \cite{mckay2010grammar}. 

A concurrent work~\citep{petersen2021deep} explores deep symbolic regression by using an RNN to search the space of expressions using autoregressive expression generation. Interestingly, the authors observed that a risk-aware reinforcement learning approach is a necessary component in their optimization, which is similar to our approach of selecting the top $\lambda$ function for optimization in Step 2 of our algorithm. A notable difference is that OccamNet does not generate the expressions autoregressively, although it still exhibits a gradual increase in modularity during training, as discussed in Section~\ref{sec:discussion}. This is also a benefit both for speed and scalability. Moreover, their entropy regularization is a potentially useful addition to our training algorithm. Marrying our approach with theirs is a promising direction for future work.

Transformer-based models can quickly and accurately identify functions given data by leveraging their extensive pretraining. However, these approaches are limited in that they are restricted to a set of primitive functions specified at training time. It may thus be fruitful to investigate combining OccamNet and such approaches in a way that increases convergence speed while maintaining OccamNet's flexibility.

\subsection{Program synthesis}
A field related to symbolic regression is program synthesis. For programs, one option to fit programs is to use EQL-based models with logic activations (step functions, $\mathsf{MIN}$, $\mathsf{MAX}$, etc.) approximated by sigmoid activations. Another is probabilistic program induction using domain-specific languages \cite{NIPS2018_7845, NIPS2018_8006, NIPS2019_9116}. Neural Turing Machines~\cite{Graves2014NeuralTM, Graves2016HybridCU} and their stable versions~\cite{CollierBeel2018} are also able to discover interpretable programs based on observations of input-output pairs. They do so by simulating programs using neural networks connected to an external memory. \citet{balog2016deepcoder} first train a machine learning model to predict a DSL based on input-output pairs and then use methods from satisfiability modulo theory~\cite{SolarLezama:EECS-2008-177} to search the space of programs built using the predicted DSL. In contrast, our DSL is lower level and can fit components like ``sort'' instead of including them in the DSL directly. \citet{kurach2015neural} develop a neural model for simple algorithmic tasks by utilizing memory access for pointer manipulation and dereferencing. However, here we achieve similar results (for example, sorting) without external memory and in only minutes on a CPU.

\subsection{Integration with deep learning}
We are not aware of classifiers that predict MNIST or ImageNet labels using symbolic rules. The closest baseline we found is using GP~\cite{10.5555/1623755.1623876}, which performs comparably well to our neural method, but cannot easily integrate with deep learning. In the reinforcement learning (RL) domain, \citet{such2017deep} and \citet{ salimans2017evolution} propose training deep models of millions of parameters on standard RL tasks using a gradient-free GP, which is competitive to gradient-based RL algorithms. Work such as \citet{Kim2019IntegrationON} performs similar tasks with EQL, a less powerful symbolic regression model.

\subsection{SCGs and pruning}
Treating the problem of finding the correct function or program as a stochastic computational graph is appealing due to efficient soft approximations to discrete distributions~\cite{DBLP:journals/corr/MaddisonMT16,45822,DBLP:journals/corr/TuckerMMS17}. Our $T$-softmax layers offer such an approximation and could further benefit from an adaptive softmax methodology~\cite{grave2016efficient}, which we leave for future work.
Furthermore, the sparsity induced by $T$-softmax layers parallels the abundant work on pruning connections and weights in neural networks~\cite{han2015learning, li2016pruning} or using regularizations, encouraging sparse connectivity~\cite{molchanov2017variational,louizos2018learning}.

\section{Information about Symbolic Regression Benchmarks}\label{sec:SymbolicRegression}
\textbf{H-1. Eureqa}

Eureqa is a software package for symbolic regression where one can specify different target expressions, building block functions (analogous to the primitives in OccamNet), and loss functions \cite{schmidt2009distilling}. For most functions, we use the absolute error as the optimization metric. We choose formula building blocks in Eureqa to match the primitive functions used in OccamNet. 

For implicit functions, we use the implicit derivative error. We also order the data to improve the performance. For the implicit functions in lines 1, 3, and 4 in Table 2 of the main text, the data is ordered by $x_0$. For the equation $x_0^2+x_1^2=1$, the data is generated by sampling $\theta\in[0,2\pi)$ and calculating $x_0=\sin(x)$ and $x_1=\cos(x)$, and is ordered by $\theta$. When the data is not ordered, the value of the implicit derivative error is much higher, resulting in the algorithm favoring incorrect equations. For equation $m_1v_1-m_2v_2$, the ordering is more ambiguous because of the higher dimensionality. We tried ordering by both $m_1$ and the product $m_1v_1$ without success.

\textbf{H-2. HeuristicLab}

Due to limits on the number of data points and feature columns in Eureqa, we instead use HeuristicLab for the image recognition tasks described in Section 5.4 of the main text. HeuristicLab is a software package for optimization and evolutionary algorithms, including symbolic regression and symbolic classification. We use the Island Genetic Algorithm with default settings.

Similar to the building block functions in Eureqa, HeuristicLab can specify the primitive symbols for each task. However, HeuristicLab does not have the primitives $\mathsf{MAX}$, $\mathsf{SIGMOID}$, or $\mathsf{tanh}$. Instead, we use the symbols $\mathsf{IfThenElse}$, $\mathsf{GreaterThan}$, $\mathsf{LessThan}$, $\mathsf{And}$, $\mathsf{Or}$, and $\mathsf{Not}$.

\textbf{H-3. Eplex}

As discussed in Section~\ref{sec:PMLBSetup}, Eplex \cite{EPLEX}, short for Epsilon-Lexicase selection, is a genetic programming population selection technique that we use as a symbolic regression benchmark in our experiments with PMLB datasets. We implement a genetic algorithm using Eplex with the DEAP \cite{DEAP} evolutionary framework, using Numpy arrays \cite{Numpy} for computation to increase speed.

Eplex selects individuals from a population by evaluating the individuals on subsets, or fitness cases, of the full data. For each fitness case, Eplex selects the top-performing individuals and proceeds to the next fitness case. This process is repeated until only one individual remains. This individual is then used as the parent for the next generation.

\textbf{H-4. AI Feynman 2.0}
We also benchmark OccamNet against AI Feynman 2.0 \cite{AIFeynman2.0}. AI Feynman 2.0 is a mixed approach that combines brute-force symbolic regression, polynomial fits, and identification for modularity in the data using neural networks. To identify modularities in the data, AI Feynman first trains a neural network on it. This  serves as an interpolating function for the true data and allows the network to search for symmetries and other forms of modularities.

\begin{figure*}[thb]
    \centering  
    \includegraphics[width=0.45\textwidth]{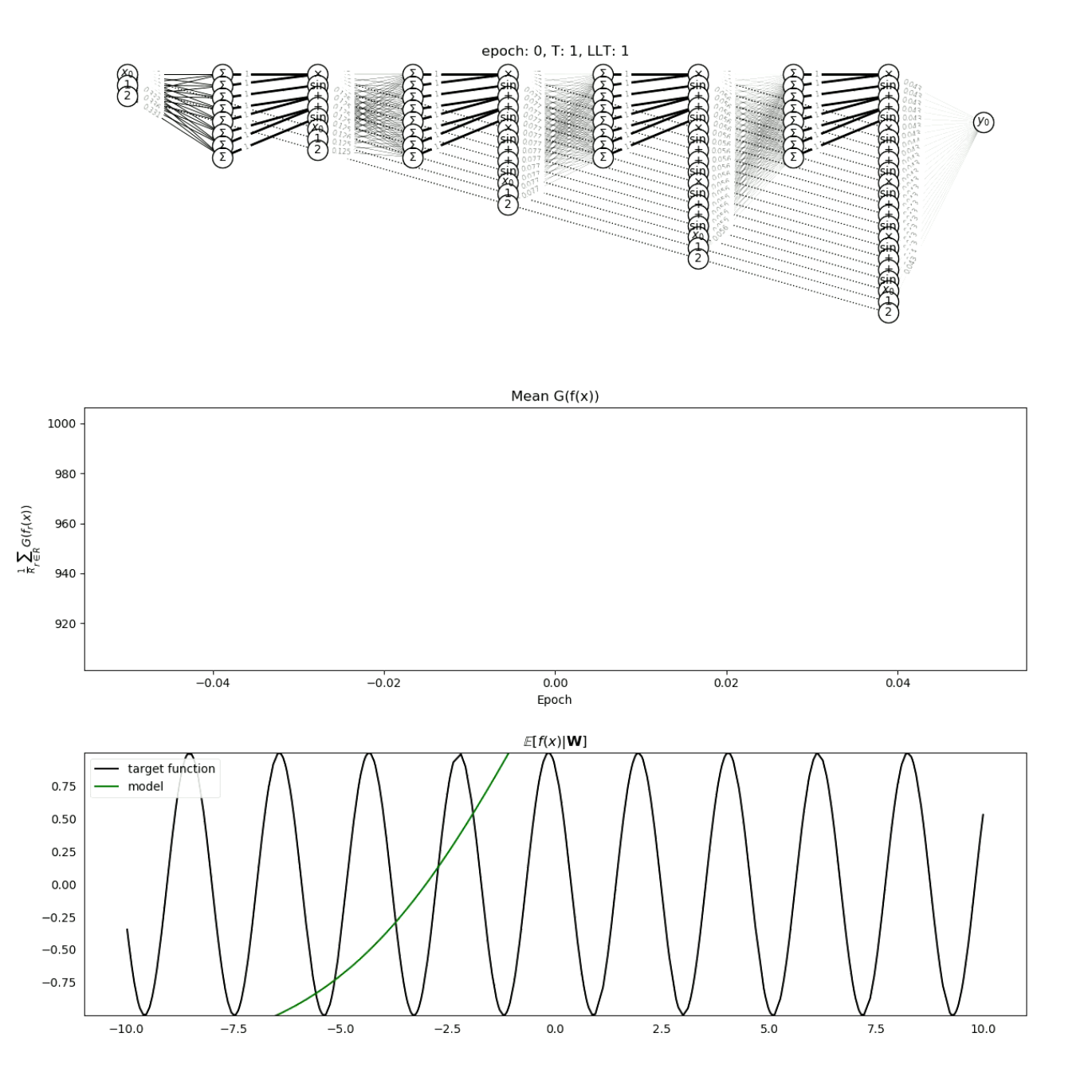}
    \includegraphics[width=0.45\textwidth]{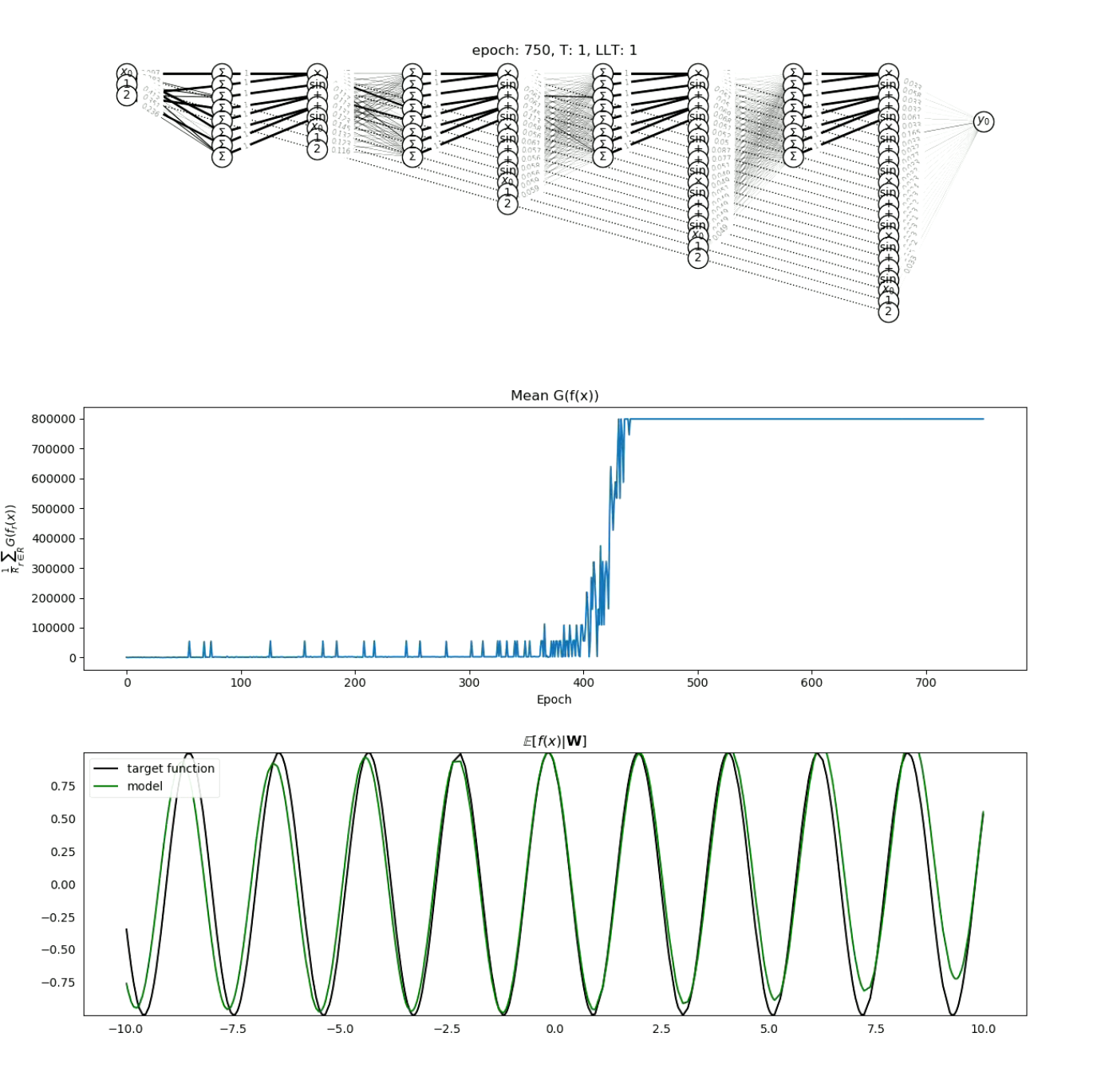}
    \caption{In this figure, we present two video frames for the target $\sin(3x+2)$, which could be accessed via \texttt{videos/sin(3x + 2).mp4} in our code files. We show the beginning of the fitting (left) and the end, where OccamNet has almost converged (right).
    }   
    \label{fig:videos1}
\end{figure*}

\begin{figure*}[thb]
    \centering  
    \includegraphics[width=0.45\textwidth]{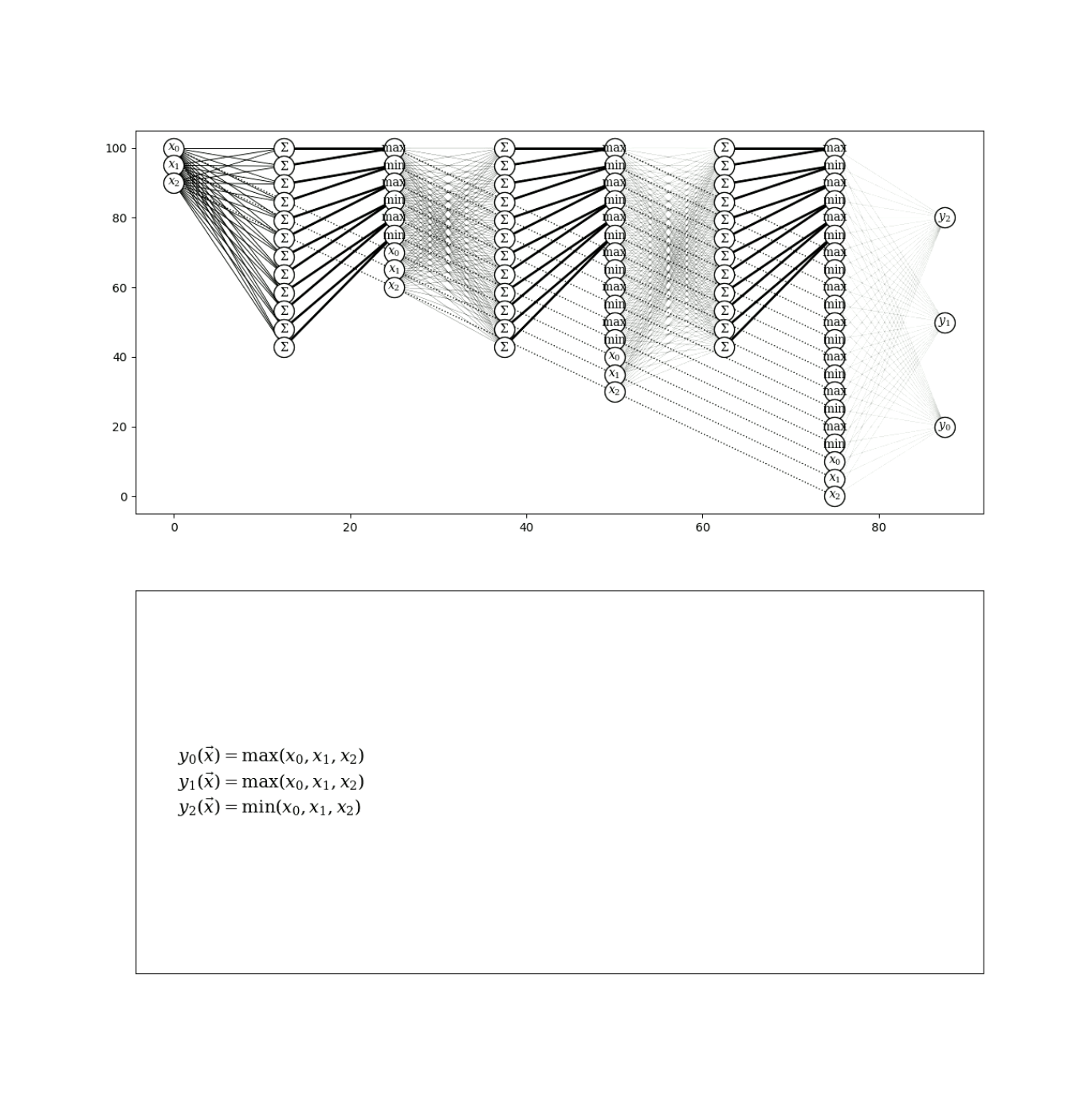}
    \includegraphics[width=0.45\textwidth]{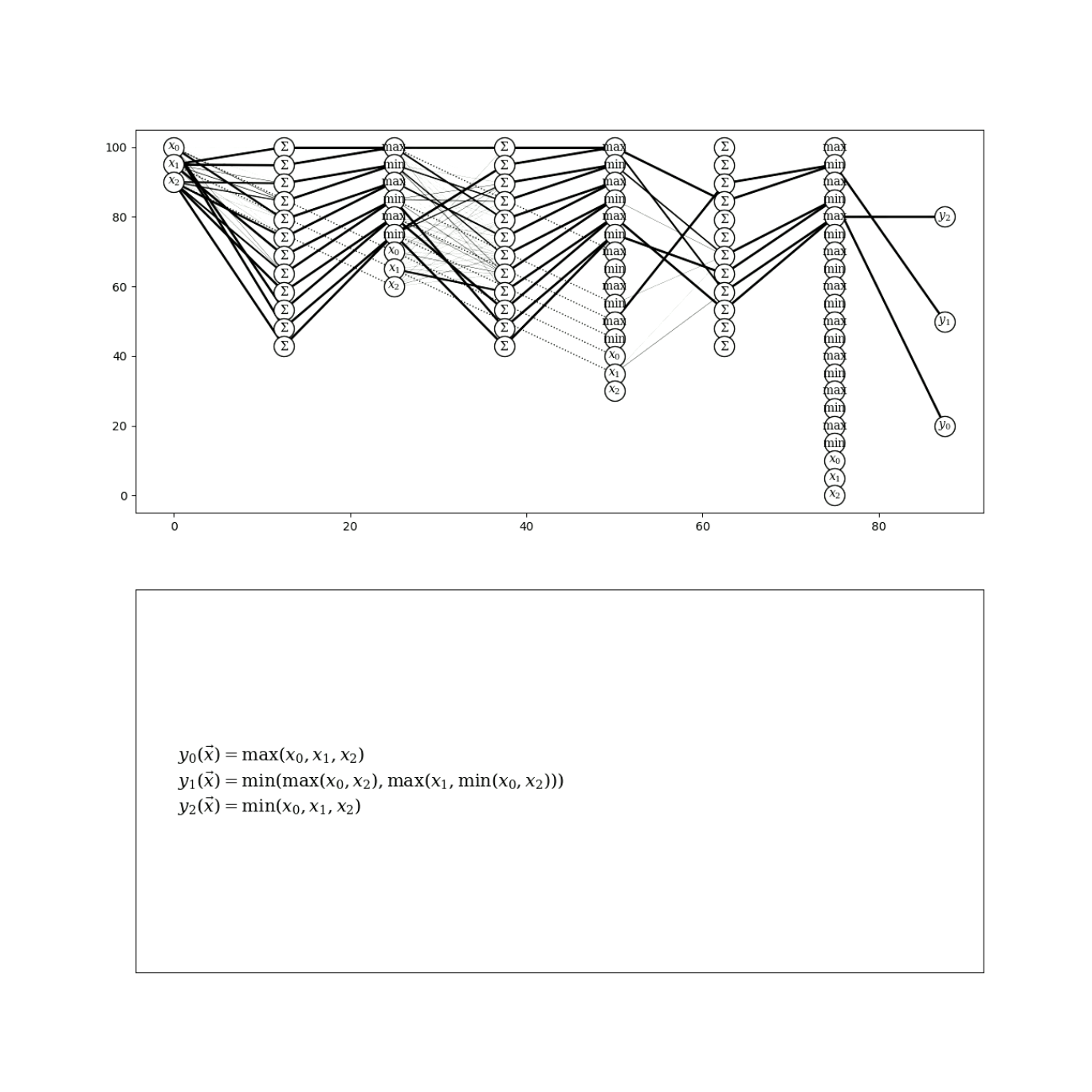}
    \caption{In this figure we present two video frames for the target $\mathsf{SORT}(x_0,x_1,x_2)$, which could be accessed via \texttt{videos/sorting.mp4} in our code files. We show the beginning of the fitting (left) and the end, where OccamNet has almost converged (right).
    }   
    \label{fig:videos2}
\end{figure*}

\section{Code, Videos, and Responsible Use}
\label{sec:codeApp}

Code for all iterations of OccamNet is available at \url{https://github.com/druidowm/OccamNet_Versions}. The code used for this paper can be found at \url{https://github.com/druidowm/OccamNet_Public}. We have grouped our code into five main folders. \texttt{analytic-and-programs} stores our network and experiments for fitting analytic functions and programs. \texttt{implicit} stores our network and experiments for implicit functions, although it also includes the three analytic functions listed in Table \ref{table:hyperparameters2}. \texttt{constant-fitting} stores code very similar to \texttt{implicit} but optimized for constant fitting. \texttt{image-recognition} stores our network and experiments for image classification. \texttt{pmlb-experiment} stores our code for benchmarking against the PMLB regression datasets. Finally, \texttt{videos} stores several videos of our model converging to various functions. In Figures~\ref{fig:videos1} and~\ref{fig:videos2}, we present snapshots of the videos.

Currently, our method is not explicitly designed against adversarial attacks. Thus, malicious stakeholders could exploit our method and manipulate the symbolic fits that OccamNet produces. A potential direction towards alleviating the problem would be to explore ways to robustify OccamNet by training it against an adversary. In the meantime, we ask that users of our code remain responsible and consider the repercussions of their use cases.

\bibliography{biblio}

\begin{thebibliography}{57}
\providecommand{\natexlab}[1]{#1}
\providecommand{\url}[1]{\texttt{#1}}
\expandafter\ifx\csname urlstyle\endcsname\relax
  \providecommand{\doi}[1]{doi: #1}\else
  \providecommand{\doi}{doi: \begingroup \urlstyle{rm}\Url}\fi

\bibitem[LeCun et~al.(2015)LeCun, Bengio, and Hinton]{lecun2015deeplearning}
Yann LeCun, Yoshua Bengio, and Geoffrey Hinton.
\newblock Deep learning.
\newblock \emph{Nature}, 521\penalty0 (7553):\penalty0 436--444, 2015.

\bibitem[Choromanska et~al.(2015)Choromanska, Henaff, Mathieu, Arous, and LeCun]{choromanska2014loss}
Anna Choromanska, Mikael Henaff, Michael Mathieu, Gérard~Ben Arous, and Yann LeCun.
\newblock The loss surfaces of multilayer networks.
\newblock In \emph{AISTATS}, 2015.

\bibitem[Lample and Charton(2020)]{Lample2020Deep}
Guillaume Lample and François Charton.
\newblock Deep learning for symbolic mathematics.
\newblock In \emph{ICLR}, 2020.

\bibitem[Schmidt and Lipson(2009)]{schmidt2009distilling}
Michael Schmidt and Hod Lipson.
\newblock Distilling free-form natural laws from experimental data.
\newblock \emph{Science}, 324\penalty0 (5923):\penalty0 81--85, 2009.

\bibitem[Udrescu and Tegmark(2020)]{Udrescu_2020}
Silviu-Marian Udrescu and Max Tegmark.
\newblock {AI} {F}eynman: A physics-inspired method for symbolic regression.
\newblock \emph{Science Advances}, 6\penalty0 (16), 2020.

\bibitem[Poli et~al.(2008)Poli, Langdon, McPhee, and Koza]{poli2008field}
Riccardo Poli, William~B Langdon, Nicholas~F McPhee, and John~R Koza.
\newblock \emph{A field guide to genetic programming}.
\newblock lulu.com, 2008.

\bibitem[{Angeline} et~al.(1994){Angeline}, {Saunders}, and {Pollack}]{265960}
P.~J. {Angeline}, G.~M. {Saunders}, and J.~B. {Pollack}.
\newblock An evolutionary algorithm that constructs recurrent neural networks.
\newblock \emph{IEEE Transactions on Neural Networks}, 5\penalty0 (1):\penalty0 54--65, 1994.

\bibitem[Arnold and Hansen(2012)]{10.1145/2330163.2330207}
Dirk~V. Arnold and Nikolaus Hansen.
\newblock A (1+1)-{CMA}-{ES} for constrained optimisation.
\newblock In \emph{GECCO}, 2012.

\bibitem[Such et~al.(2017)Such, Madhavan, Conti, Lehman, Stanley, and Clune]{such2017deep}
Felipe~Petroski Such, Vashisht Madhavan, Edoardo Conti, Joel Lehman, Kenneth~O. Stanley, and Jeff Clune.
\newblock Deep neuroevolution: Genetic algorithms are a competitive alternative for training deep neural networks for reinforcement learning.
\newblock \emph{arXiv preprint arXiv:1712.06567}, 2017.

\bibitem[Hansen(2016)]{hansen2016cma}
Nikolaus Hansen.
\newblock The {CMA} evolution strategy: A tutorial.
\newblock \emph{arXiv preprint arXiv:1604.00772}, 2016.

\bibitem[Loshchilov and Hutter(2016)]{loshchilov2016cmaes}
Ilya Loshchilov and Frank Hutter.
\newblock {CMA}-{ES} for hyperparameter optimization of deep neural networks.
\newblock \emph{arXiv preprint arXiv:1604.07269}, 2016.

\bibitem[Mckay et~al.(2010)Mckay, Hoai, Whigham, Shan, and O’neill]{mckay2010grammar}
Robert~I Mckay, Nguyen~Xuan Hoai, Peter~Alexander Whigham, Yin Shan, and Michael O’neill.
\newblock Grammar-based genetic programming: a survey.
\newblock \emph{Genetic Programming and Evolvable Machines}, 11\penalty0 (3-4):\penalty0 365--396, 2010.

\bibitem[Petersen et~al.(2021)Petersen, Larma, Mundhenk, Santiago, Kim, and Kim]{petersen2021deep}
Brenden~K Petersen, Mikel~Landajuela Larma, Terrell~N. Mundhenk, Claudio~Prata Santiago, Soo~Kyung Kim, and Joanne~Taery Kim.
\newblock Deep symbolic regression: Recovering mathematical expressions from data via risk-seeking policy gradients.
\newblock In \emph{ICLR}, 2021.

\bibitem[{Biggio} et~al.(2021){Biggio}, {Bendinelli}, {Neitz}, {Lucchi}, and {Parascandolo}]{BiggioNeuroSymResScales}
Luca {Biggio}, Tommaso {Bendinelli}, Alexander {Neitz}, Aurelien {Lucchi}, and Giambattista {Parascandolo}.
\newblock {Neural Symbolic Regression that Scales}.
\newblock \emph{arXiv e-prints}, art. arXiv:2106.06427, June 2021.
\newblock \doi{10.48550/arXiv.2106.06427}.

\bibitem[Ellis et~al.(2018{\natexlab{a}})Ellis, Ritchie, Solar-Lezama, and Tenenbaum]{NIPS2018_7845}
Kevin Ellis, Daniel Ritchie, Armando Solar-Lezama, and Josh Tenenbaum.
\newblock Learning to infer graphics programs from hand-drawn images.
\newblock In \emph{NIPS}. 2018{\natexlab{a}}.

\bibitem[Ellis et~al.(2018{\natexlab{b}})Ellis, Morales, Sabl\'{e}-Meyer, Solar-Lezama, and Tenenbaum]{NIPS2018_8006}
Kevin Ellis, Lucas Morales, Mathias Sabl\'{e}-Meyer, Armando Solar-Lezama, and Josh Tenenbaum.
\newblock Learning libraries of subroutines for neurally\textendash guided bayesian program induction.
\newblock In \emph{NIPS}. 2018{\natexlab{b}}.

\bibitem[Ellis et~al.(2019)Ellis, Nye, Pu, Sosa, Tenenbaum, and Solar-Lezama]{NIPS2019_9116}
Kevin Ellis, Maxwell Nye, Yewen Pu, Felix Sosa, Josh Tenenbaum, and Armando Solar-Lezama.
\newblock Write, execute, assess: Program synthesis with a {REPL}.
\newblock In \emph{NeurIPS}. 2019.

\bibitem[Balog et~al.(2016)Balog, Gaunt, Brockschmidt, Nowozin, and Tarlow]{balog2016deepcoder}
Matej Balog, Alexander~L. Gaunt, Marc Brockschmidt, Sebastian Nowozin, and Daniel Tarlow.
\newblock Deepcoder: Learning to write programs.
\newblock In \emph{ICLR}, 2016.

\bibitem[Solar~Lezama(2008)]{SolarLezama:EECS-2008-177}
Armando Solar~Lezama.
\newblock \emph{Program Synthesis By Sketching}.
\newblock PhD thesis, EECS Department, University of California, Berkeley, 2008.

\bibitem[Trask et~al.(2018)Trask, Hill, Reed, Rae, Dyer, and Blunsom]{NIPS2018_8027}
Andrew Trask, Felix Hill, Scott~E Reed, Jack Rae, Chris Dyer, and Phil Blunsom.
\newblock Neural arithmetic logic units.
\newblock In \emph{NIPS}. 2018.

\bibitem[Madsen and Johansen(2020)]{Madsen2020Neural}
Andreas Madsen and Alexander~Rosenberg Johansen.
\newblock Neural arithmetic units.
\newblock In \emph{ICLR}, 2020.

\bibitem[Graves et~al.(2014)Graves, Wayne, and Danihelka]{Graves2014NeuralTM}
Alex Graves, Greg Wayne, and Ivo Danihelka.
\newblock Neural turing machines.
\newblock \emph{arXiv preprint arXiv:1410.5401}, 2014.

\bibitem[Graves et~al.(2016)Graves, Wayne, Reynolds, Harley, Danihelka, Grabska-Barwinska, Colmenarejo, Grefenstette, Ramalho, Agapiou, Badia, Hermann, Zwols, Ostrovski, Cain, King, Summerfield, Blunsom, Kavukcuoglu, and Hassabis]{Graves2016HybridCU}
Alex Graves, Greg Wayne, Malcolm Reynolds, Tim Harley, Ivo Danihelka, Agnieszka Grabska-Barwinska, Sergio~G{\'o}mez Colmenarejo, Edward Grefenstette, Tiago Ramalho, John Agapiou, Adri{\`a}~Puigdom{\`e}nech Badia, Karl~Moritz Hermann, Yori Zwols, Georg Ostrovski, Adam Cain, Helen. King, C.~Summerfield, Phil Blunsom, Koray Kavukcuoglu, and Demis Hassabis.
\newblock Hybrid computing using a neural network with dynamic external memory.
\newblock \emph{Nature}, 538:\penalty0 471--476, 2016.

\bibitem[Collier and Beel(2018)]{CollierBeel2018}
Mark Collier and Joeran Beel.
\newblock Implementing neural turing machines.
\newblock In \emph{ICANN}, page 94–104, 2018.

\bibitem[{Martius} and {Lampert}(2016)]{EQLOriginal}
Georg {Martius} and Christoph~H. {Lampert}.
\newblock {Extrapolation and learning equations}.
\newblock \emph{arXiv e-prints}, art. arXiv:1610.02995, October 2016.

\bibitem[Sahoo et~al.(2018)Sahoo, Lampert, and Martius]{EQLWithDivision}
Subham Sahoo, Christoph Lampert, and Georg Martius.
\newblock Learning equations for extrapolation and control.
\newblock In Jennifer Dy and Andreas Krause, editors, \emph{Proceedings of the 35th International Conference on Machine Learning}, volume~80 of \emph{Proceedings of Machine Learning Research}, pages 4442--4450, Stockholmsmässan, Stockholm Sweden, 10--15 Jul 2018. PMLR.
\newblock URL \url{http://proceedings.mlr.press/v80/sahoo18a.html}.

\bibitem[Kim et~al.(2020)Kim, Lu, Mukherjee, Gilbert, Jing, Čeperić, and Soljačić]{Kim2019IntegrationON}
Samuel Kim, Peter~Y. Lu, Srijon Mukherjee, Michael Gilbert, Li~Jing, Vladimir Čeperić, and Marin Soljačić.
\newblock Integration of neural network-based symbolic regression in deep learning for scientific discovery.
\newblock \emph{IEEE Transactions on Neural Networks and Learning Systems}, pages 1--12, 2020.

\bibitem[He et~al.(2016)He, Zhang, Ren, and Sun]{he2016deep}
Kaiming He, Xiangyu Zhang, Shaoqing Ren, and Jian Sun.
\newblock Deep residual learning for image recognition.
\newblock In \emph{CVPR}, pages 770--778, 2016.

\bibitem[Balla et~al.(2022)Balla, Huang, Dugan, Dangovski, and Soljacic]{balla2022ai}
Julia Balla, Sihao Huang, Owen Dugan, Rumen Dangovski, and Marin Soljacic.
\newblock Ai-assisted discovery of quantitative and formal models in social science.
\newblock \emph{arXiv preprint arXiv:2210.00563}, 2022.

\bibitem[Fowler(2010)]{10.5555/1809745}
Martin Fowler.
\newblock \emph{Domain Specific Languages}.
\newblock Addison-Wesley Professional, 1st edition, 2010.

\bibitem[Huang et~al.(2017)Huang, Liu, Van Der~Maaten, and Weinberger]{huang2017densely}
Gao Huang, Zhuang Liu, Laurens Van Der~Maaten, and Kilian~Q Weinberger.
\newblock Densely connected convolutional networks.
\newblock In \emph{CVPR}, 2017.

\bibitem[{La Cava} et~al.(2016){La Cava}, {Spector}, and {Danai}]{EPLEX}
William {La Cava}, Lee {Spector}, and Kourosh {Danai}.
\newblock {Epsilon-Lexicase Selection for Regression}.
\newblock In \emph{GECCO}, 2016.

\bibitem[Udrescu et~al.(2020)Udrescu, Tan, Feng, Neto, Wu, and Tegmark]{AIFeynman2.0}
Silviu-Marian Udrescu, Andrew Tan, Jiahai Feng, Orisvaldo Neto, Tailin Wu, and Max Tegmark.
\newblock {AI} {F}eynman 2.0: Pareto-optimal symbolic regression exploiting graph modularity.
\newblock 2020.

\bibitem[Wagner et~al.(2014)Wagner, Kronberger, Beham, Kommenda, Scheibenpflug, Pitzer, Vonolfen, Kofler, Winkler, Dorfer, and Affenzeller]{wagner2014}
Stefan Wagner, Gabriel Kronberger, Andreas Beham, Michael Kommenda, Andreas Scheibenpflug, Erik Pitzer, Stefan Vonolfen, Monika Kofler, Stephan Winkler, Viktoria Dorfer, and Michael Affenzeller.
\newblock \emph{Advanced Methods and Applications in Computational Intelligence}, volume~6, chapter Architecture and Design of the HeuristicLab Optimization Environment, pages 197--261.
\newblock Springer, 2014.

\bibitem[LeCun et~al.(1998)LeCun, Bottou, Bengio, and Haffner]{lecun-gradientbased-learning-applied-1998}
Yann LeCun, Léon Bottou, Yoshua Bengio, and Patrick Haffner.
\newblock Gradient-based learning applied to document recognition.
\newblock In \emph{IEEE}, 1998.

\bibitem[Deng et~al.(2009)Deng, Dong, Socher, Li, Li, and Fei-Fei]{imagenet_cvpr09}
J.~Deng, W.~Dong, R.~Socher, L.-J. Li, K.~Li, and L.~Fei-Fei.
\newblock {ImageNet: A Large-Scale Hierarchical Image Database}.
\newblock In \emph{CVPR}, 2009.

\bibitem[Olson et~al.(2017)Olson, La~Cava, Orzechowski, Urbanowicz, and Moore]{PMLB}
Randal~S. Olson, William La~Cava, Patryk Orzechowski, Ryan~J. Urbanowicz, and Jason~H. Moore.
\newblock {PMLB}: a large benchmark suite for machine learning evaluation and comparison.
\newblock \emph{BioData Mining}, 10\penalty0 (1):\penalty0 36, 2017.

\bibitem[Schmidt and Lipson(2010)]{ImplicitFitting}
Michael Schmidt and Hod Lipson.
\newblock \emph{Symbolic Regression of Implicit Equations}, pages 73--85.
\newblock Springer US, 2010.

\bibitem[Williams(1992)]{10.1007/BF00992696}
Ronald~J. Williams.
\newblock Simple statistical gradient-following algorithms for connectionist reinforcement learning.
\newblock \emph{Mach. Learn.}, 8\penalty0 (3–4):\penalty0 229–256, May 1992.

\bibitem[Kingma and Ba(2015)]{Kingma2015AdamAM}
Diederik~P. Kingma and Jimmy Ba.
\newblock Adam: A method for stochastic optimization.
\newblock In \emph{ICLR}, 2015.

\bibitem[Meurer et~al.(2017)Meurer, Smith, Paprocki, \v{C}ert\'ik, Kirpichev, Rocklin, Kumar, Ivanov, Moore, Singh, Rathnayake, Vig, Granger, Muller, Bonazzi, Gupta, Vats, Johansson, Pedregosa, Curry, Terrel, Rou\v{c}ka, Saboo, Fernando, Kulal, Cimrman, and Scopatz]{SymPy}
Aaron Meurer, Christopher~P. Smith, Mateusz Paprocki, Ond\v{r}ej \v{C}ert\'ik, Sergey~B. Kirpichev, Matthew Rocklin, AMiT Kumar, Sergiu Ivanov, Jason~K. Moore, Sartaj Singh, Thilina Rathnayake, Sean Vig, Brian~E. Granger, Richard~P. Muller, Francesco Bonazzi, Harsh Gupta, Shivam Vats, Fredrik Johansson, Fabian Pedregosa, Matthew~J. Curry, Andy~R. Terrel, {\v{S}}t\v{e}p\'an Rou\v{c}ka, Ashutosh Saboo, Isuru Fernando, Sumith Kulal, Robert Cimrman, and Anthony Scopatz.
\newblock {SymPy}: symbolic computing in python.
\newblock \emph{Peer J Computer Science}, 3:\penalty0 103, 2017.

\bibitem[{Chen} and {Guestrin}(2016)]{XGBoost}
Tianqi {Chen} and Carlos {Guestrin}.
\newblock {XGBoost: A Scalable Tree Boosting System}.
\newblock \emph{arXiv e-prints}, art. arXiv:1603.02754, March 2016.

\bibitem[{Orzechowski} et~al.(2018){Orzechowski}, {La Cava}, and {Moore}]{WhereAreWe}
Patryk {Orzechowski}, William {La Cava}, and Jason~H. {Moore}.
\newblock {Where are we now? A large benchmark study of recent symbolic regression methods}.
\newblock In \emph{GECCO}, 2018.

\bibitem[Fortin et~al.(2012)Fortin, {De Rainville}, Gardner, Parizeau, and Gagn\'e]{DEAP}
F\'elix-Antoine Fortin, Fran\c{c}ois-Michel {De Rainville}, Marc-Andr\'e Gardner, Marc Parizeau, and Christian Gagn\'e.
\newblock {DEAP}: Evolutionary algorithms made easy.
\newblock \emph{Journal of Machine Learning Research}, 13:\penalty0 2171--2175, 2012.

\bibitem[{Melo} et~al.(2019){Melo}, {Vasconcellos Vargas}, and {Banzhaf}]{EplexIsSlow}
Vinicius~V. {Melo}, Danilo {Vasconcellos Vargas}, and Wolfgang {Banzhaf}.
\newblock {Batch Tournament Selection for Genetic Programming}.
\newblock In \emph{GECCO}, 2019.

\bibitem[Kurach et~al.(2016)Kurach, Andrychowicz, and Sutskever]{kurach2015neural}
Karol Kurach, Marcin Andrychowicz, and Ilya Sutskever.
\newblock Neural random-access machines.
\newblock In \emph{ICLR}, 2016.

\bibitem[Montana and Davis(1989)]{10.5555/1623755.1623876}
David~J. Montana and Lawrence Davis.
\newblock Training feedforward neural networks using genetic algorithms.
\newblock In \emph{IJCAI}. Morgan Kaufmann Publishers Inc., 1989.

\bibitem[Salimans et~al.(2017)Salimans, Ho, Chen, Sidor, and Sutskever]{salimans2017evolution}
Tim Salimans, Jonathan Ho, Xi~Chen, Szymon Sidor, and Ilya Sutskever.
\newblock Evolution strategies as a scalable alternative to reinforcement learning.
\newblock \emph{arXiv preprint arXiv:1703.03864}, 2017.

\bibitem[Maddison et~al.(2017)Maddison, Mnih, and Teh]{DBLP:journals/corr/MaddisonMT16}
Chris~J. Maddison, Andriy Mnih, and Yee~Whye Teh.
\newblock The concrete distribution: {A} continuous relaxation of discrete random variables.
\newblock In \emph{ICLR}, 2017.

\bibitem[Jang et~al.(2017)Jang, Gu, and Poole]{45822}
Eric Jang, Shixiang Gu, and Ben Poole.
\newblock Categorical reparameterization with gumbel-softmax.
\newblock In \emph{ICLR}, 2017.

\bibitem[Tucker et~al.(2017)Tucker, Mnih, Maddison, and Sohl{-}Dickstein]{DBLP:journals/corr/TuckerMMS17}
George Tucker, Andriy Mnih, Chris~J. Maddison, and Jascha Sohl{-}Dickstein.
\newblock {REBAR:} low-variance, unbiased gradient estimates for discrete latent variable models.
\newblock In \emph{NIPS}, 2017.

\bibitem[Grave et~al.(2017)Grave, Joulin, Cissé, Grangier, and Jégou]{grave2016efficient}
Edouard Grave, Armand Joulin, Moustapha Cissé, David Grangier, and Hervé Jégou.
\newblock Efficient softmax approximation for {GPU}s.
\newblock In \emph{ICML}, 2017.

\bibitem[Han et~al.(2015)Han, Pool, Tran, and Dally]{han2015learning}
Song Han, Jeff Pool, John Tran, and William~J. Dally.
\newblock Learning both weights and connections for efficient neural networks.
\newblock 2015.

\bibitem[Li et~al.(2017)Li, Kadav, Durdanovic, Samet, and Graf]{li2016pruning}
Hao Li, Asim Kadav, Igor Durdanovic, Hanan Samet, and Hans~Peter Graf.
\newblock Pruning filters for efficient convnets.
\newblock In \emph{ICLR}, 2017.

\bibitem[Molchanov et~al.(2017)Molchanov, Ashukha, and Vetrov]{molchanov2017variational}
Dmitry Molchanov, Arsenii Ashukha, and Dmitry Vetrov.
\newblock Variational dropout sparsifies deep neural networks.
\newblock In \emph{ICML}, 2017.

\bibitem[Louizos et~al.(2018)Louizos, Welling, and Kingma]{louizos2018learning}
Christos Louizos, Max Welling, and Diederik~P. Kingma.
\newblock Learning sparse neural networks through {$L_0$} regularization.
\newblock In \emph{ICLR}, 2018.

\bibitem[Harris et~al.(2020)Harris, Millman, van~der Walt, Gommers, Virtanen, Cournapeau, Wieser, Taylor, Berg, Smith, Kern, Picus, Hoyer, van Kerkwijk, Brett, Haldane, del R{\'{i}}o, Wiebe, Peterson, G{\'{e}}rard-Marchant, Sheppard, Reddy, Weckesser, Abbasi, Gohlke, and Oliphant]{Numpy}
Charles~R. Harris, K.~Jarrod Millman, St{\'{e}}fan~J. van~der Walt, Ralf Gommers, Pauli Virtanen, David Cournapeau, Eric Wieser, Julian Taylor, Sebastian Berg, Nathaniel~J. Smith, Robert Kern, Matti Picus, Stephan Hoyer, Marten~H. van Kerkwijk, Matthew Brett, Allan Haldane, Jaime~Fern{\'{a}}ndez del R{\'{i}}o, Mark Wiebe, Pearu Peterson, Pierre G{\'{e}}rard-Marchant, Kevin Sheppard, Tyler Reddy, Warren Weckesser, Hameer Abbasi, Christoph Gohlke, and Travis~E. Oliphant.
\newblock Array programming with {NumPy}.
\newblock \emph{Nature}, 585\penalty0 (7825):\penalty0 357--362, 2020.

\end{thebibliography}
\bibliographystyle{unsrtnat}

\end{document}